\PassOptionsToPackage{unicode}{hyperref}
\PassOptionsToPackage{hyphens}{url}
\PassOptionsToPackage{dvipsnames,svgnames,x11names}{xcolor}
\documentclass[
  authoryear,
  preprint]{elsarticle}

\usepackage{amsmath,amssymb}
\usepackage{iftex}
\ifPDFTeX
  \usepackage[T1]{fontenc}
  \usepackage[utf8]{inputenc}
  \usepackage{textcomp} 
\else 
  \usepackage{unicode-math}
  \defaultfontfeatures{Scale=MatchLowercase}
  \defaultfontfeatures[\rmfamily]{Ligatures=TeX,Scale=1}
\fi
\usepackage{lmodern}
\ifPDFTeX\else  
\fi
\IfFileExists{upquote.sty}{\usepackage{upquote}}{}
\IfFileExists{microtype.sty}{
  \usepackage[]{microtype}
  \UseMicrotypeSet[protrusion]{basicmath} 
}{}
\makeatletter
\@ifundefined{KOMAClassName}{
  \IfFileExists{parskip.sty}{%
    \usepackage{parskip}
  }{
    \setlength{\parindent}{0pt}
    \setlength{\parskip}{6pt plus 2pt minus 1pt}}
}{
  \KOMAoptions{parskip=half}}
\makeatother
\usepackage{xcolor}
\ifx\paragraph\undefined\else
  \let\oldparagraph\paragraph
  \renewcommand{\paragraph}[1]{\oldparagraph{#1}\mbox{}}
\fi
\ifx\subparagraph\undefined\else
  \let\oldsubparagraph\subparagraph
  \renewcommand{\subparagraph}[1]{\oldsubparagraph{#1}\mbox{}}
\fi

\providecommand{\tightlist}{%
  \setlength{\itemsep}{0pt}\setlength{\parskip}{0pt}}\usepackage{longtable,booktabs,array}
\usepackage{calc} 
\usepackage{etoolbox}
\makeatletter
\patchcmd\longtable{\par}{\if@noskipsec\mbox{}\fi\par}{}{}
\makeatother
\IfFileExists{footnotehyper.sty}{\usepackage{footnotehyper}}{\usepackage{footnote}}
\makesavenoteenv{longtable}
\usepackage{graphicx}
\makeatletter
\def\maxwidth{\ifdim\Gin@nat@width>\linewidth\linewidth\else\Gin@nat@width\fi}
\def\maxheight{\ifdim\Gin@nat@height>\textheight\textheight\else\Gin@nat@height\fi}
\makeatother
\setkeys{Gin}{width=\maxwidth,height=\maxheight,keepaspectratio}
\makeatletter
\def\fps@figure{htbp}
\makeatother

\makeatletter
\@ifpackageloaded{caption}{}{\usepackage{caption}}
\AtBeginDocument{%
\ifdefined\contentsname
  \renewcommand*\contentsname{Table of contents}
\else
  \newcommand\contentsname{Table of contents}
\fi
\ifdefined\listfigurename
  \renewcommand*\listfigurename{List of Figures}
\else
  \newcommand\listfigurename{List of Figures}
\fi
\ifdefined\listtablename
  \renewcommand*\listtablename{List of Tables}
\else
  \newcommand\listtablename{List of Tables}
\fi
\ifdefined\figurename
  \renewcommand*\figurename{Figure}
\else
  \newcommand\figurename{Figure}
\fi
\ifdefined\tablename
  \renewcommand*\tablename{Table}
\else
  \newcommand\tablename{Table}
\fi
}
\@ifpackageloaded{float}{}{\usepackage{float}}
\floatstyle{ruled}
\@ifundefined{c@chapter}{\newfloat{codelisting}{h}{lop}}{\newfloat{codelisting}{h}{lop}[chapter]}
\floatname{codelisting}{Listing}

\makeatother
\makeatletter
\@ifpackageloaded{caption}{}{\usepackage{caption}}
\@ifpackageloaded{subcaption}{}{\usepackage{subcaption}}
\makeatother
\journal{—}
\ifLuaTeX
  \usepackage{selnolig}  
\fi
\usepackage[]{natbib}
\usepackage{bookmark}

\IfFileExists{xurl.sty}{\usepackage{xurl}}{} 
\hypersetup{
  pdftitle={Machine learning enabled velocity model building with uncertainty quantification},
  pdfauthor={Rafael Orozco; Huseyin Tuna Erdinc; Yunlin Zeng; Mathias Louboutin; Felix J. Herrmann},
  colorlinks=true,
  linkcolor={blue},
  filecolor={Maroon},
  citecolor={Blue},
  urlcolor={Blue},
  pdfcreator={LaTeX via pandoc}}

\begin{document}

\begin{frontmatter}
\title{Machine learning enabled velocity model building with uncertainty
quantification}
\author[1]{Rafael Orozco%
}
 \ead{rorozco@gatech.edu} 
\author[1]{Huseyin Tuna Erdinc%
}

\author[1]{Yunlin Zeng%
}

\author[2]{Mathias Louboutin%
}

\author[1]{Felix J. Herrmann%
}

\affiliation[1]{organization={Georgia Institute of
Technology},,postcodesep={}}
\affiliation[2]{organization={Devito Codes},,postcodesep={}}

\cortext[cor1]{Corresponding author}

\begin{abstract}
Accurately characterizing migration velocity models is crucial for a
wide range of geophysical applications, from hydrocarbon exploration to
monitoring of CO\textsubscript{2} sequestration projects. Traditional
velocity model building methods such as Full-Waveform Inversion (FWI)
are powerful but often struggle with the inherent complexities of the
inverse problem, including noise, limited bandwidth, receiver aperture
and computational constraints. To address these challenges, we propose a
scalable methodology that integrates generative modeling, in the form of
Diffusion networks, with physics-informed summary statistics, making it
suitable for complicated imaging problems including field datasets. By
defining these summary statistics in terms of subsurface-offset image
volumes for poor initial velocity models, our approach allows for
computationally efficient generation of Bayesian posterior samples for
migration velocity models that offer a useful assessment of uncertainty.
To validate our approach, we introduce a battery of tests that measure
the quality of the inferred velocity models, as well as the quality of
the inferred uncertainties. With modern synthetic datasets, we reconfirm
gains from using subsurface-image gathers as the conditioning
observable. For complex velocity model building involving salt, we
propose a new iterative workflow that refines amortized posterior
approximations with salt flooding and demonstrate how the uncertainty in
the velocity model can be propagated to the final product reverse time
migrated images. Finally, we present a proof of concept on field
datasets to show that our method can scale to industry-sized problems.
\end{abstract}

\end{frontmatter}
    
\section{Introduction}\label{introduction}

Subsurface characterization of the earth's subsurface is important for
hydrocarbon exploration \citep{zhang2022high}, monitoring of
CO\textsubscript{2} storage projects \citep{gahlot2024uncertainty},
geothermal energy projects \citep{louie2012advanced} and various other
applications \citep{wagner2021overview}. Generally, subsurface
characterization is achieved by observing the interaction between
specific physical phenomena (such as electrodynamics, gravity, and
acoustic wave propagation) and subsurface properties. This tomographic
information is then resolved into images that are analyzed for different
characterization requirements. Although our framework is generally
applicable, we focus on characterization of acoustic properties by means
of probing the subsurface with acoustic waves. Out of the various
methods, FWI stands out as a powerful tool due to its ability to resolve
high-quality acoustic images in complex structures
\citep{tarantola1984inversion}. In spite of its advantages, FWI still
has shortcomings due in part to the nature of the problem but also due
to the specific computational challenges that FWI brings since it
requires solving many wave-equation partial differential equations
(PDEs).

The particular challenges of the FWI inverse problem are that the
observations are corrupted by noise, are limited in frequency bandwidth,
computational simulations will always contain some approximation to the
true physics and due to practical engineering considerations are mostly
restricted to sensing the upcoming waves at the surface so will suffer
from some sort of limited aperture. All of these factors contribute to
FWI's ill-posed nature in the sense that many subsurface scenarios are
capable of explaining the limited data available. Traditional workflows
approach this challenge by introducing prior information to regularize
the vast search space, such as Total Variation (TV)
\citep{esser2018total}. While this has served well to produce
deterministic solutions, it does not express the multi-solution nature
of FWI and does not represent a realistic prior of the multiscale
complexity of Earth structures.

We use Bayesian inference as a principled framework to combine
observable data (seismic shot gathers) with prior knowledge (training
samples) and output a family of Earth models that give users practical
uncertainty quantification (UQ). One of the major gaps in geophysical
inversion is the difficulty of computing Bayesian posterior
distributions that are grounded in useful prior information and
incorporate the complex physics of the problem. Existing approaches
either are too expensive to scale to large-scale problems
(sampling-based methods), fail to capture the full extent of uncertainty
(local methods), or rely on approximations that weaken the physical
fidelity of the results (convolution-based methods).

As a form of variational inference \citep{jordan1999introduction}, our
approach sidesteps the computational problem of sampling the posterior
distribution by optimizing instead for an amortized (read generalized)
approximation to the posterior that is learned from training examples
and can be applied computationally efficiently at test time with small
compute (as measured by PDE solves) on various datasets.

\section{Paper outline}\label{paper-outline}

First, we introduce the wave-based inversion problem that we aim to
solve. Due to the ill-posed nature of the problem, we use a Bayesian
formulation to solve it. Then, we discuss related work that has solved
various aspects of this problem, highlighting the gap in this literature
that we aim to fill. We explain our particular methods, which include
simulation-based inference (SBI) and conditional Diffusion networks as
the core generative network. We then explain the use of physical summary
statistics to efficiently incorporate the wave physics into the method.
To evaluate the quality of our posterior distributions, we propose four
distinct metrics, each targeting a different aspect of UQ. These metrics
are applied to assess the improvements gained from using Common-Image
Gathers (CIGs) in comparison to Reverse-Time Migrations (RTMs, which are
CIGs at zero subsurface offset). Next, we address the challenges
presented by the complex salt structures in the SEAM model
\citep{fehler2011model}. To overcome these, we recognize the need for
additional guidance from the physics and propose an iterative algorithm,
ASPIRE, that leverages the wave PDE while minimizing the total number of
PDEs at inference time. Finally, we test the robustness and scalability
of our method by applying it to field datasets. The results demonstrate
that the method is adaptable to changes in the test distribution and can
handle large-scale 2D inversion problems (e.g., \(512\times7024\) grid
sizes).

\section{Problem statement}\label{problem-statement}

The core of subsurface characterization involves solving an inverse
problem where the objective is to infer unknown subsurface properties
from observational data. In this context, the forward process,
represented as \begin{equation}\phantomsection\label{eq-forward}{
\mathbf{y}=\mathcal{F}(\mathbf{x}) + \boldsymbol{\varepsilon},
}\end{equation} describes how the observations (shot records)
\(\mathbf{y}\) are generated from the underlying subsurface properties
\(\mathbf{x}\), with \(\boldsymbol{\varepsilon}\) representing
bandwidth-limited noise. Here, we focus on wave-based inversion, where
the forward operator \(\mathcal{F}\) corresponds to the solution of the
wave-equation PDE with the wavefield being restricted to the positions
of the receivers. The complexity of this problem arises from the
null-space present in the forward operator, compounded by noise in the
measurements, which makes direct inversion unreliable as it fails to
characterize the full solution.

To model the noise in the system, one would assume that the noise
\(\boldsymbol{\varepsilon}\) follows a known distribution, such as a
normal distribution \(N(0, \sigma)\). This leads to a likelihood
function \(p(\mathbf{y} \mid \mathbf{x})\), which quantifies the
probability of observing the data \(\mathbf{y}\) given the unknown
parameters \(\mathbf{x}\). If the noise is additive, the likelihood can
be written as a well-known \(\ell_2\)-normed misfit,
\(p(\mathbf{y}|\mathbf{x}) = \frac{1}{2\sigma^2}\|\mathcal{F}(\mathbf{x}) -\mathbf{y} \|_2^2\).
Minimizing data misfits of this form underpins FWI and other variational
methods that use the forward model's fit to the data, but on their own,
it is insufficient to fully resolve the inverse problem due to the
non-uniqueness of solutions and the presence of possibly adverse
parasitic local minima.

In this paper, we address this limitation by adopting a Bayesian
approach to the inverse problem. Our target in this paper is that given
an observation from the field \(\mathbf{y}\) we aim to find samples from
the posterior distribution
\(\mathbf{x}_{\textrm{post}} \sim p(\mathbf{x}|\mathbf{y})\), which can
be defined using Bayes's rule as the combination of the prior and the
data likelihood
\(p(\mathbf{x} \mid\mathbf{y}) \propto p(\mathbf{y}\mid \mathbf{x}) p(\mathbf{x})\).
This posterior distribution combines the prior information about the
subsurface properties, encoded in \(p(\mathbf{x})\), with the likelihood
of observing the data, \(p(\mathbf{y} \mid \mathbf{x})\), resulting in a
probabilistic representation of the subsurface model that accounts for
both the uncertainty in the data and the prior knowledge. This Bayesian
framework allows for a more robust characterization of subsurface
properties, as it provides not just a single estimate but a distribution
of plausible solutions, incorporating uncertainty into the
interpretation.

\section{Related work and our
contributions}\label{related-work-and-our-contributions}

The literature on solving the FWI problem with uncertainty has primarily
focused on implementations of Stein Variational Gradient Descent (SVGD)
methods. The foundational work in this area was introduced by
\citep{zhao2022bayesian}, who applied Bayesian seismic tomography using
normalizing flows. Later, the same authors extended their method to 3D
inversion \citep{zhang20233}, though this approach still required
numerous forward and adjoint PDE solutions for each observation, which
made it computationally expensive.

In an effort to reduce the computational burden of SVGD methods,
\citep{izzatullah2024physics} proposed a technique that minimizes the
number of optimization iterations by carefully defining the prior
distribution. This approach begins with solving the FWI problem under
the assumption of convergence and then adds perturbations to the
solution to create a prior distribution. This prior is used as an
initial ensemble in an SVGD technique to optimize toward individual
maximum likelihood estimates (MLEs) that do not collapse into the same
solution due to a repulsion term in the method. Although the authors did
not use true prior terms---thus these solutions are not precise Bayesian
samples---they presented samples that revealed interesting variations
between them and were mostly focused on the areas of the image with low
illumination from the source-receiver configuration.
\citep{yang2024conditional} extended these concepts by combining ideas
from the Deep Image Prior (DPI) with conditional networks. They
performed an SVGD-like update starting from an ensemble of models that
surrounded a precomputed FWI solution, as in
\citep{izzatullah2024physics}, to fine-tune the weights of a pretrained
conditional network. This pretraining process enabled them to sample
various Earth models by post hoc changes to the network conditions.
While this takes regularization benefits of both the untrained DPI and
the features learned during pretraining, it also does not have a clear
prior used during optimization, so does not correspond to true Bayesian
samples over the target in this case the uncertainty over network
parameters.

In the context of defining realistic Earth priors,
\citep{zhao2024variational} assumed constrained Gaussian distributions
over the prior, which were fitted to real Earth data. They then
implemented a secondary optimization that bypasses the likelihood,
making it efficient to swap different priors at inference time.
\citep{qu2024uncertainty} explored another important aspect of UQ by
employing importance sampling and ensemble methods. Their objective was
to capture uncertainty stemming not only from the ill-posed nature of
the inverse problem but also from epistemic uncertainty in the network
weights.

Similar work using normalizing flows includes \citep{sun2024enabling},
which targets time-lapse inversion. They formed training samples by
performing FWI inversion on prior examples to construct the training
dataset. The prior samples assumed access to a non-cycle-skipped
starting model, to which various perturbations were added, resulting in
a diverse set of training samples. In a related approach,
\citep{sun2024invertible} utilized invertible networks with a maximum
mean discrepancy (MMD)-based loss. Given that MMD estimation requires
large batch sizes, the authors applied model and data reduction
techniques to minimize memory usage and make the training feasible.
Though not specifically aimed at UQ, \citep{muller2023deep} presented a
method similar to our iterative refinement, by using an iterative scheme
to refine the migration velocity models. Targeting our same application
of salt velocity model building, \citep{alali2023deep} combine trained
convolutional neural networks with FWI updates to automatically build
salt velocity models. Here, we aim to deliver automatic salt model
building but without the cost associated with FWI workflows, while also
providing uncertainty quantification.

This paper builds on the methods introduced in previous literature of
applying variational inference methods towards seismic imaging
\citep{siahkoohi2023reliable, orozco2024aspire} that have the goal of
calculating Bayesian samples while making frugal use of the physical
operator by defining an expensive offline training phase to achieve a
cheap online inference phase. In particular, we build on the WISE
framework \citep{yin2024wise} with the use of conditional Diffusion
networks to improve results by leveraging the prior learning
capabilities of Diffusion \citep{ho2020denoising}. While Structural
Similarity Index Metric (SSIM) \citep{brunet2011mathematical} and Root
Mean Squared Error (RMSE) can compare the quality between image
reconstructions, it remains unclear in the literature how to compare two
results for uncertainty. We therefore propose and discuss a battery of
metrics that can be used to benchmark the quality of the uncertainty.
Furthermore, we discuss a method to improve on amortized results with
minimal extra computation at test time and then demonstrate these
methods on field data in preliminary proof of concepts.

Our key contributions include:

\begin{enumerate}
\def\labelenumi{\arabic{enumi}.}
\tightlist
\item
  Extending the WISE framework to incorporate conditional Diffusion
  networks.
\item
  Proposing four benchmark metrics to assess the quality of UQ.
\item
  Testing our method on the Compass and \texttt{Synthoseis} training
  datasets, demonstrating the value of using Common Image Gathers (CIGs)
  over Reverse Time Migrations (RTMs) alone from the standpoint of image
  quality and also uncertainty quality.
\item
  Introducing the iterative ASPIRE algorithm for seismic inversion,
  particularly for complex salt structures, such as those in the SEAM
  model.
\item
  Presenting a proof-of-concept application on field datasets to
  demonstrate the scalability of the method.
\end{enumerate}

\section{Methods}\label{methods}

Our approach builds on the simulation-based inference (SBI) framework
\citep{cranmer2020frontier}, a powerful tool for solving inverse
problems that leverages numerical simulations and conditional generative
networks to approximate posterior distributions. While SBI is, in
principle, a general method, directly applying it to the complex problem
of seismic subsurface characterization presents significant challenges
due to the intricacies of waveform data. To address these challenges, we
incorporate conditional Diffusion networks as the generative backbone,
enabling the learning of an expressive prior and scalable posterior
sampling. Additionally, we employ physics-based summary statistics to
compress observational data while preserving crucial information about
the subsurface properties.

\subsection{Simulation-based
inference}\label{simulation-based-inference}

SBI combines the strengths of numerical simulations and conditional
generative modeling, providing a powerful framework for solving complex
inverse problems \citep{cranmer2020frontier}. Numerical simulations are
used to generate data pairs,
\(\mathcal{D}=\{\mathbf{x}^{i},\mathbf{y}^{i})\}_{i=0}^{N}\), where each
pair consists of a set of subsurface properties \(\mathbf{x}^{i}\) and
the corresponding simulated observation \(\mathbf{y}^{i}\) derived using
the forward simulation Equation~\ref{eq-forward}. These data pairs are
then used to train a conditional generative network which learns the
posterior distribution of the unknown properties given the observations.
By integrating physics-based simulations with modern generative modeling
techniques, SBI provides an amortized method---that is, it generalizes
over all data simulated from the realizations of the prior. This means
that after an initial training phase, inference is inexpensive and can
be done on many unseen test observations without retraining or expensive
applications of the forward/adjoint operator.

\subsection{Conditional generative modeling with Diffusion
networks}\label{conditional-generative-modeling-with-diffusion-networks}

Diffusion networks are density estimation algorithms based on learning
the score of the target distribution
\(\nabla_{\mathbf{x}}\log p(\mathbf{x})\) \citep{song2020score}.
Specifically, we define a family of mollified distributions
\(\nabla_{\mathbf{x}}\log p(\mathbf{x},\sigma(t))\), where \(\sigma(t)\)
represents a noise schedule such that, as time \(t\) increases, the
distribution is mollified towards the Gaussian distribution. After
learning the score for all time steps \(t\), Diffusion networks can
evaluate likelihoods \citep{zheng2023improved} and sample new generative
instances from the target distribution \(p(\mathbf{x})\). To learn the
score, Diffusion networks use one of the many forms of the denoising
objective \citep{karras2022elucidating},

\[
\hat \theta = \underset{\boldsymbol{\theta}}{\operatorname{arg min}} \, \,{\mathbb{E}}_{\mathbf{x}\sim p(\mathbf{x})}{\mathbb{E}}_{\mathbf{n}\sim \mathcal{N}(0,t^2 I)}\| s_{\boldsymbol{\theta}}(\mathbf{{x}} + \mathbf{n};t)-\mathbf{x} \|^2_2
\]

where the approximation the score at time \(t\) is given by the
evaluation of the trained network \(s_{\hat \theta}(\mathbf{x},t)\). New
samples from the target distribution can be generated by solving the
following stochastic differential equation (SDE):

\begin{equation}\phantomsection\label{eq-sampling}{
d\mathbf{x} = -\dot\sigma(t)\sigma(t) \nabla_\mathbf{x} p(\mathbf{x};\sigma(t))dt.
}\end{equation}

Here, we use the formulation in \citet{karras2022elucidating} to
simplify terms. Many strategies exist for solving this SDE. Here, we
follow the method in \citet{karras2022elucidating} and use a procedure
similar to the predictor-corrector sampler of \citet{song2020score}.

To extend Diffusion networks towards conditional distributions, we aim
to find the conditional score
\(\nabla_{\mathbf{x}} \log p(\mathbf{x}|\mathbf{y})\). Based on the
theory laid out in \citet{batzolis2021conditional} and
\citet{baldassari2024conditional}, we can approximate this conditional
score with a simple modification of Diffusion networks by incorporating
the observation \(\mathbf{y}\) as an additional condition to the
denoising network

\[
\hat \theta = \underset{\boldsymbol{\theta}}{\operatorname{arg min}} \, \,{\mathbb{E}}_{\mathbf{y}\sim p(\mathbf{y}\mid\mathbf{x})} {\mathbb{E}}_{\mathbf{x}\sim p(\mathbf{x})}{\mathbb{E}}_{\mathbf{n}\sim \mathcal{N}(0,t^2 I)}\| s_{\boldsymbol{\theta}}(\mathbf{{x}} +\mathbf{n},\mathbf{y},t)-\mathbf{x} \|^2_2.
\]

In many cases, implementing a conditional Diffusion network involves
simple modifications to existing non-conditional Diffusion networks. In
\href{https://github.com/slimgroup/GeneralizedDiffusion}{Github Repo},
we share our conditional Diffusion implementation derived from
non-conditional networks from Elucidating Diffusion Models
\citep{karras2022elucidating}. In practice, we approximate the
expectation over prior samples with a set of ground truth examples and
sample from the likelihood by running the simulator in
Equation~\ref{eq-forward} to form a paired dataset
\(\mathcal{D}=\{(\mathbf{x}^{i},\mathbf{y}^{i}\}_{i=0}^{N}\), then
perform stochastic gradient descent on the following objective:
\begin{equation}\phantomsection\label{eq-cond-training}{
\hat \theta = \underset{\boldsymbol{\theta}}{\operatorname{arg min}}  \, \, \mathcal{L}(\mathcal{D},\theta) = \underset{\boldsymbol{\theta}}{\operatorname{arg min}} \, \, \sum_{i=0}^{N} {\mathbb{E}}_{\mathbf{n}\sim \mathcal{N}(0,t^2 I)}\| s_{\boldsymbol{\theta}}(\mathbf{{x}}^{i} +\mathbf{n},\mathbf{y}^{i},t)-\mathbf{x}^{i} \|^2_2.
}\end{equation}

After training, conditional Diffusion generates samples from the
posterior \(p(\mathbf{x}\mid\mathbf{y}^{\textrm{obs}})\) by conditioning
the network on \(\mathbf{y}^\textrm{obs}\) and following the reverse
process on the same SDE as in Equation~\ref{eq-sampling} to generate a
posterior sample \(\mathbf{x}_{\textrm{post}}\). While the physics of
the inverse problem is present in the formation of the dataset, this is
insufficient in the case of complex forward operators associated with
the wave equation \citep{orozco2023adjoint}. Therefore, we will explore
a method to alleviate this challenge next by incorporating more physics
into the methodology.

\subsection{Physics-based summary
statistics}\label{physics-based-summary-statistics}

While simulation-based inference (SBI) has shown promise across various
fields, applying it to problems with complex forward operators, such as
seismic inversion, presents unique challenges. These challenges stem
from the difficulty of extracting the rich information contained in
waveforms. To address this, we employ summary statistics that compress
the data while preserving information needed to infer unknown subsurface
properties. Summary statistics can either be hand-crafted by domain
experts or learned by exploiting probabilistic symmetries within the
data \citep{bloem2020probabilistic}. In this work, we follow the methods
of \citet{orozco2023adjoint} and \citet{yin2024wise} by using
physics-based summary statistics derived from the physical forward
operator. Specifically, we utilize the gradient of the data likelihood,
which allows us to inject physics into the generative process at the
cost of a single PDE solve per set of posterior samples, where the PDE
corresponds to the computations required for a single migration.

Using this gradient serves as a natural way to undo the complexity of
the waveforms since it brings the data into the model space.
Theoretically, this approach is well-founded, as it optimally saturates
the information inequality \citep{alsing2018generalized}, a bound that
limits how much information a summary statistic can extract about an
unknown variable. Empirical results have shown that incorporating these
summary statistics accelerates training and improves convergence with
fewer data samples \citep{orozco2023adjoint}. Given an observation
\(\mathbf{y}\) and a migration-velocity model \(\mathbf{x}_0\), the
summary statistic is calculated as follows:
\begin{equation}\phantomsection\label{eq-score}{
\mathbf{\overline y} := h(\mathbf{y}) = \overline{\nabla \mathcal{F}}(\mathbf{x}_0)^{\top}(  \mathcal{F}(\mathbf{x}_0) - \mathbf{y}),
}\end{equation}

where \(\overline{\nabla \mathcal{F}}\) is the subsurface-offset
extended Jacobian, following the method of
\citep{hou2016accelerating, yin2024wise}. \(\mathbf{\overline y}\)
contains the CIGs, which represent the summarized waveform but
transformed to model space, while preserving more information compared
to RTMs in situations where the migration-velocity model is poor. We
summarize all training data
\(\overline{\mathbf{y}}^{i}=h({\mathbf{y}}^{i})\) and form a new
training dataset
\(\overline{\mathcal{D}} = \{\mathbf{x}^{i},\overline{\mathbf{y}}^{i}\}_{i=0}^{N}\)
and train a conditional Diffusion network with
Equation~\ref{eq-cond-training}. At inference time, the observation
\(\mathbf{y}^{\mathrm{obs}}\) is summarized in the same manner,
\(\mathbf{\overline y}^{\mathrm{obs}} = h(\mathbf{y}^{\mathrm{obs}})\)
and then passed into the same sampling process as
Equation~\ref{eq-sampling}. The computational cost of this approach is
low because it requires only a single extended migration per observation
at inference time, making it more computationally efficient than
non-amortized methods that typically demand hundreds of PDE solves
\citep{zhao2022bayesian} for each dataset. Our approach is also
amortized and therefore can be applied to many unseen observations
without repeating the costly training process.

\section{Stylized examples}\label{stylized-examples}

We separate our experiments into two parts. First, we train our
generative networks on synthetic seismic model sets and validate them on
unseen datasets from the same synthetic distribution as used during
training. Secondly, we apply these trained networks to field datasets to
assess their robustness in practical scenarios and also to understand
the prior that they learn.

\subsection{Training dataset
generation}\label{training-dataset-generation}

For the synthetic experiments, the migration-velocity models used are
intentionally not kinematically accurate. For the two salt imaging
experiments, we first remove the salt body from the model, replace it
with the average velocity of the sedimentary layers, and then smooth the
resulting model using a Gaussian kernel. To minimize artifacts from
tomographic updates, the migrations are generated using the
inverse-scattering imaging condition (ISIC) \citep{witte2017sparsity},
which implies that our method is purely reflection-based. The sources
are given by Ricker wavelets that are filtered to remove unrealistically
low frequencies below 3 Hz. When noise is added to the simulated shot
gathers, we apply a filter to ensure the noise contains the same
frequency content as the source wavelet. We use JUDI and Devito for wave
simulations {[}\citet{witte2017sparsity}; \citet{louboutin2019devito};
\citet{luporini2020architecture};{]}. The extended Jacobian operator was
created using \citep{louboutin2024ImageGather} with a subsurface range
that was chosen to contain the majority of the non-zero offset energy.

Across all experiments, we create between \(700–800\) training pairs.
The Diffusion networks are trained until the image quality metrics
(RMSE, SSIM) for the posterior means stop improving on a validation set
held out during training. To calculate posterior statistics, we generate
\(N\) posterior samples and calculate Monte Carlo estimations of the
posterior mean,
\(\mathbf{x}_{\textrm{mean}} = \sum_{i=0}^{N} \mathbf{x}_{\textrm{post}}^{i} / N\),
and standard deviation,
\(\mathbf{x}_{\textrm{std}} = \sqrt{\sum_{i=0}^{N} (\mathbf{x}_{\textrm{post}}^{i} - \mathbf{x}_{\textrm{mean}})^2 / N}\).
The number of posterior samples to use is decided by increasing the
number until the posterior standard deviation converges; in our case,
\(N=64\).

\subsection{Compass model}\label{compass-model}

\begin{figure}

\begin{minipage}{0.50\linewidth}

\centering{

\includegraphics[width=1\textwidth,height=\textheight]{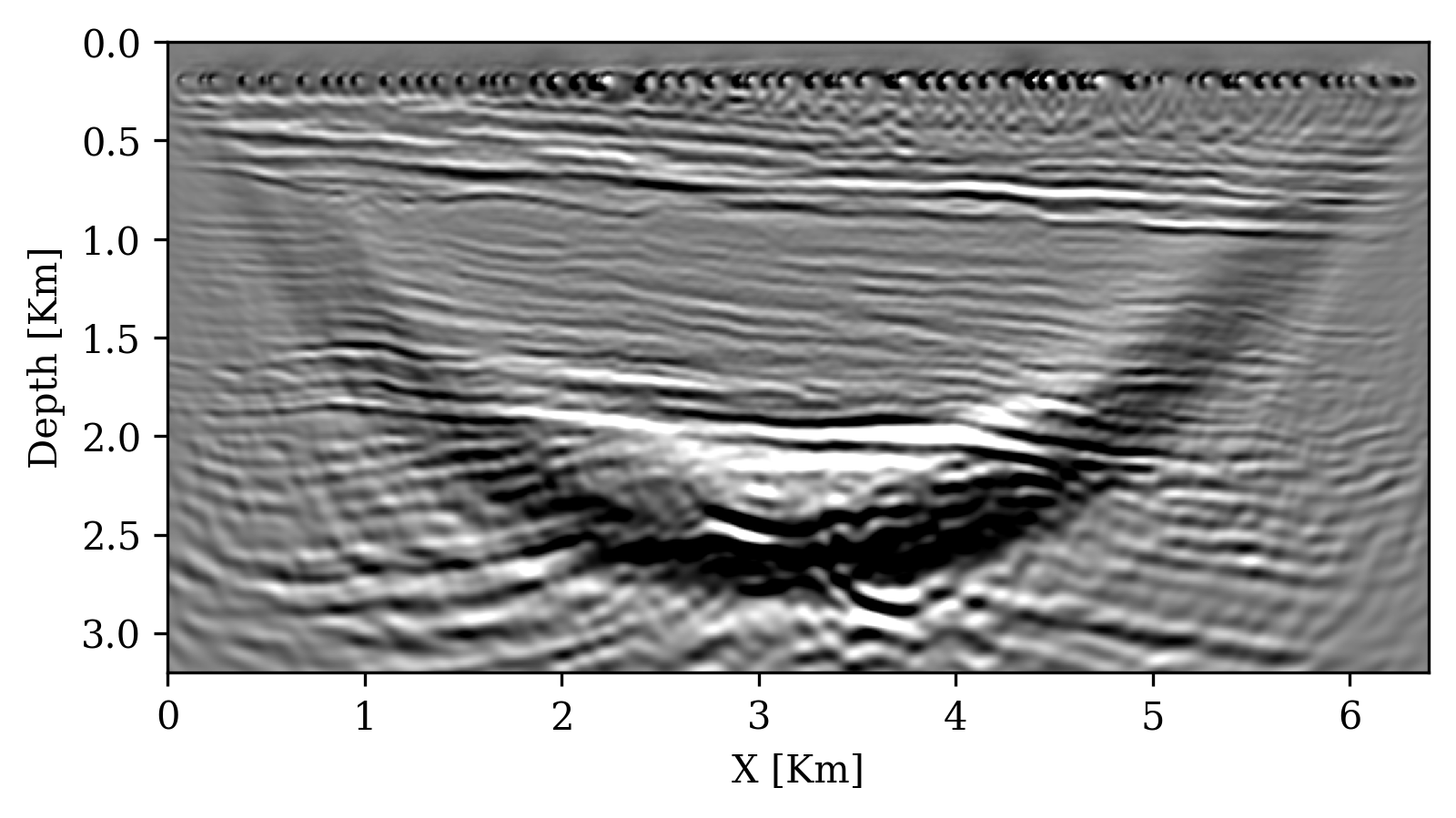}

}

\subcaption{\label{fig-compass-a}Reverse-time migration}

\end{minipage}%
\begin{minipage}{0.50\linewidth}

\centering{

\includegraphics[width=1\textwidth,height=\textheight]{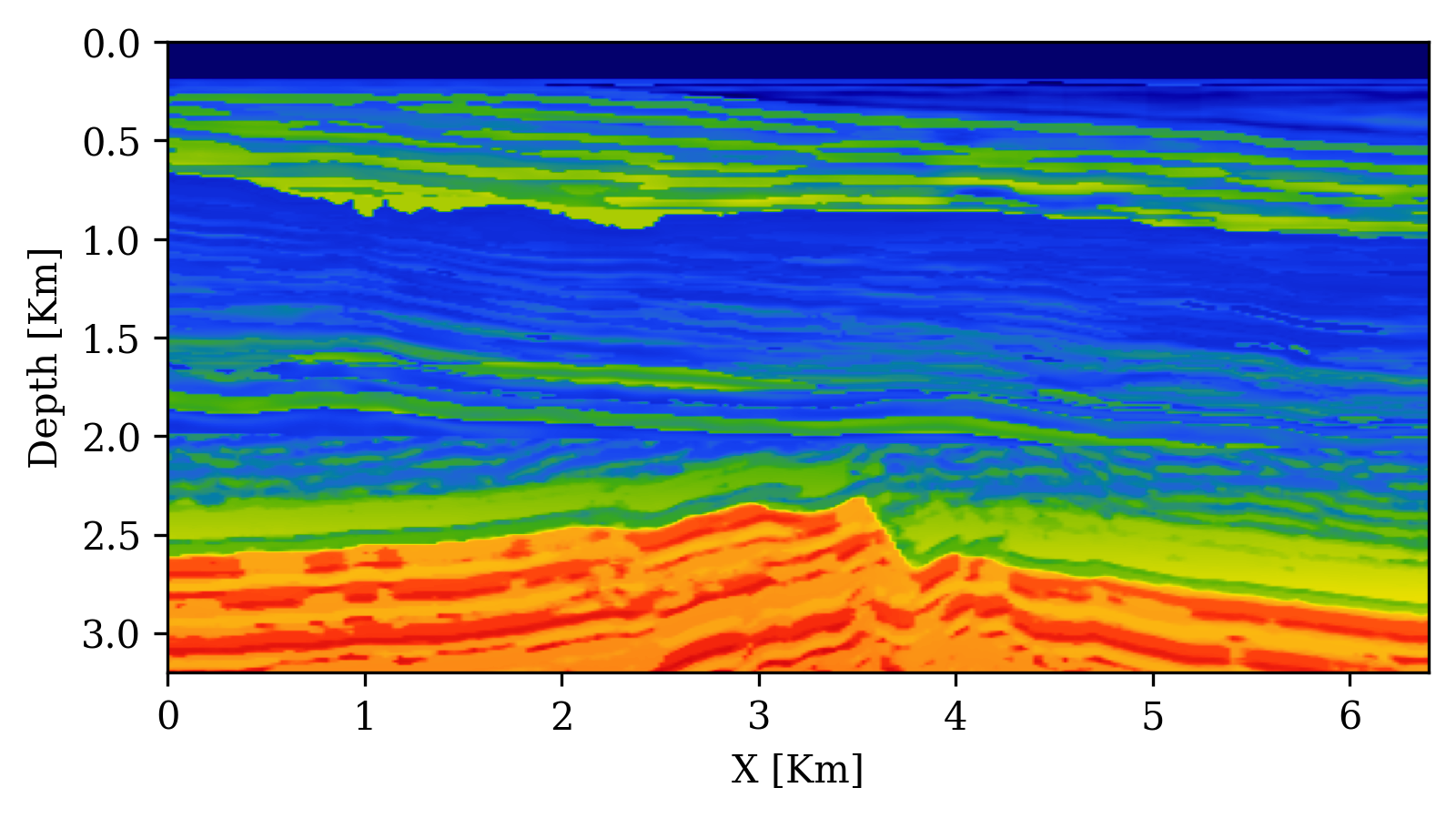}

}

\subcaption{\label{fig-compass-b}Ground truth velocity model}

\end{minipage}%
\newline
\begin{minipage}{0.50\linewidth}

\centering{

\includegraphics[width=1\textwidth,height=\textheight]{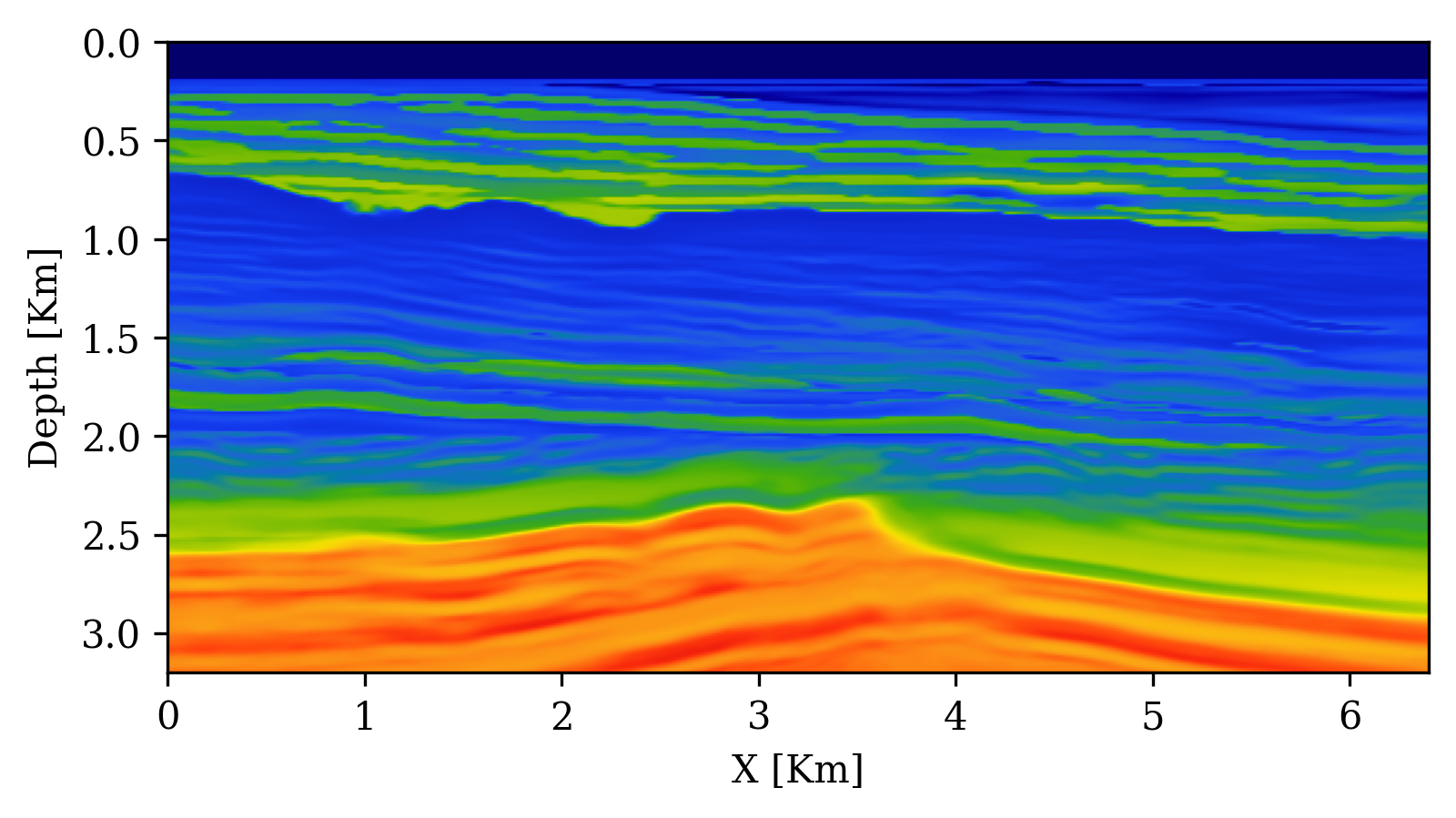}

}

\subcaption{\label{fig-compass-c}Posterior mean w/ RTMs SSIM=\(0.78\)}

\end{minipage}%
\begin{minipage}{0.50\linewidth}

\centering{

\includegraphics[width=1\textwidth,height=\textheight]{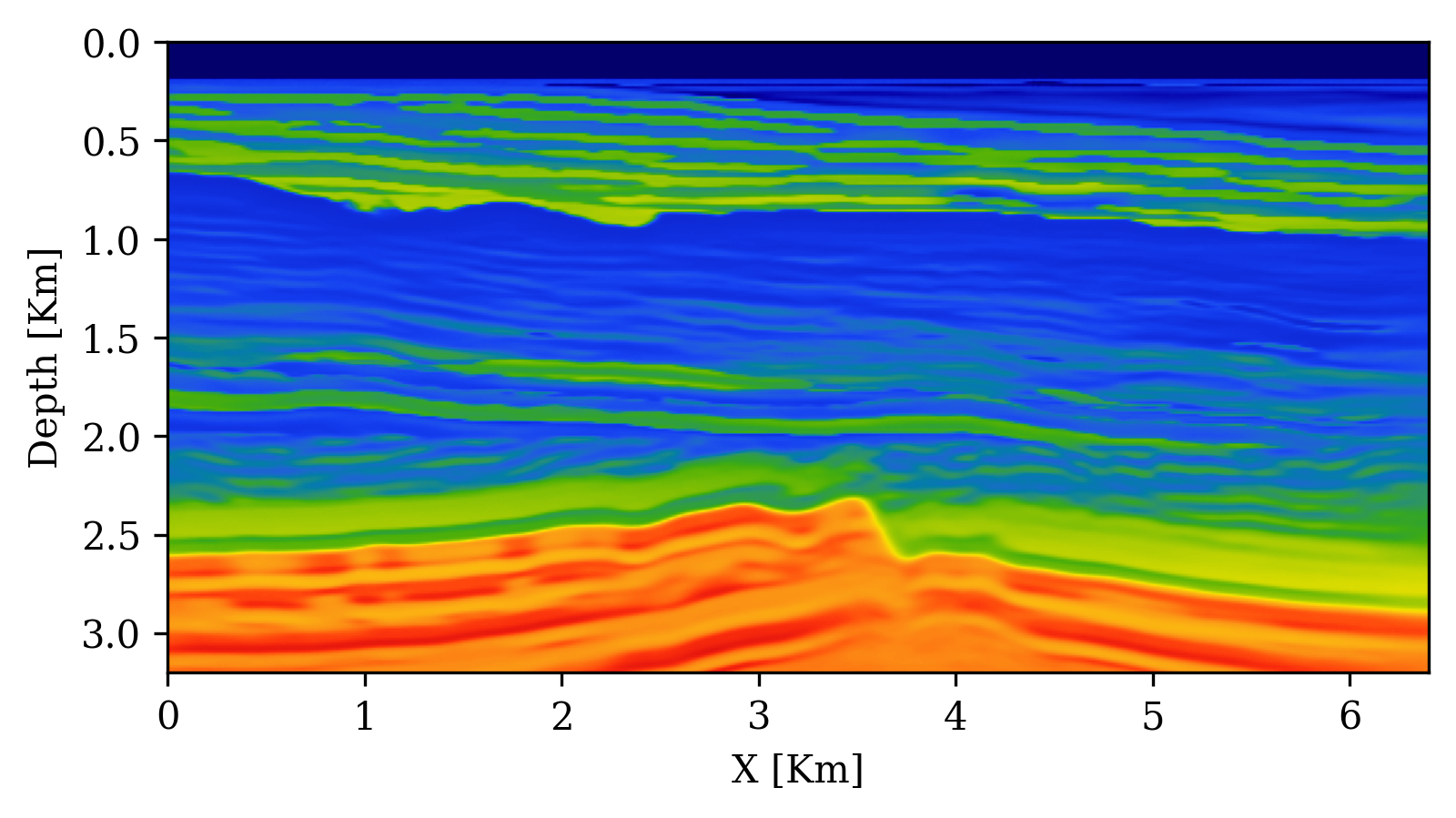}

}

\subcaption{\label{fig-compass-d}Posterior mean w/ CIGs SSIM=\(0.84\)}

\end{minipage}%
\newline
\begin{minipage}{0.50\linewidth}

\centering{

\includegraphics[width=1\textwidth,height=\textheight]{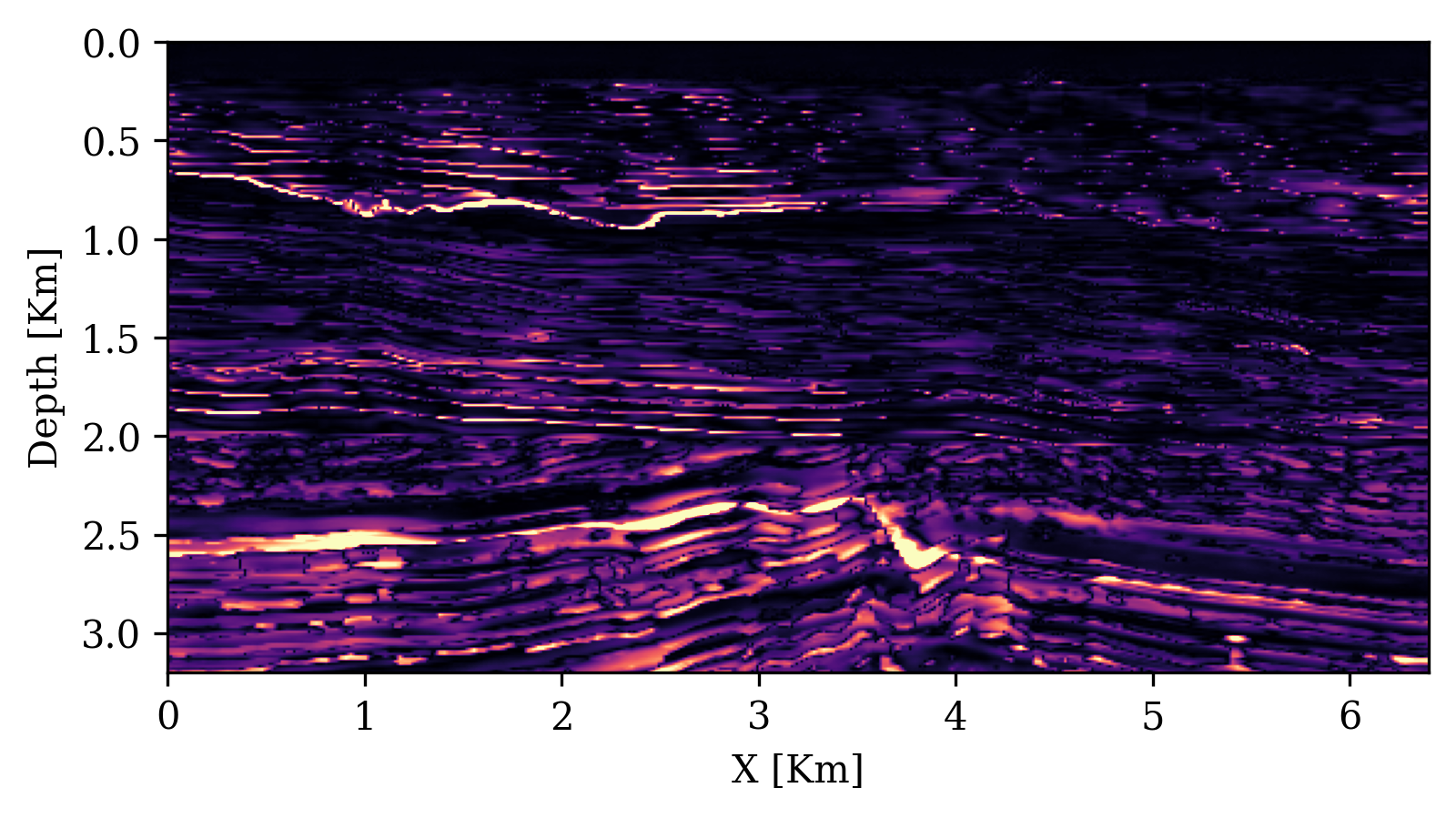}

}

\subcaption{\label{fig-compass-e}Error w/ RTMs RMSE=\(0.13\)}

\end{minipage}%
\begin{minipage}{0.50\linewidth}

\centering{

\includegraphics[width=1\textwidth,height=\textheight]{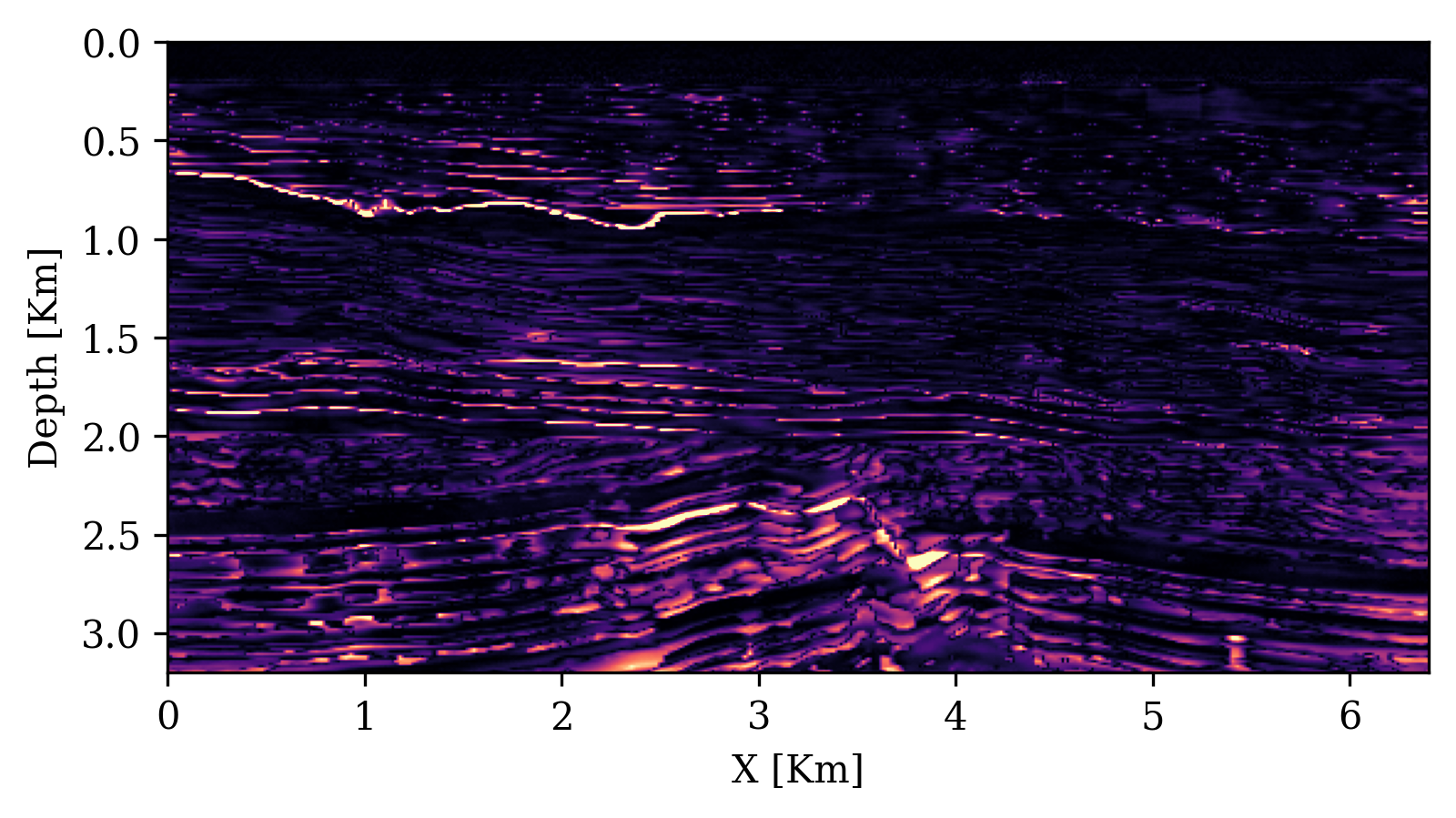}

}

\subcaption{\label{fig-compass-f}Error w/ CIGs RMSE=\(0.10\)}

\end{minipage}%
\newline
\begin{minipage}{0.50\linewidth}

\centering{

\includegraphics[width=1\textwidth,height=\textheight]{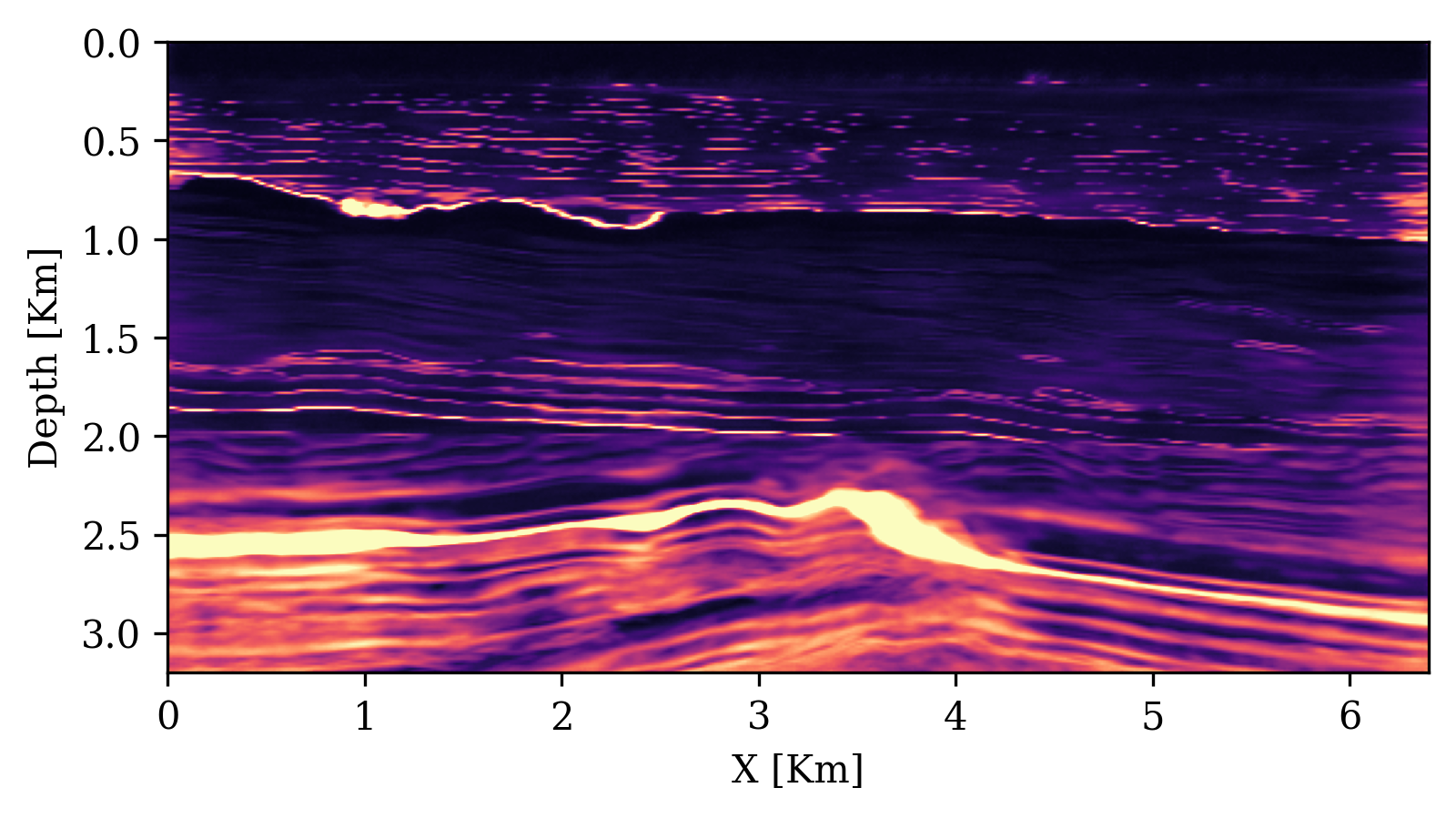}

}

\subcaption{\label{fig-compass-g}Posterior standard deviation w/ RTMs}

\end{minipage}%
\begin{minipage}{0.50\linewidth}

\centering{

\includegraphics[width=1\textwidth,height=\textheight]{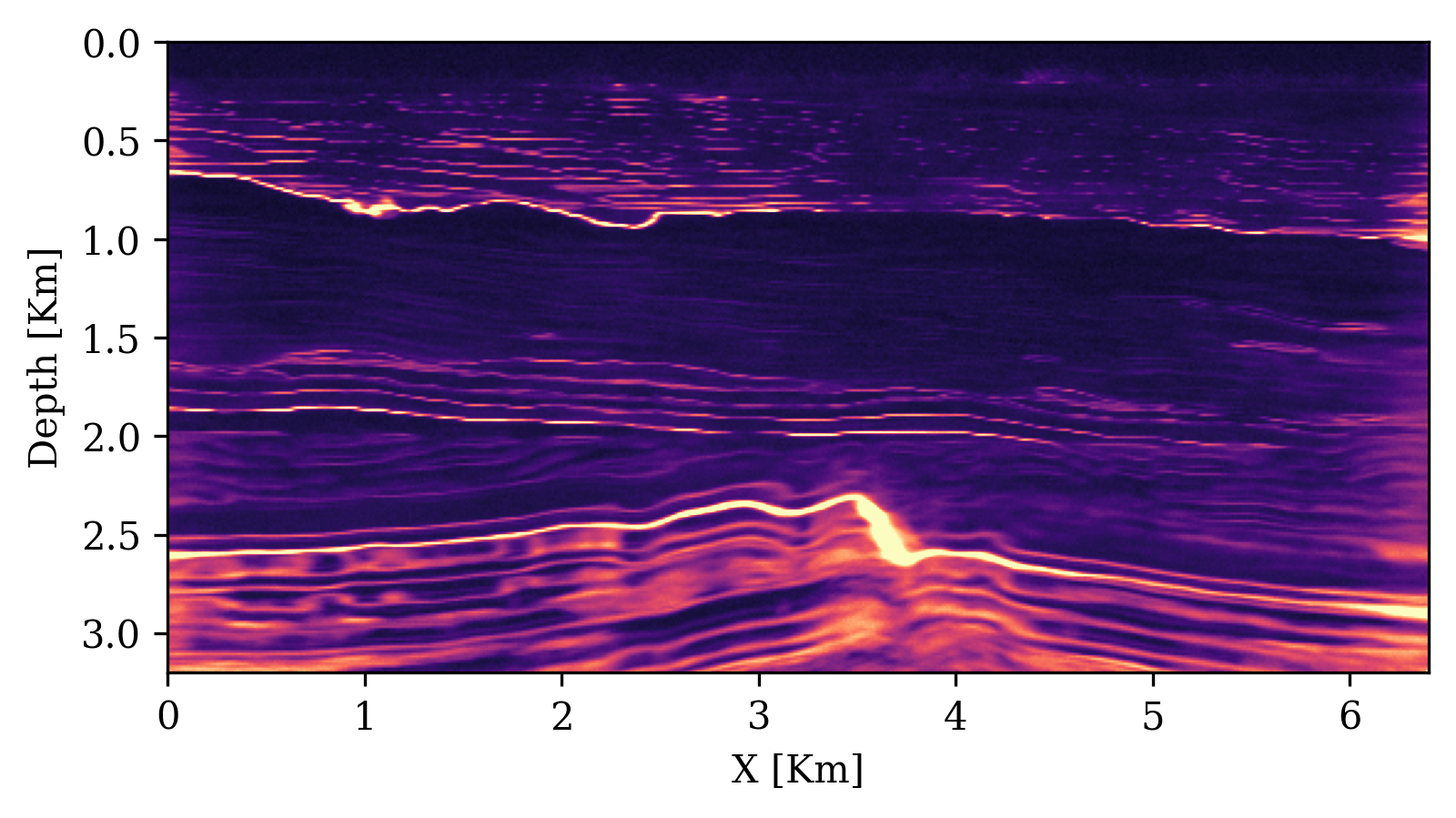}

}

\subcaption{\label{fig-compass-h}Posterior standard deviation w/ CIGs}

\end{minipage}%

\caption{\label{fig-compass-test}Posterior sampling on Compass dataset.
We observe an increase in quality of the inferred velocity model when
using CIGs instead of RTMs.}

\end{figure}

Our evaluation begins by using the Compass model
\citep{jones2012building}. The training dataset consists of the same
velocity and CIGs pairs as in WISE \citep{yin2024wise}. These include
\(N=800\) training pairs of velocity models of size (\(512\times256\))
that are discretized with \(12.5\) m and CIGs that are generated with
\(50\) equally spaced offsets between \(-500\) m to \(500\) m without
ISIC. The migration-velocity models for the Compass dataset are created
using a single 1D velocity profile that is derived from the training
dataset. Our only modification is that we use conditional Diffusion as
the generative network instead of conditional normalizing flows.
Training took \(14\) GPU hours, while posterior sampling takes \(440\)
sec for the \(64\)-shot CIGs in addition to \(2\) sec for each posterior
sample. The results summarized in Figure~\ref{fig-compass-test} confirm
the observation from \citet{yin2024wise} that CIGs drastically improve
posterior sampling performance. As we discuss later, these results
improve on the normalizing flow results from \citet{yin2024wise}.
Overall, we observe increasing error and uncertainty with depth.
Additionally, there is correlation between the complexity of the
velocity model and the error and uncertainty. Aside from uncertainty
near the strong lateral variation along the major unconformity in the
model, we also observe increased errors and uncertainty associated with
topology on the velocity kickback that occurs around \(750\) m depth on
the left side of the model. In the section on uncertainty benchmarks, we
expand this rudimentary analysis and detail how we can determine which
uncertainty is of higher quality according to well-defined metrics.

\subsection{Synthoseis models}\label{synthoseis-models}

While the results presented in the previous section are encouraging, the
geological setting in the North Sea Compass dataset lacks sufficient
complexity to put our inference methodology to the test. In addition,
its training was limited to neighboring 2D slices from the same area and
therefore may lack in diversity. To address this potential lack of
diversity and complexity, we consider an example where we train on
synthetically generated models produced by the open-source software
package
\texttt{Synthoseis}{[}https://github.com/sede-open/synthoseis{]}.
\texttt{Synthoseis} is an algorithm designed to generate realistic and
diverse synthetic 3D seismic models tailored for deep learning
applications \citep{merrifield2022synthetic}. Our approach specifically
uses the algorithm's velocity-model generation routines, focusing on the
creation of the acoustic velocity property (\(V_p\)) of Earth models.
The workflow consists of several key steps, namely initialization of
random Earth parameters ranges, generating 3D depth horizons, embedding
of rock property models, and ultimately producing 3D volumes. In this
study, we examine scenarios with Earth models that feature salt bodies.
Although each model produced by the algorithm exhibits full 3D
structures, for our experiments, we extracted 2D slices from the 3D
models to manage computational resources needed during
\texttt{Synthoseis} generation. It is important to note that while these
models are not created by the same generative model that we train, thus
using these samples does not fall under the regime of autophagy as
defined in \citep{alemohammad2023self}.

To create the training dataset \(\overline{\mathcal{D}}\), we generate
\(N=800\) pairs. Each \(\mathbf{x}\) has a grid size of
\(512 \times 256\) with a spatial discretization of \(12.5\) m. We
modeled towed sources at the ocean surface with an interval of \(12.5\)
m. The receivers are placed on the ocean bottom with an average sampling
of \(200\) m, they are positioned using a random jittered sampling
scheme \citep{hennenfent2008simply}. The shot records \(\mathbf{y}\) are
simulated by recording for \(3.2\) s, while the active sources are
modeled with a Ricker wavelet with central frequency of \(20\) Hz. The
simulated noise was band-limited to contain the same frequency content
as the source and had a magnitude of \(25\) dB. To make
migration-velocity models, we first replace values where salt is located
with the average value of the sedimentary layers. Then, we convert the
model from depth to time, smooth in time with an anisotropic Gaussian
kernel of size \((40\times80)\), and then convert it back to depth. We
follow this nonlinear procedure to intentionally create poor
migration-velocity models. The horizontal subsurface-offset migrations
\(\overline{\mathbf{y}}\) include \(24\) equally spaced offsets between
-500 m and 500 m. The conditional Diffusion network was trained for
\(12\) GPU hours. For inference, the algorithm takes \(277\) sec to
generate CIGs for \(32\) shots and \(2\) sec for each posterior sample.

\begin{figure}

\begin{minipage}{0.50\linewidth}

\centering{

\includegraphics[width=1\textwidth,height=\textheight]{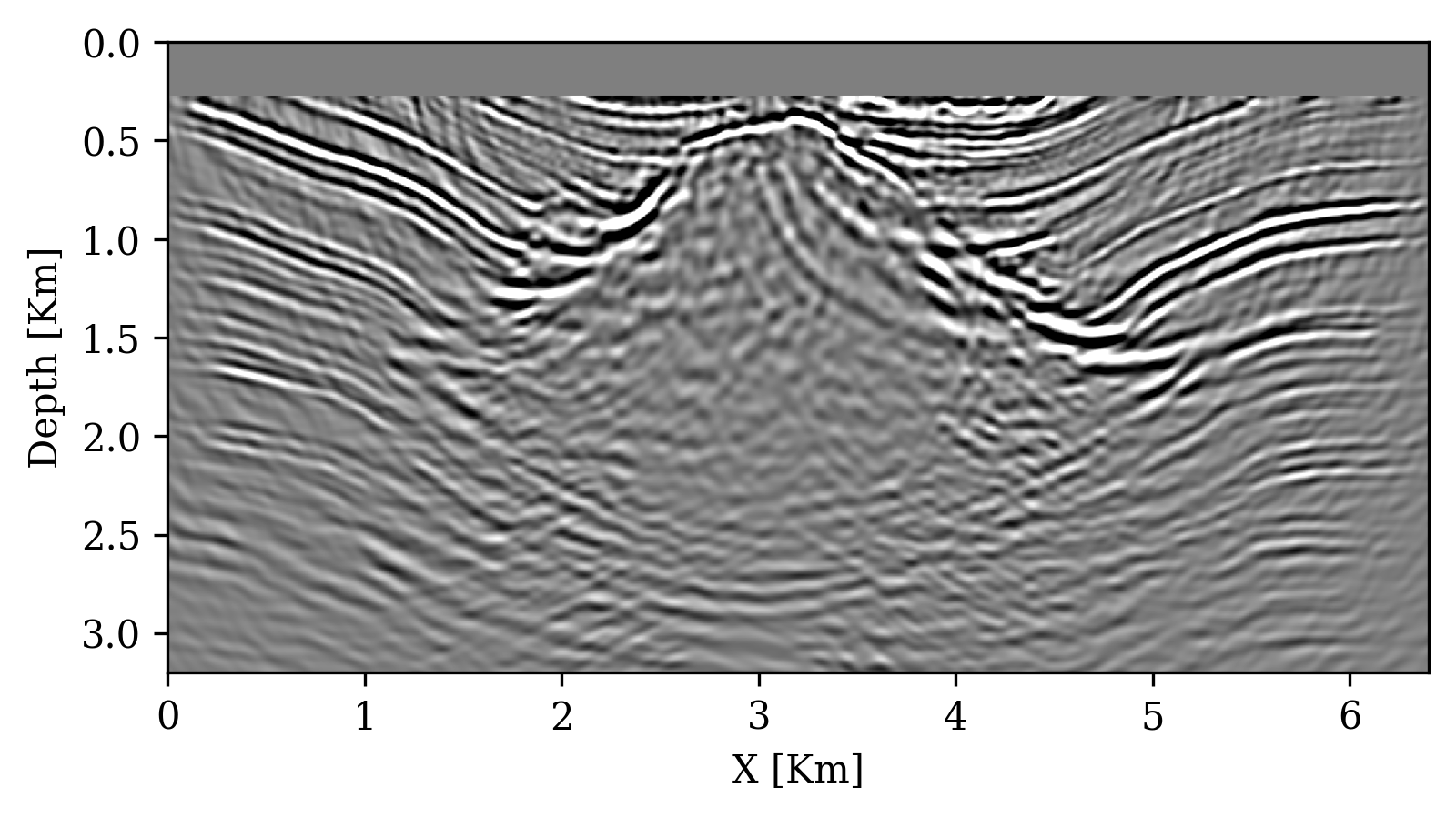}

}

\subcaption{\label{fig-synth-a}Reverse-time migration}

\end{minipage}%
\begin{minipage}{0.50\linewidth}

\centering{

\includegraphics[width=1\textwidth,height=\textheight]{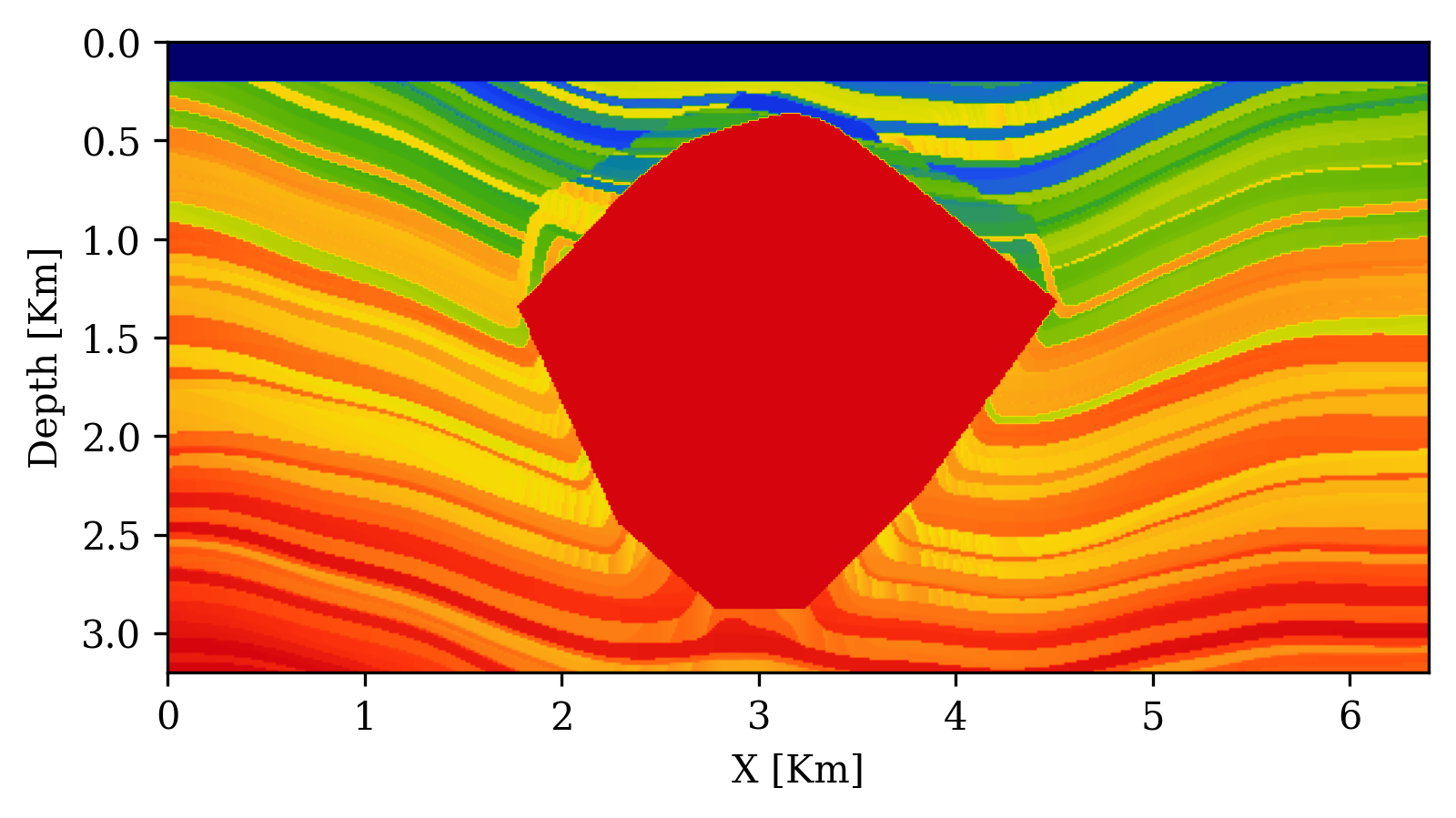}

}

\subcaption{\label{fig-synth-b}Ground truth velocity model}

\end{minipage}%
\newline
\begin{minipage}{0.50\linewidth}

\centering{

\includegraphics[width=1\textwidth,height=\textheight]{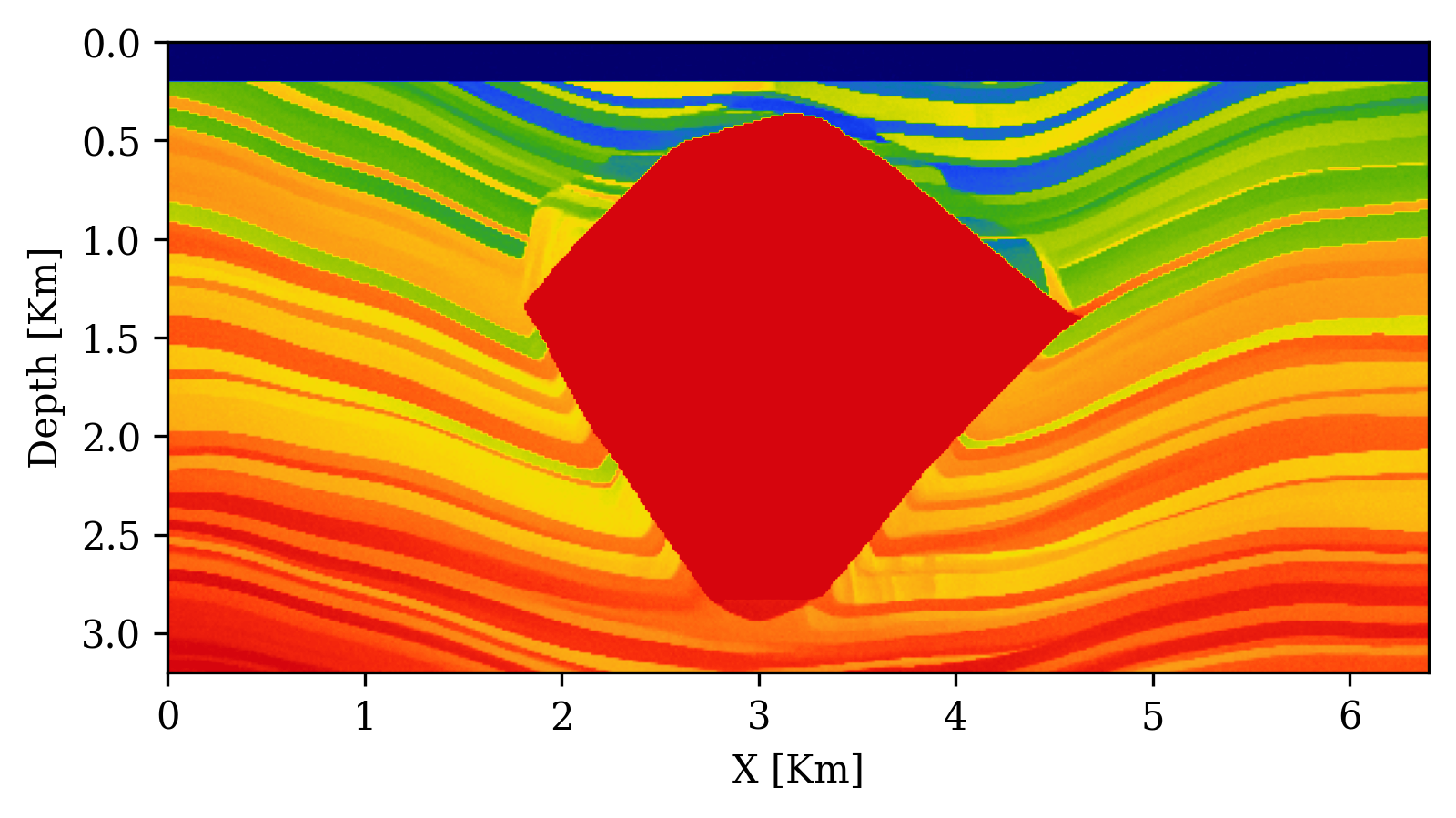}

}

\subcaption{\label{fig-synth-c}Posterior Sample w/ RTMs}

\end{minipage}%
\begin{minipage}{0.50\linewidth}

\centering{

\includegraphics[width=1\textwidth,height=\textheight]{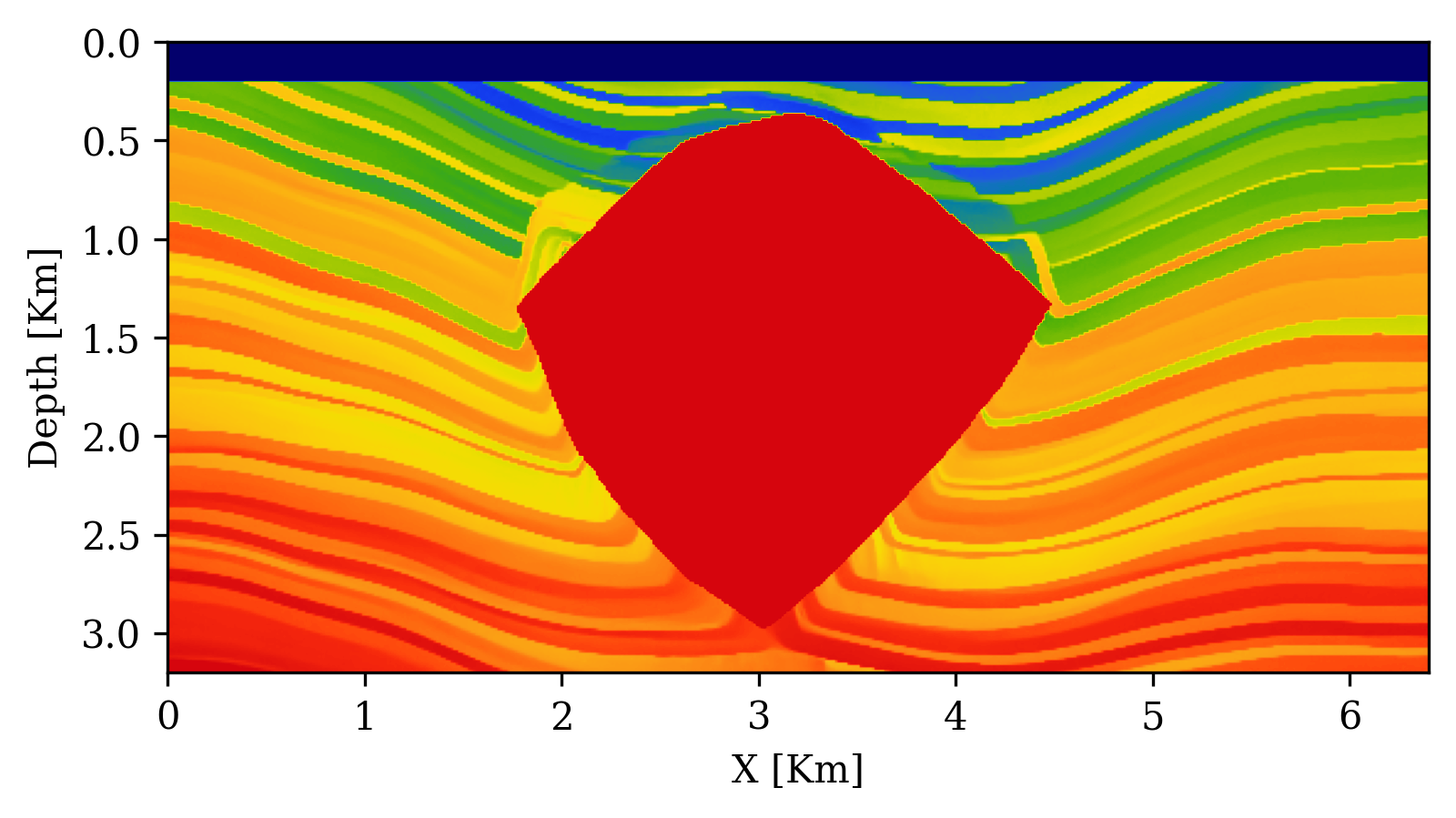}

}

\subcaption{\label{fig-synth-d}Posterior Sample w/ CIGs}

\end{minipage}%
\newline
\begin{minipage}{0.50\linewidth}

\centering{

\includegraphics[width=1\textwidth,height=\textheight]{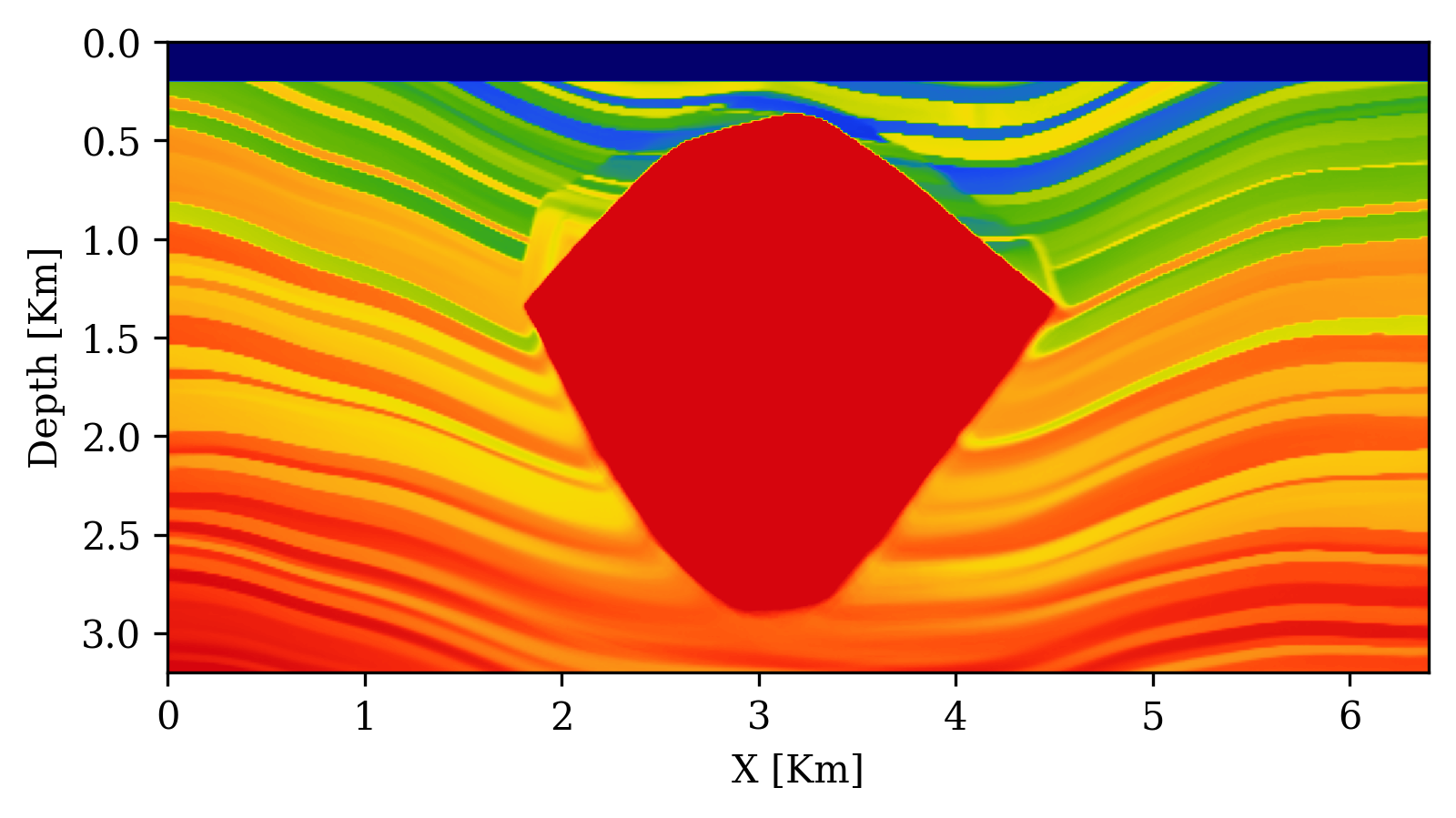}

}

\subcaption{\label{fig-synth-e}Posterior mean w/ RTMs SSIM=\(0.84\)}

\end{minipage}%
\begin{minipage}{0.50\linewidth}

\centering{

\includegraphics[width=1\textwidth,height=\textheight]{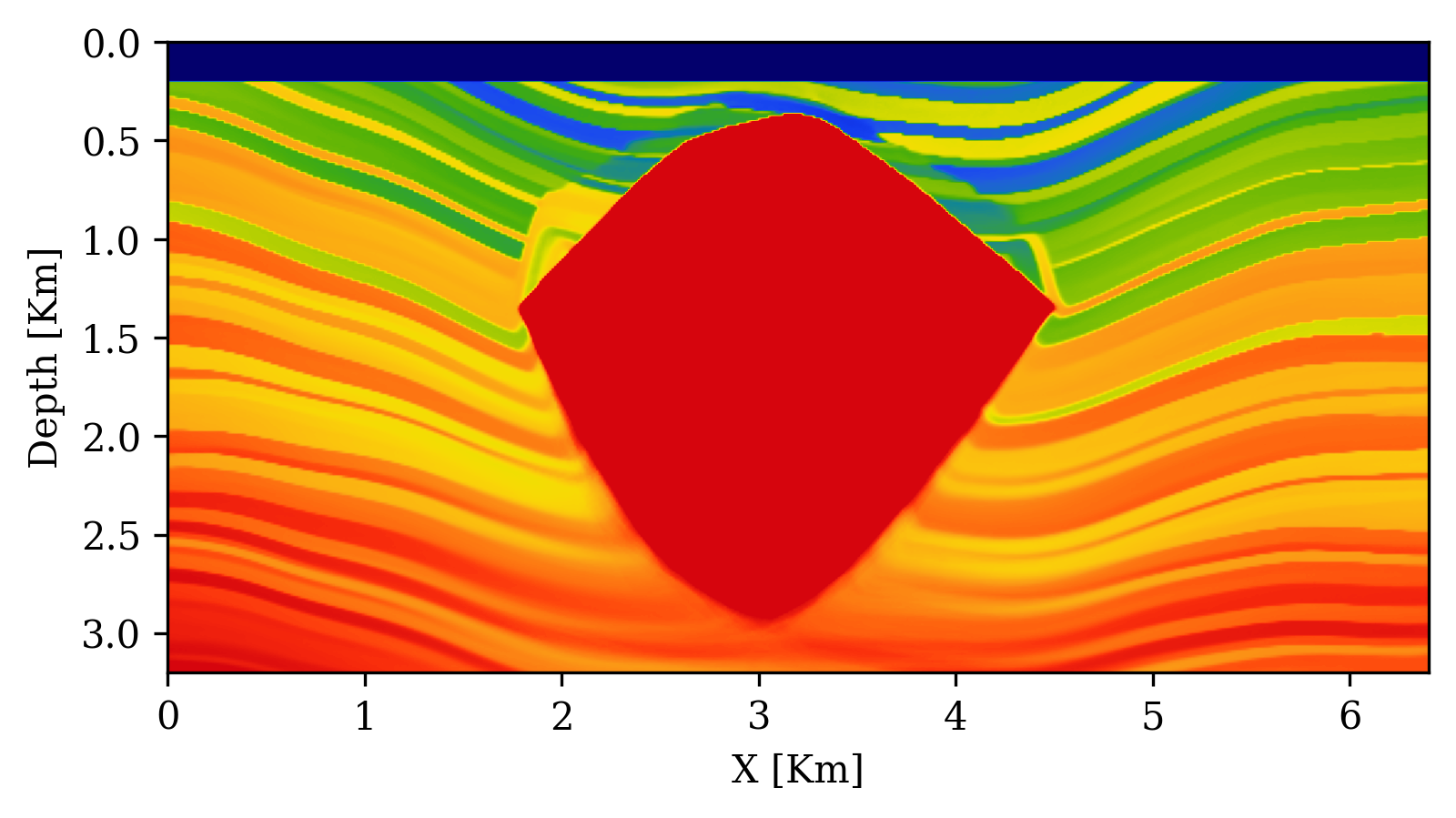}

}

\subcaption{\label{fig-synth-f}Posterior mean w/ CIGs SSIM=\(0.89\)}

\end{minipage}%
\newline
\begin{minipage}{0.50\linewidth}

\centering{

\includegraphics[width=1\textwidth,height=\textheight]{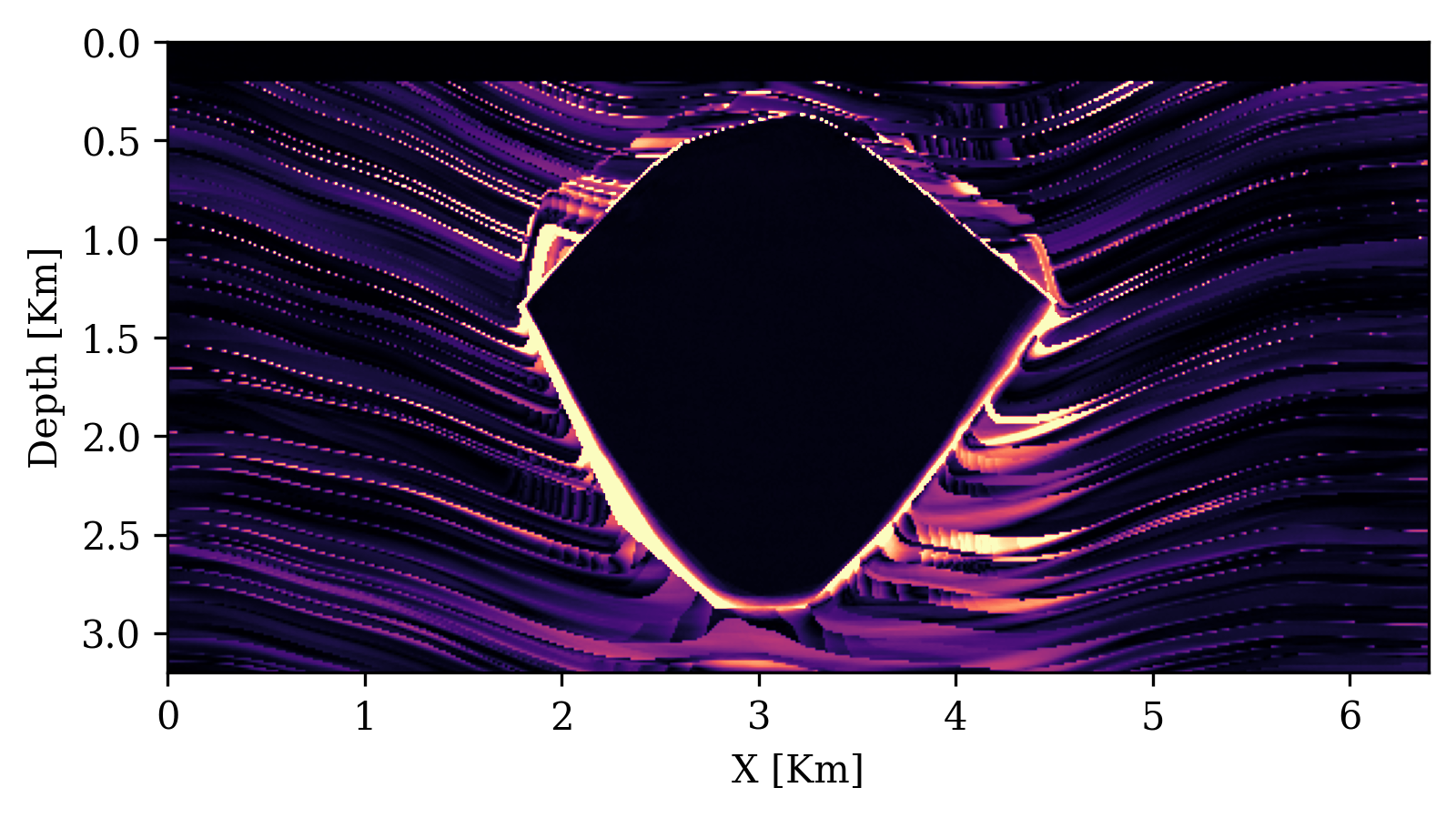}

}

\subcaption{\label{fig-synth-g}Error w/ RTMs RMSE=\(0.15\)}

\end{minipage}%
\begin{minipage}{0.50\linewidth}

\centering{

\includegraphics[width=1\textwidth,height=\textheight]{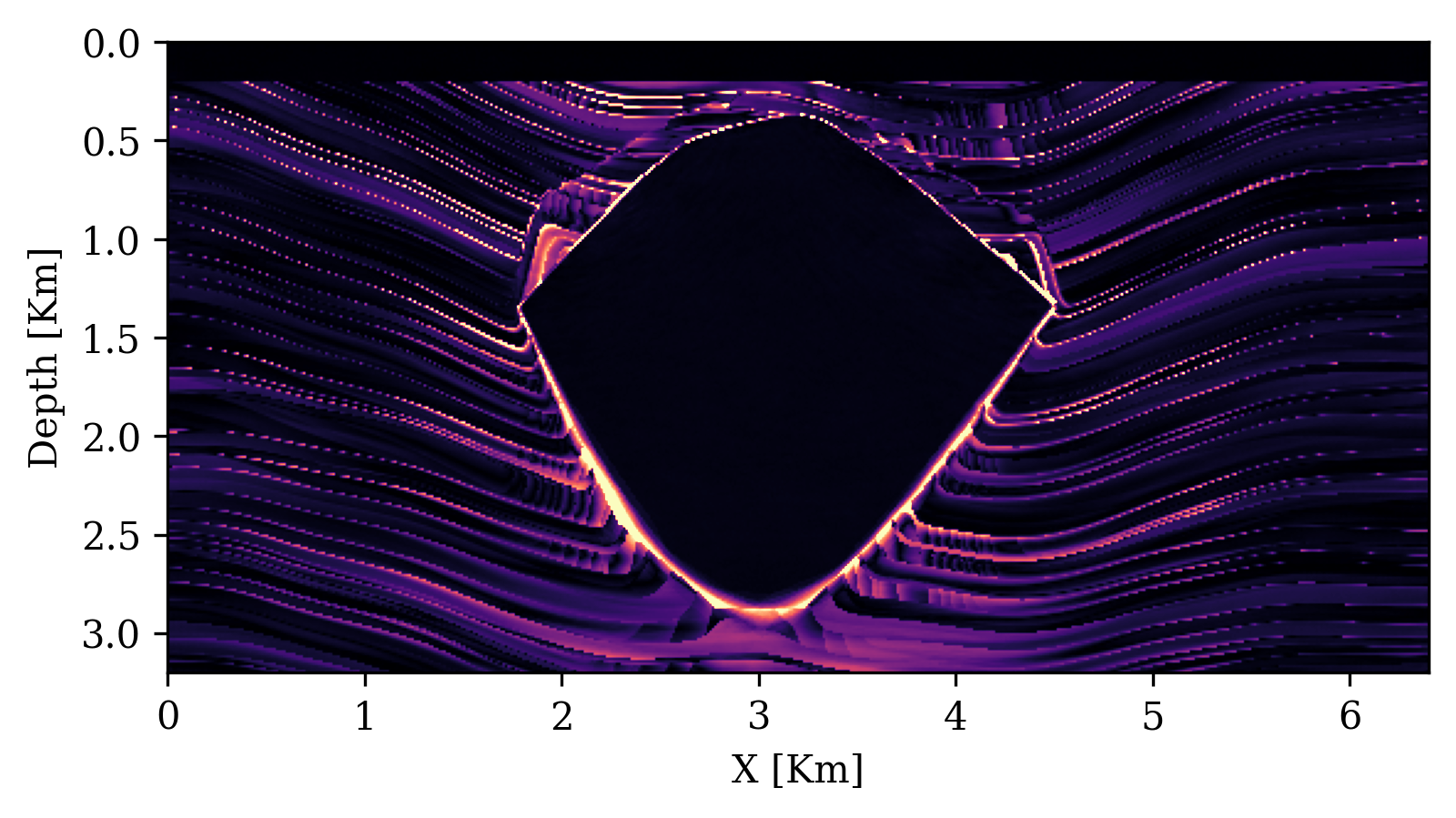}

}

\subcaption{\label{fig-synth-h}Error w/ CIGs RMSE=\(0.10\)}

\end{minipage}%
\newline
\begin{minipage}{0.50\linewidth}

\centering{

\includegraphics[width=1\textwidth,height=\textheight]{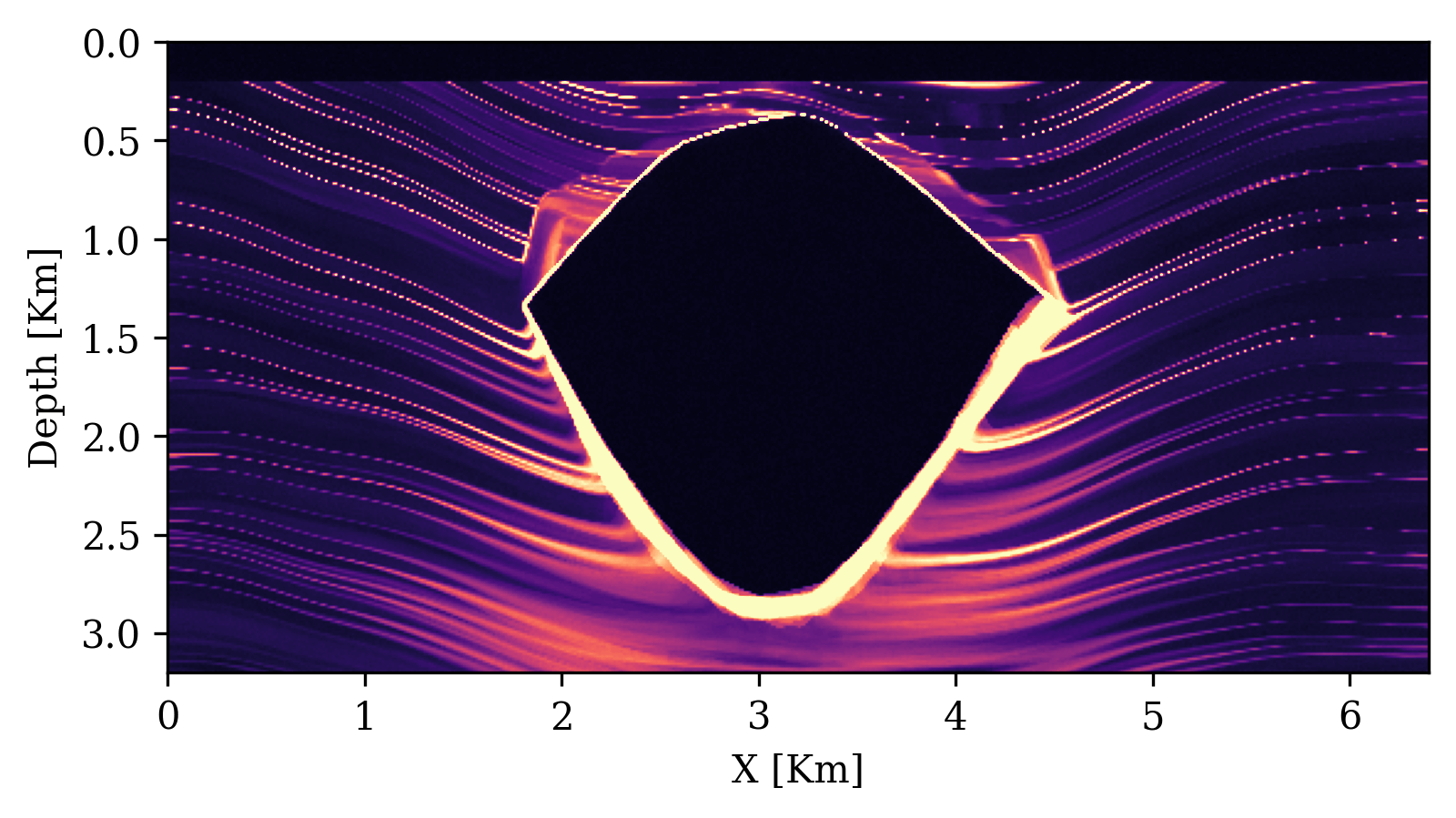}

}

\subcaption{\label{fig-synth-i}Posterior standard deviation w/o CIGs}

\end{minipage}%
\begin{minipage}{0.50\linewidth}

\centering{

\includegraphics[width=1\textwidth,height=\textheight]{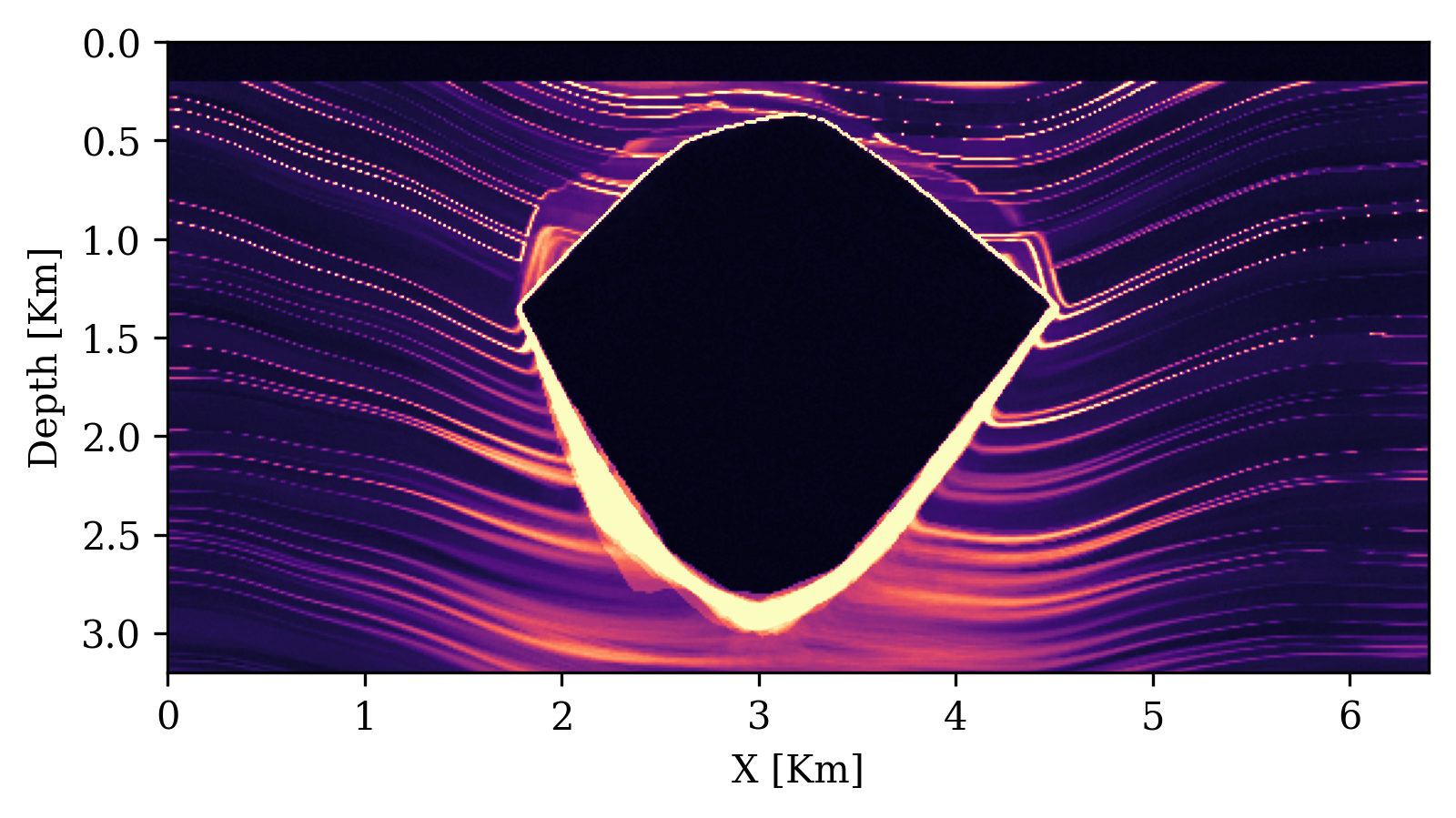}

}

\subcaption{\label{fig-synth-j}Posterior standard deviation w/ CIGs}

\end{minipage}%

\caption{\label{fig-synthoseis-test}Posterior sampling on velocity
models generated by \texttt{Synthoseis}. Based on the image quality
metrics, the CIGs more accurately inform the posterior inference.}

\end{figure}

After training, we consider an example not seen during training for
posterior sampling. The results are shown in
Figure~\ref{fig-synthoseis-test}. Our first observation is that, over
both image quality metrics (SSIM and RMSE) the network trained with CIGs
outperforms the network trained on RTMs only. Furthermore, we are
pleased that although there is uncertainty below the salt structure, as
evidenced by the high standard deviation
Figure~\ref{fig-synth-i},Figure~\ref{fig-synth-j} and posterior means
Figure~\ref{fig-synth-e},Figure~\ref{fig-synth-f} that appear blurred in
regions where the uncertainty is large, the individual posterior samples
in Figure~\ref{fig-synth-c},Figure~\ref{fig-synth-d} exhibit an accurate
approximation of the prior in the way the salt is sharply delineated. As
before, the posterior mean \(\mathbf{x}_{\textrm{mean}}\) and posterior
standard deviation \(\mathbf{x}_{\textrm{std}}\) are calculated from
\(64\) posterior samples. Again, we see reasonably good correlation
between the errors and uncertainty where as expected, these occur mainly
at the bottom of the salt and in the subsalt areas. It is also
encouraging that the unlapping sedimentary layers are well recovered
albeit the inference struggles in the area beneath the deepest point of
the salt.

\subsection{Quantitative assessment of
UQ}\label{quantitative-assessment-of-uq}

Evaluating the quality of the UQ---i.e., quality of the inference of the
posterior, from posterior samples is crucial for understanding how well
the network captures the true variability in the posterior distribution.
In this section, we introduce four approaches to assess the performance
of UQ, namely (1) the ability to warn areas with high error; (2) the
degree of correlation between uncertainty and error; (3) the posterior
coverage, which is defined as the proportion of pixels for which the
range of posterior samples contains the ground truth; and (4) the
ability of posterior samples to fit the observed shot data. For all
metrics, we take the uncertainty to be the standard deviation between
posterior samples and the error calculated between the posterior mean
and a known ground truth velocity model.

\subsubsection{Percentage of regions with large errors but low
uncertainty}\label{percentage-of-regions-with-large-errors-but-low-uncertainty}

\begin{figure}

\begin{minipage}{0.50\linewidth}

\centering{

\includegraphics[width=1\textwidth,height=\textheight]{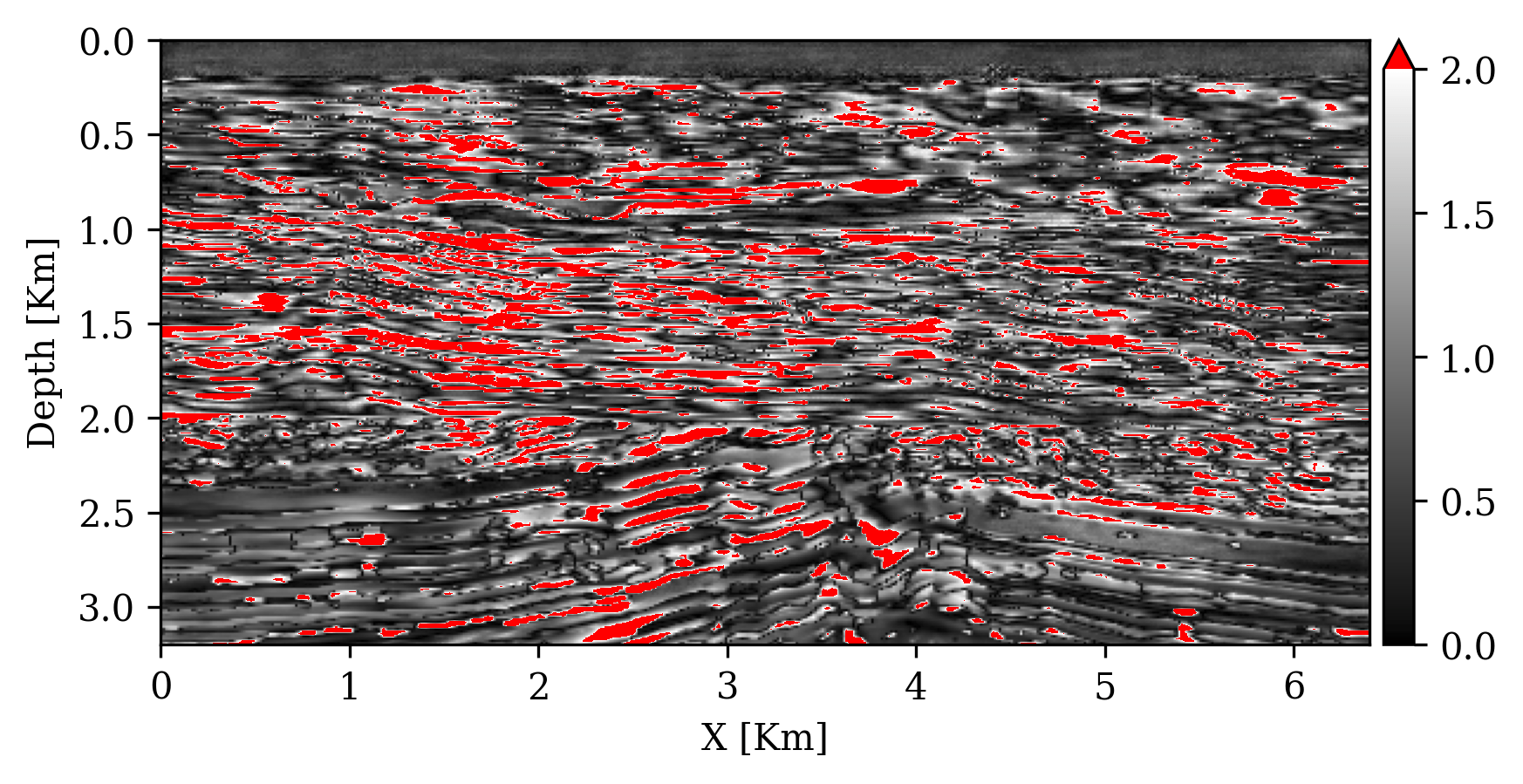}

}

\subcaption{\label{fig-support-synth-salt}Compass w/ RTMs
\(z\)-score=\(10.55\%\)}

\end{minipage}%
\begin{minipage}{0.50\linewidth}

\centering{

\includegraphics[width=1\textwidth,height=\textheight]{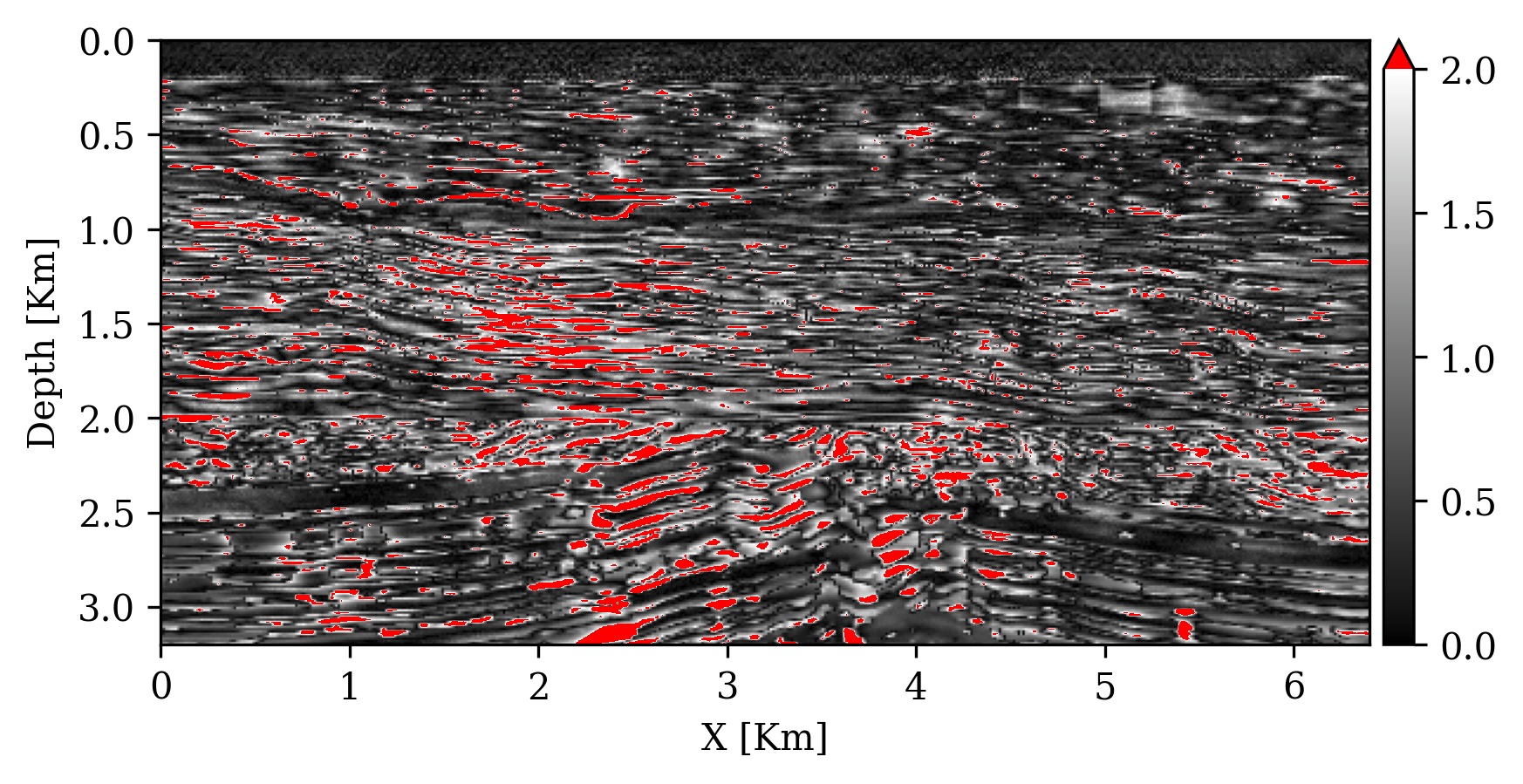}

}

\subcaption{\label{fig-support-synth-salt}Compass w/ CIGs
\(z\)-score=\(6.08\%\)}

\end{minipage}%
\newline
\begin{minipage}{0.50\linewidth}

\centering{

\includegraphics[width=1\textwidth,height=\textheight]{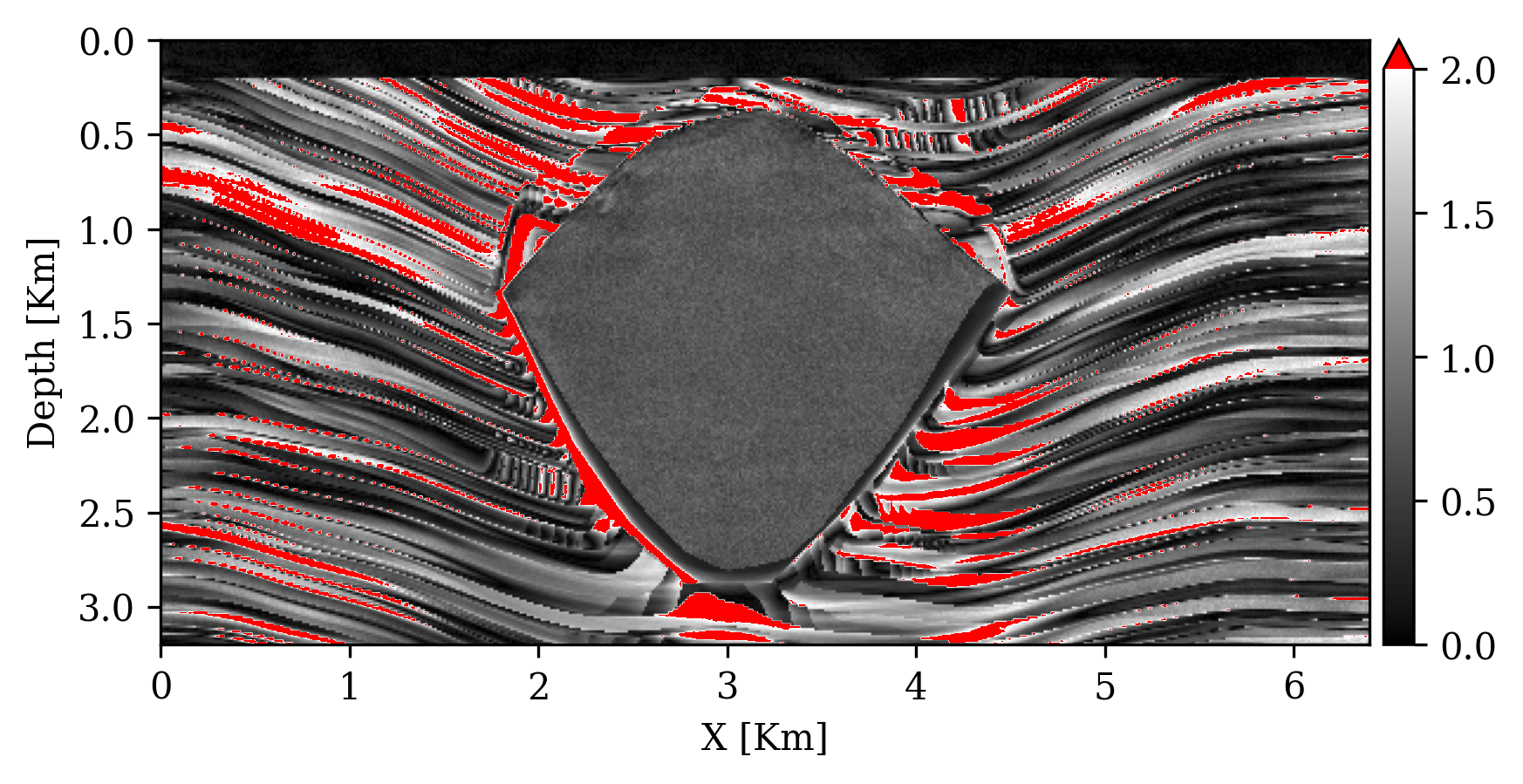}

}

\subcaption{\label{fig-support-synth-salt}Sythoseis w/ RTMs
\(z\)-score=\(7.36\%\)}

\end{minipage}%
\begin{minipage}{0.50\linewidth}

\centering{

\includegraphics[width=1\textwidth,height=\textheight]{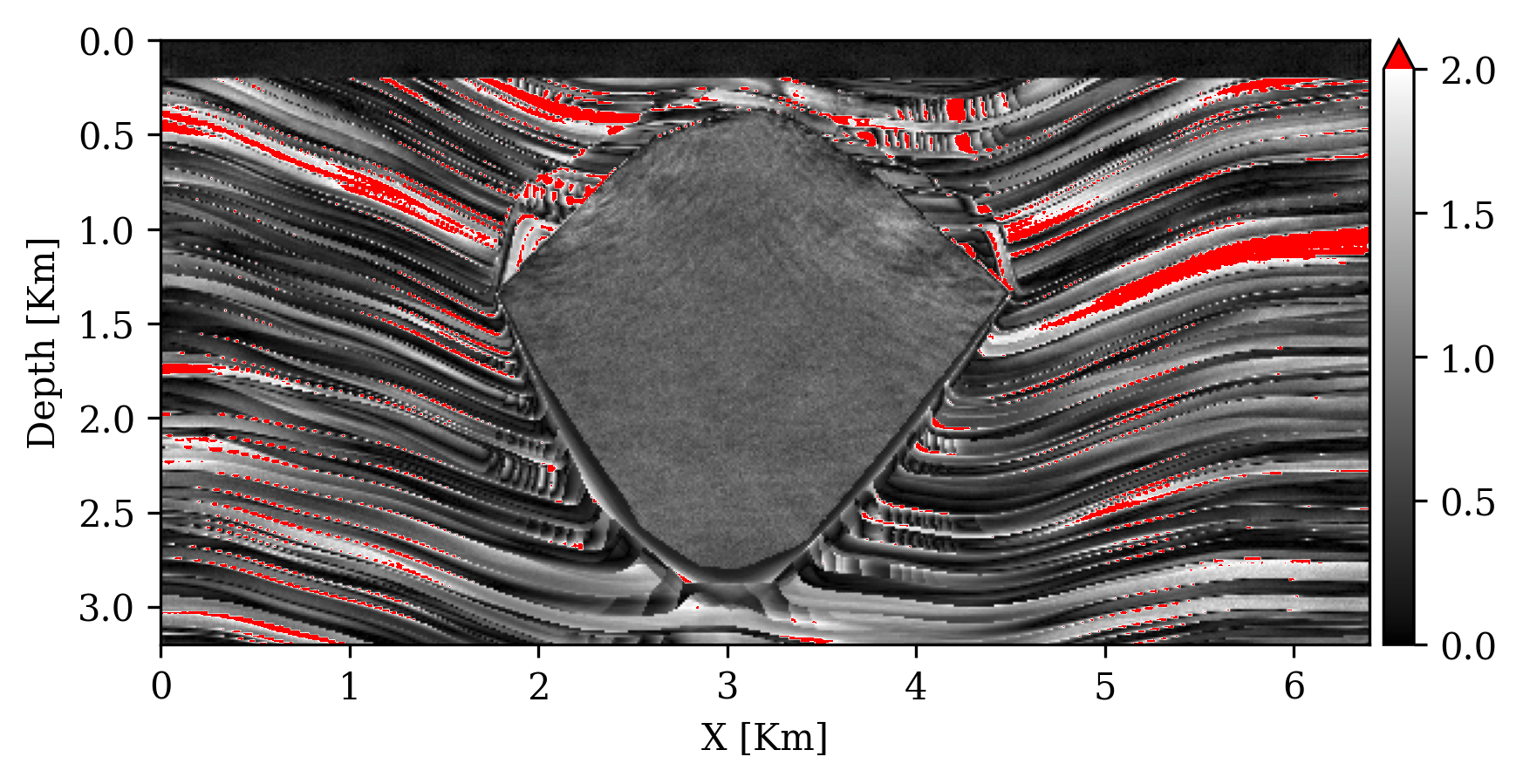}

}

\subcaption{\label{fig-support-synth-salt}Sythoseis w/ CIGs
\(z\)-score=\(4.48\%\)}

\end{minipage}%

\caption{\label{fig-bouman}Comparison of the \(z\)-score between
conditional Diffusion networks trained on RTMs and networks trained on
CIGs with \(24\) non-zero offsets. We desire the \(z\)-score (percentage
of pixels where error is \(2\times\) higher than UQ) to be low.}

\end{figure}

An important feature of UQ is its ability to warn the user of areas that
may have high errors in the reconstruction. By warning, we mean that if
an area has high uncertainty, then that area should have high error.
Thus, we want to avoid high errors that are not followed by high
uncertainty, which could be an indication of poor UQ. To quantify and
visualize this, we use the \(z\)-score as defined
\citep{wu2024principled} by the pixel-wise division of the error by the
uncertainty: \(z\)-score
\(= \lvert\mathbf{x}^{\ast}-\mathbf{x}_{\textrm{mean}} \rvert / \mathbf{x}_{\textrm{std}}\)
where \(\mathbf{x}^\ast\) is the ground truth velocity,
\(\mathbf{x}_{\textrm{mean}}\) the average of the posterior samples, and
\(\mathbf{x}_{\textrm{std}}\) their pixelwise standard deviation. In
Figure~\ref{fig-bouman}, we show that high errors are ``acceptable'' as
long as it is followed by high uncertainty in that same area, while
areas with high errors and low uncertainty will have high values for the
\(z\)-score in this plot. To highlight these areas, grid points with
errors that are \(2\times\) larger than the uncertainty are shown in
red.

To ensure good UQ, we want to minimize the total red area. Thus, we
define the quantitative metric as the percentage of pixels that are red
and aim for this metric to be low. In Table~\ref{tbl-performance}, we
show the average \(z\)-score percentage over posterior sampling results
of \(50\) test unseen observations for both RTMs and CIGs trained
networks. From Figure~\ref{fig-bouman} we make the following
observations: for both models, the total area of the red regions is
significantly smaller when the posterior is conditioned on CIGs. We also
observe that red regions for the Compass model tend to be associated
with areas where the velocity kickback exhibits lateral variations. With
few exceptions, we see that red regions for the salt models generated by
\texttt{Synthoseis} are located near steeply dipping events. Overall,
percentages of red area are significantly reduced thanks to the use of
CIGs.

\subsubsection{Degree of calibration}\label{degree-of-calibration}

\begin{figure}

\begin{minipage}{0.25\linewidth}

\includegraphics[width=1\textwidth,height=\textheight]{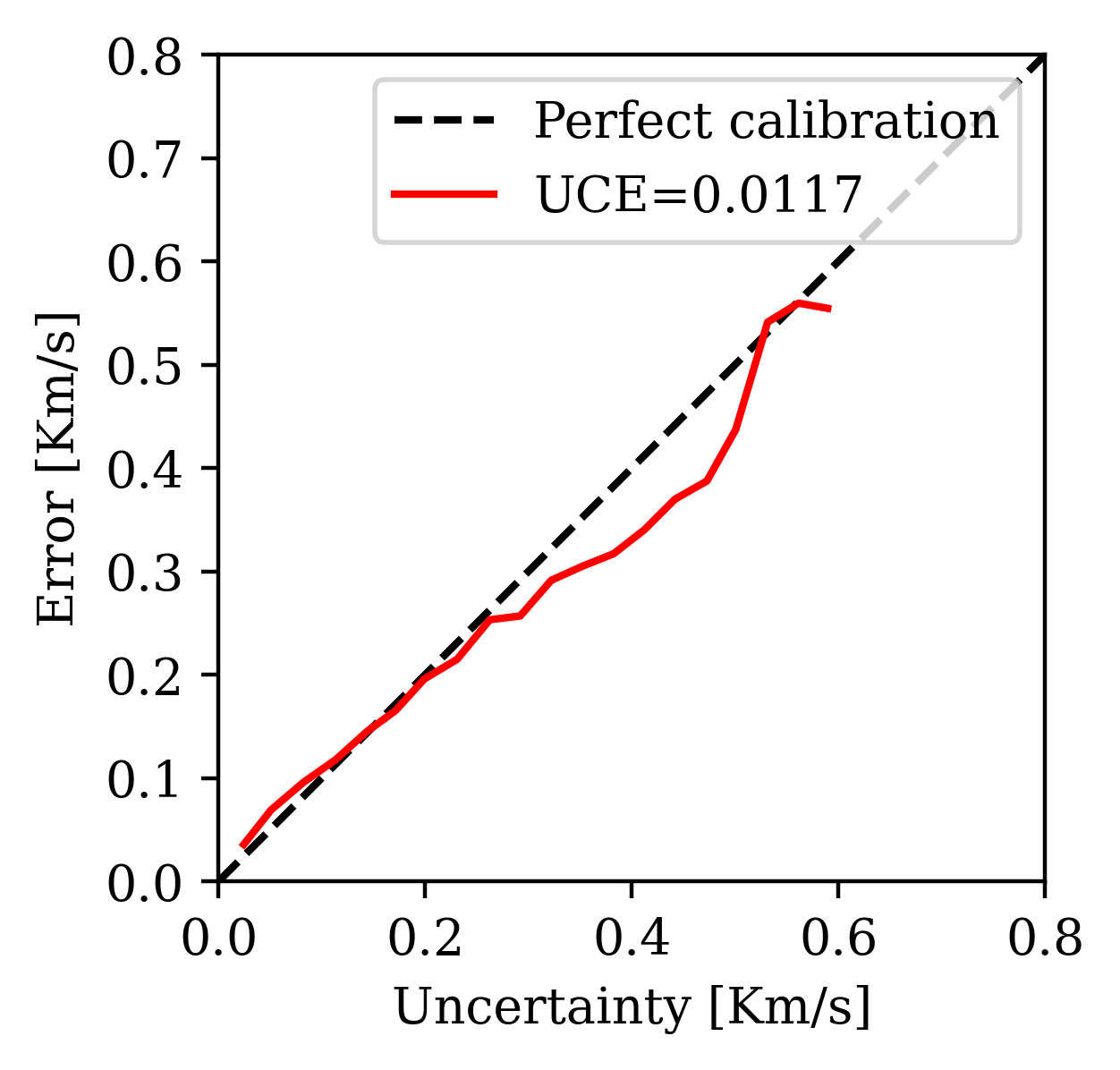}

\subcaption{\label{}Compass w/ RTMs}
\end{minipage}%
\begin{minipage}{0.25\linewidth}

\includegraphics[width=1\textwidth,height=\textheight]{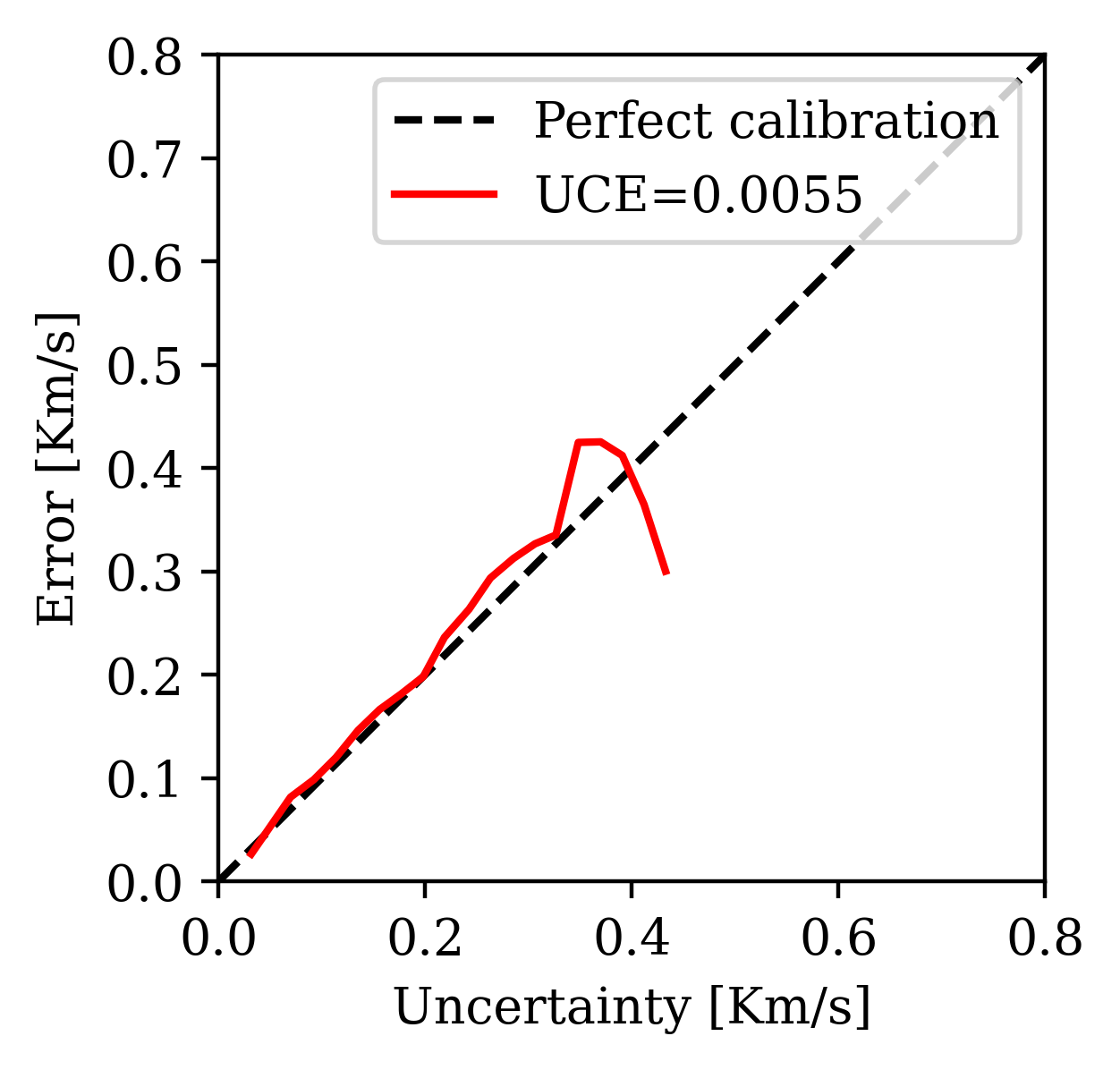}

\subcaption{\label{}Compass w/ CIGs}
\end{minipage}%
\begin{minipage}{0.25\linewidth}

\includegraphics[width=1\textwidth,height=\textheight]{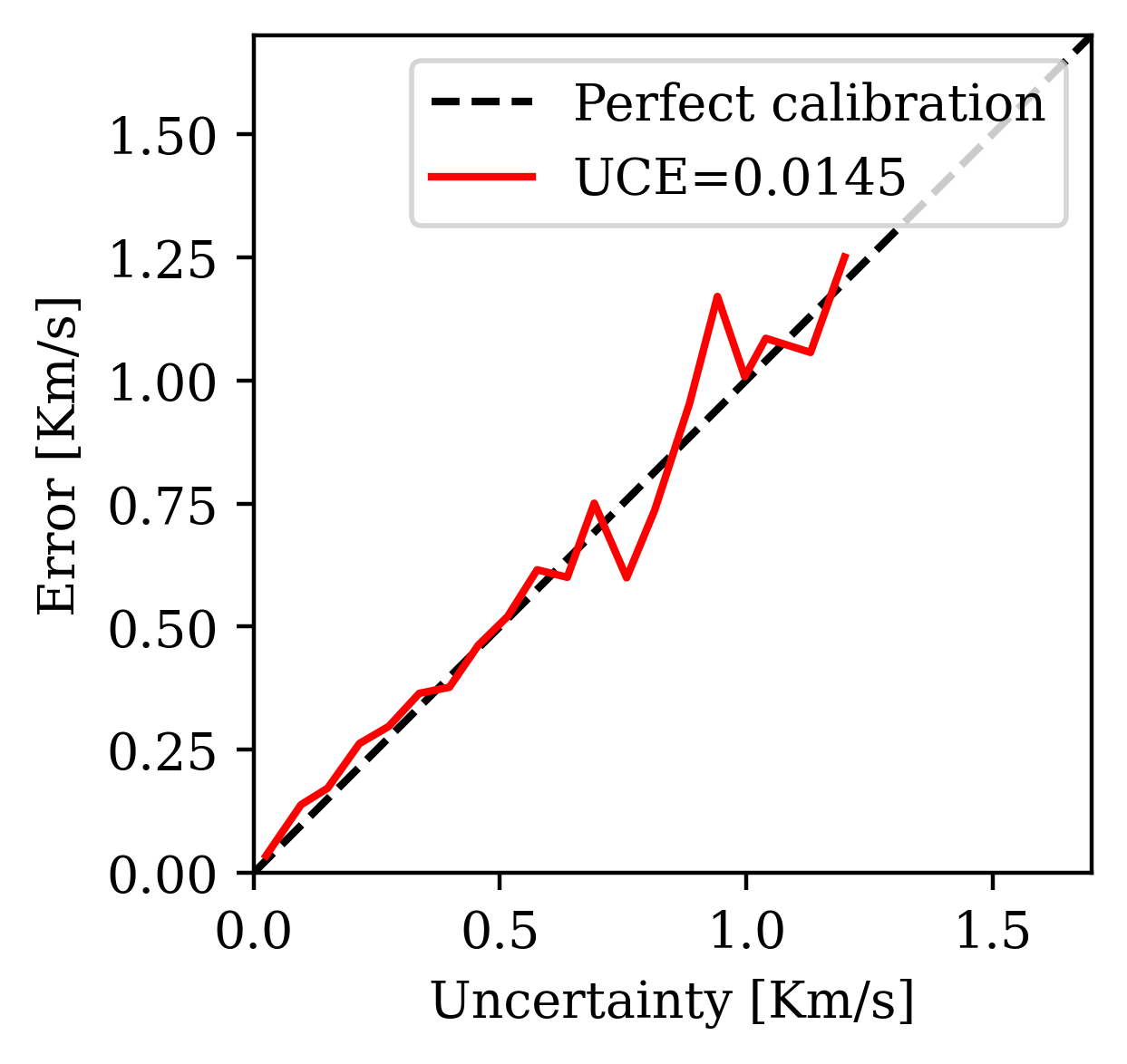}

\subcaption{\label{}\texttt{Synthoseis} w/ RTMs}
\end{minipage}%
\begin{minipage}{0.25\linewidth}

\includegraphics[width=1\textwidth,height=\textheight]{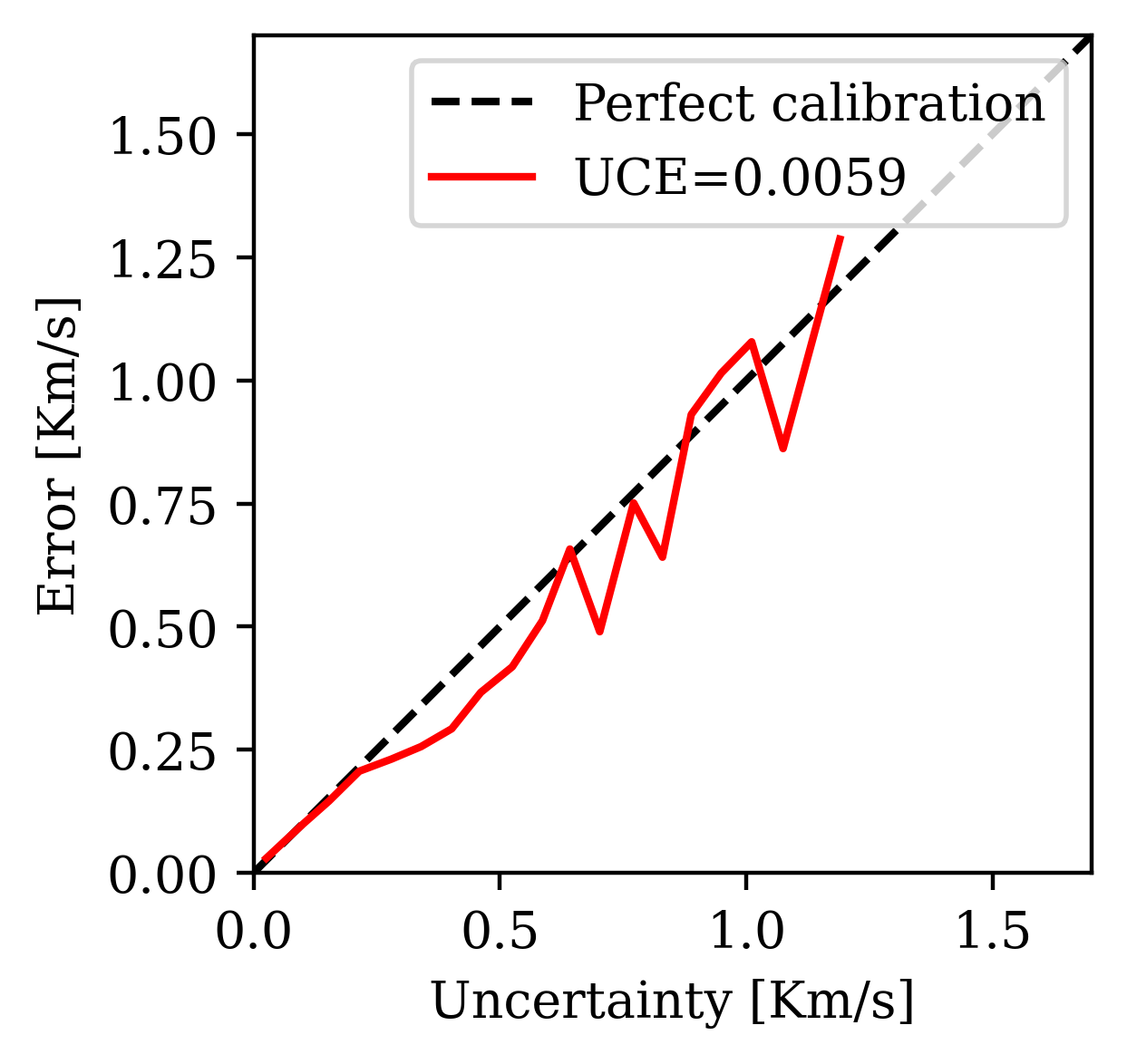}

\subcaption{\label{}\texttt{Synthoseis} w/ CIGs}
\end{minipage}%

\caption{\label{fig-calibration}Comparison of the uncertainty
calibration of a conditional Diffusion network trained on RTMs compared
with network trained on CIGs with \(24\) non-zero offsets. For both
synthetic datasets, the CIGs trained network is better calibrated as
evidenced by the lower UCE metric.}

\end{figure}

While correctly warning of high errors is an indication that the UQ is
reasonable, it does not provide a precise quantitative metric for the
correspondence between error and UQ at various magnitudes. For this
purpose, we use the calibration test from
\citep{laves2020well, guo2017calibration} to quantify the correlation
between predicted uncertainty and error. By binning the pixels over
different magnitudes, we can build a correlation plot wherein we want
pixels that are placed into bins of certain errors magnitudes to have
similar uncertainty magnitudes. Figure~\ref{fig-calibration} shows the
application of this test to the UQ results for the Compass example in
Figure~\ref{fig-compass-test} and the \texttt{Synthoseis} example in
Figure~\ref{fig-synthoseis-test}. The uncertainty calibration error
(UCE) is a single scalar summarizing the performance of this test by
calculating the area between the red curve and the optimal calibration
on the diagonal dashed line. Therefore, lower UCE values indicate
better-calibrated uncertainty. Again, we observe major improvements due
to the use of CIGs for both models. In Table~\ref{tbl-performance}, we
show the average UCE for posterior sampling results on \(50\) unseen
test observations for both RTMs and CIGs networks.

\subsubsection{Posterior coverage
percentage}\label{posterior-coverage-percentage}

\begin{figure}

\begin{minipage}{0.50\linewidth}

\includegraphics[width=1\textwidth,height=\textheight]{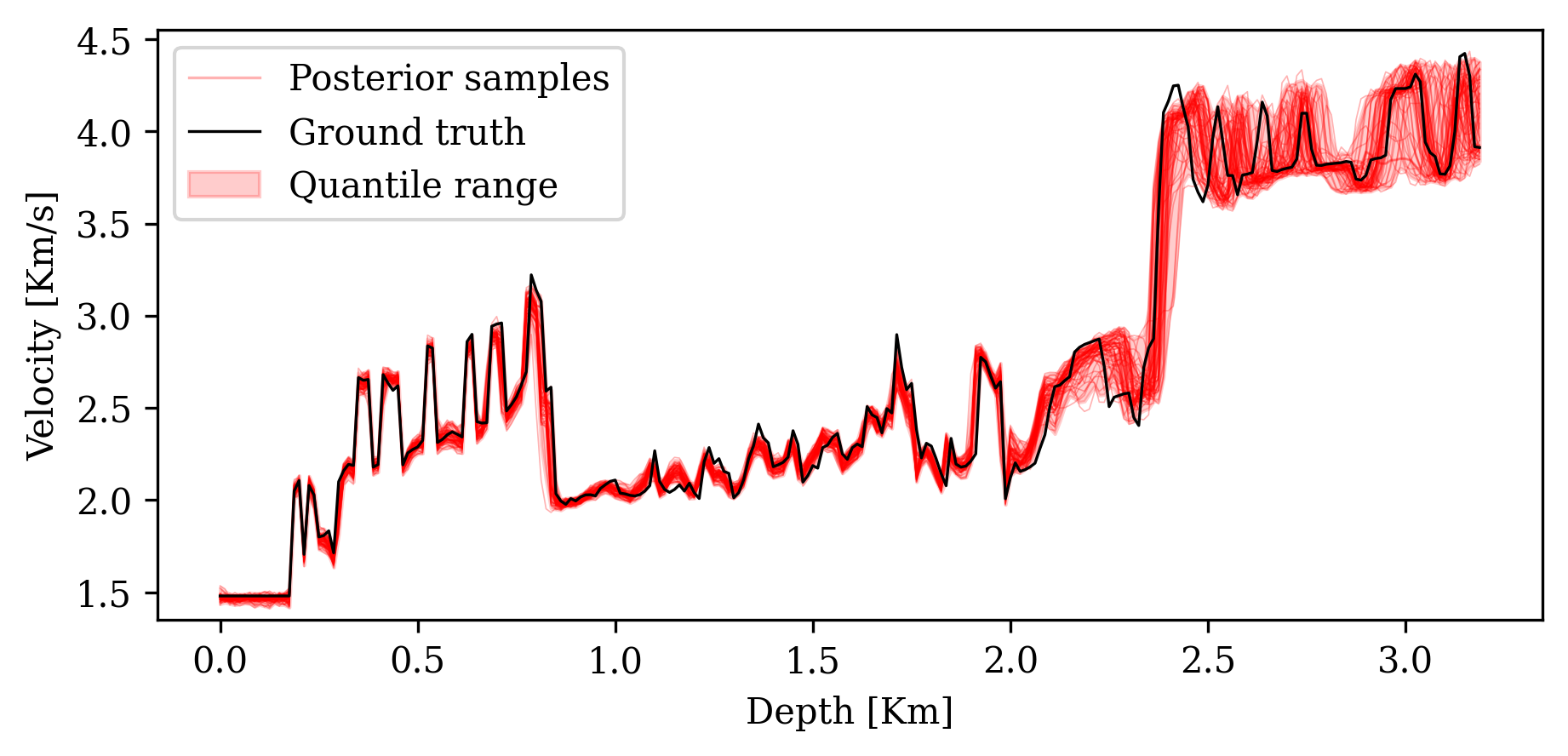}

\subcaption{\label{}Traces Compass w/ RTMs Coverage=\(82.1\%\)}
\end{minipage}%
\begin{minipage}{0.50\linewidth}

\includegraphics[width=1\textwidth,height=\textheight]{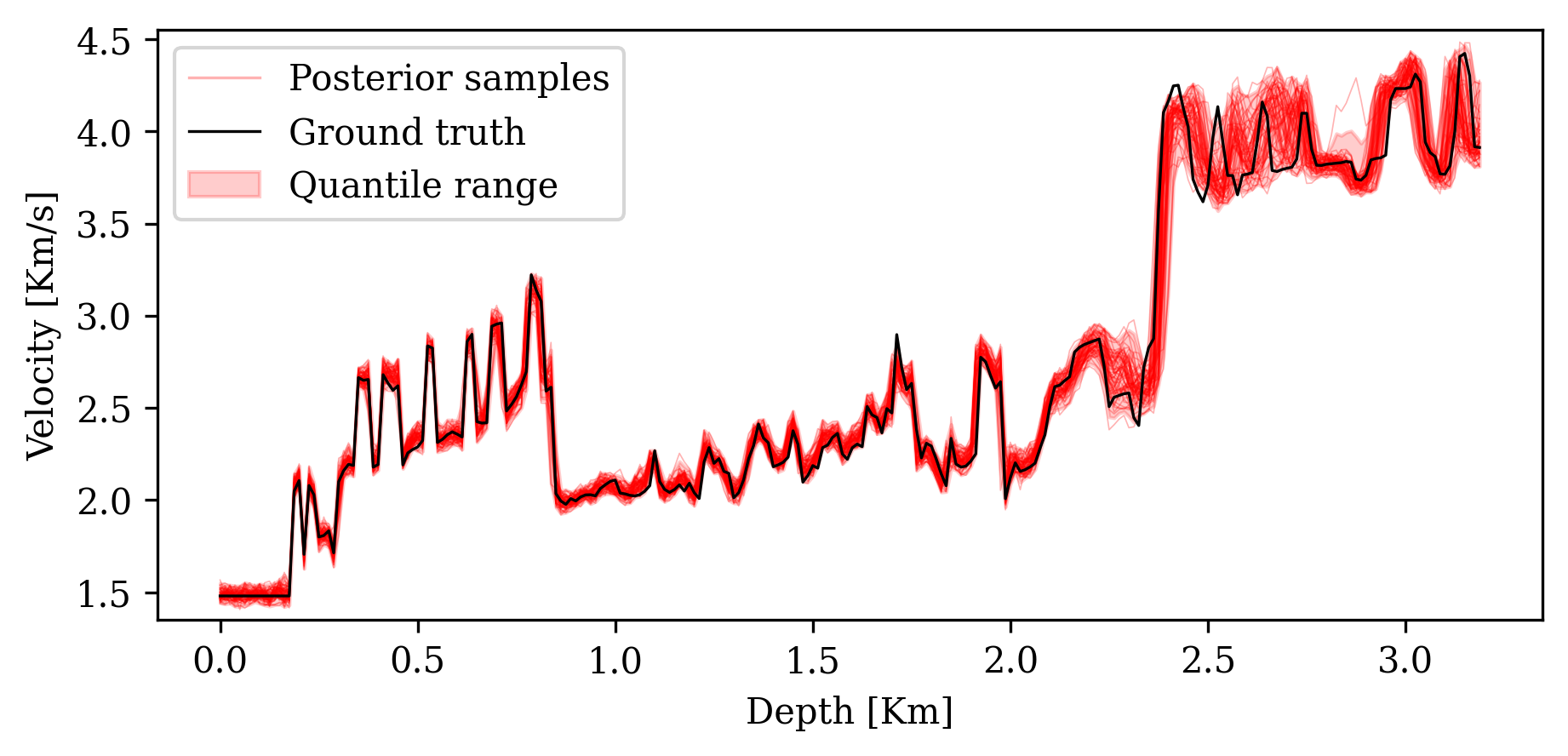}

\subcaption{\label{}Traces Compass w/ CIGs Coverage=\(90.6\%\)}
\end{minipage}%
\newline
\begin{minipage}{0.50\linewidth}

\centering{

\includegraphics[width=1\textwidth,height=\textheight]{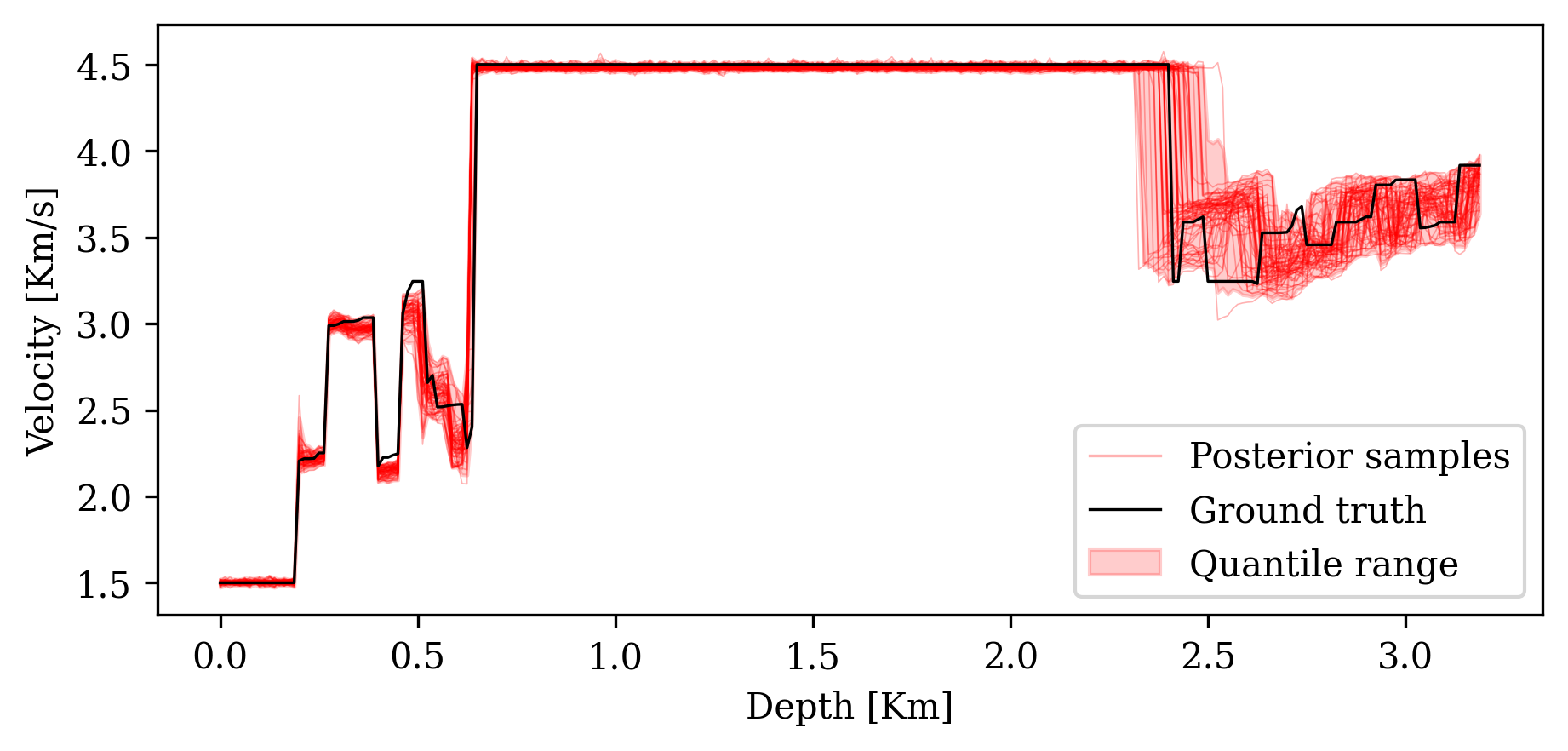}

}

\subcaption{\label{fig-synth-salt-traces}Traces \texttt{Synthoseis} w/
RTMs Coverage=\(87.0\%\)}

\end{minipage}%
\begin{minipage}{0.50\linewidth}

\includegraphics[width=1\textwidth,height=\textheight]{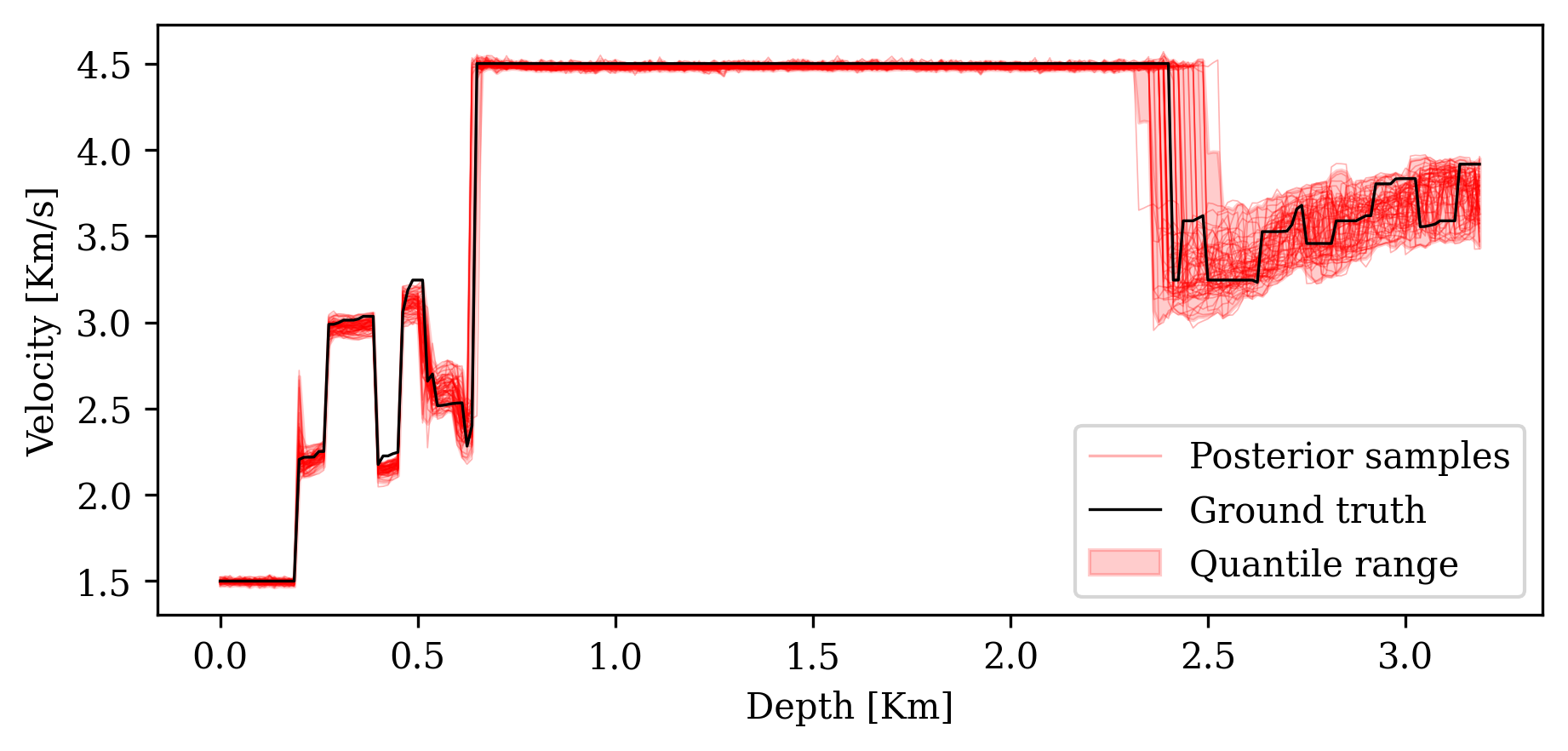}

\subcaption{\label{}Traces \texttt{Synthoseis} w/ CIGs
Coverage=\(89.7\%\)}
\end{minipage}%

\caption{\label{fig-coverage}Vertical traces through the posterior
samples and the ground truth velocity model used to calculate the
posterior coverage metric.}

\end{figure}

Next, we quantitatively evaluate the coverage of the posterior samples.
Coverage measures whether the true velocity model is contained within
the spread of the posterior distribution. This is an important
characteristic of Bayesian methods, as we want the ground truth velocity
model to be included within the posterior samples
\citep{tachella2023equivariant}. To quantify coverage in percentages, we
compute the lower and upper percentiles of the posterior samples at each
pixel and test if the ground truth velocity model is contained within
that range. The quantified metric corresponds to the percentage of
pixels for which the calculated range contains the ground truth velocity
model. Ideally, this coverage percentage should be high. Visually this
test is intuitively understood in Figure~\ref{fig-coverage} where we
plot samples of the posterior over a single vertical trace and compare
them with the ground truth velocity. For our metric, we calculate the
range using the upper \(99\%\) and lower \(1\%\) percentiles. Although
we only show a trace in these figures, note that the coverage metric is
calculated over all pixels in the reconstructed velocity models. In
Table~\ref{tbl-performance}, we include the average coverage percentage
over \(50\) test samples for both RTMs and CIGs trained networks. We
want this metric to be high because we wish our posterior distribution
to always contain the ground truth velocity model.

\subsection{Shot data residual of posterior
samples}\label{shot-data-residual-of-posterior-samples}

\begin{figure}

\begin{minipage}{0.50\linewidth}

\centering{

\includegraphics[width=0.8\textwidth,height=\textheight]{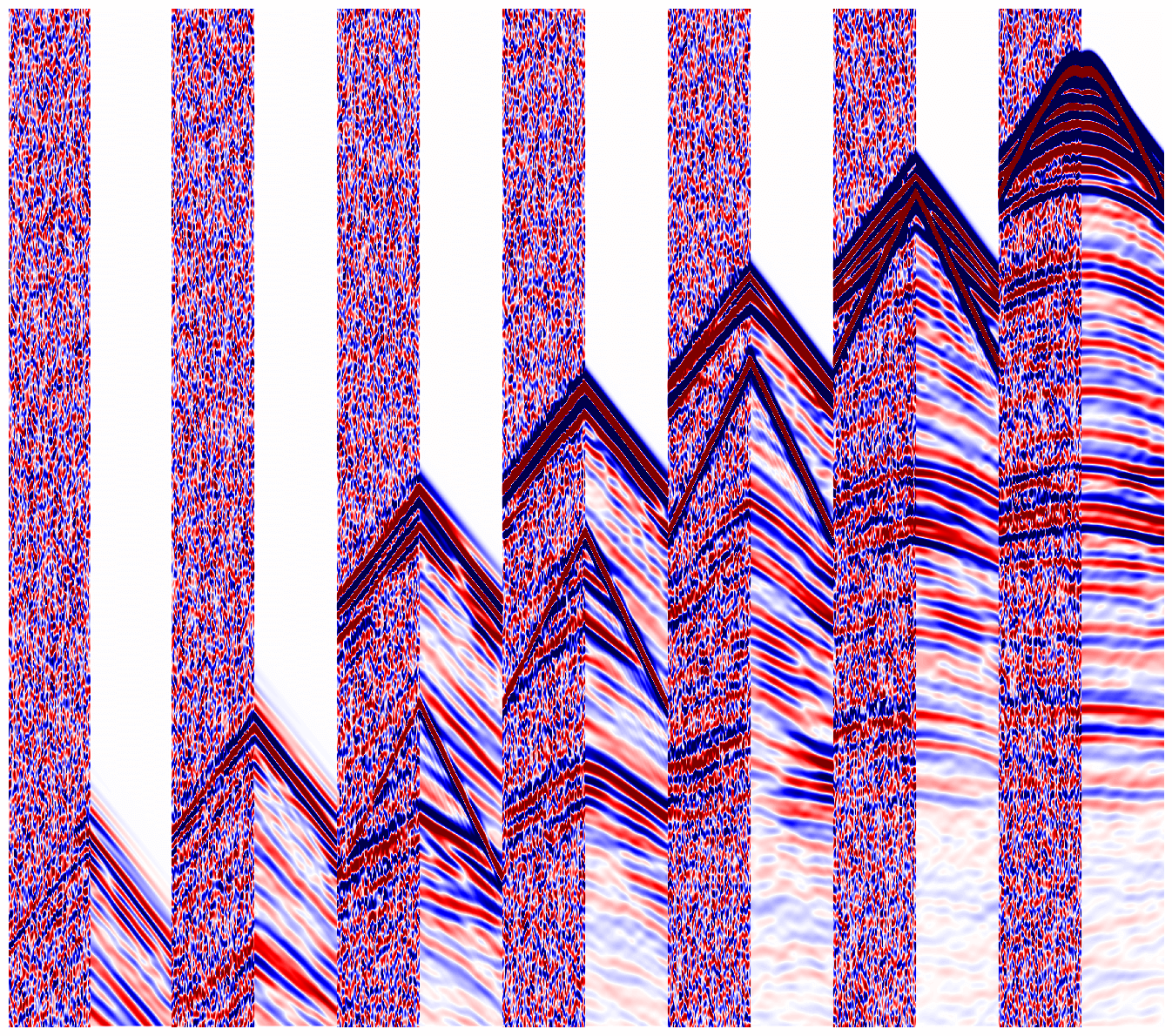}

}

\subcaption{\label{fig-misfit-compass-rtm}Compass w/ RTM data fit
\(56\%\)}

\end{minipage}%
\begin{minipage}{0.50\linewidth}

\centering{

\includegraphics[width=0.8\textwidth,height=\textheight]{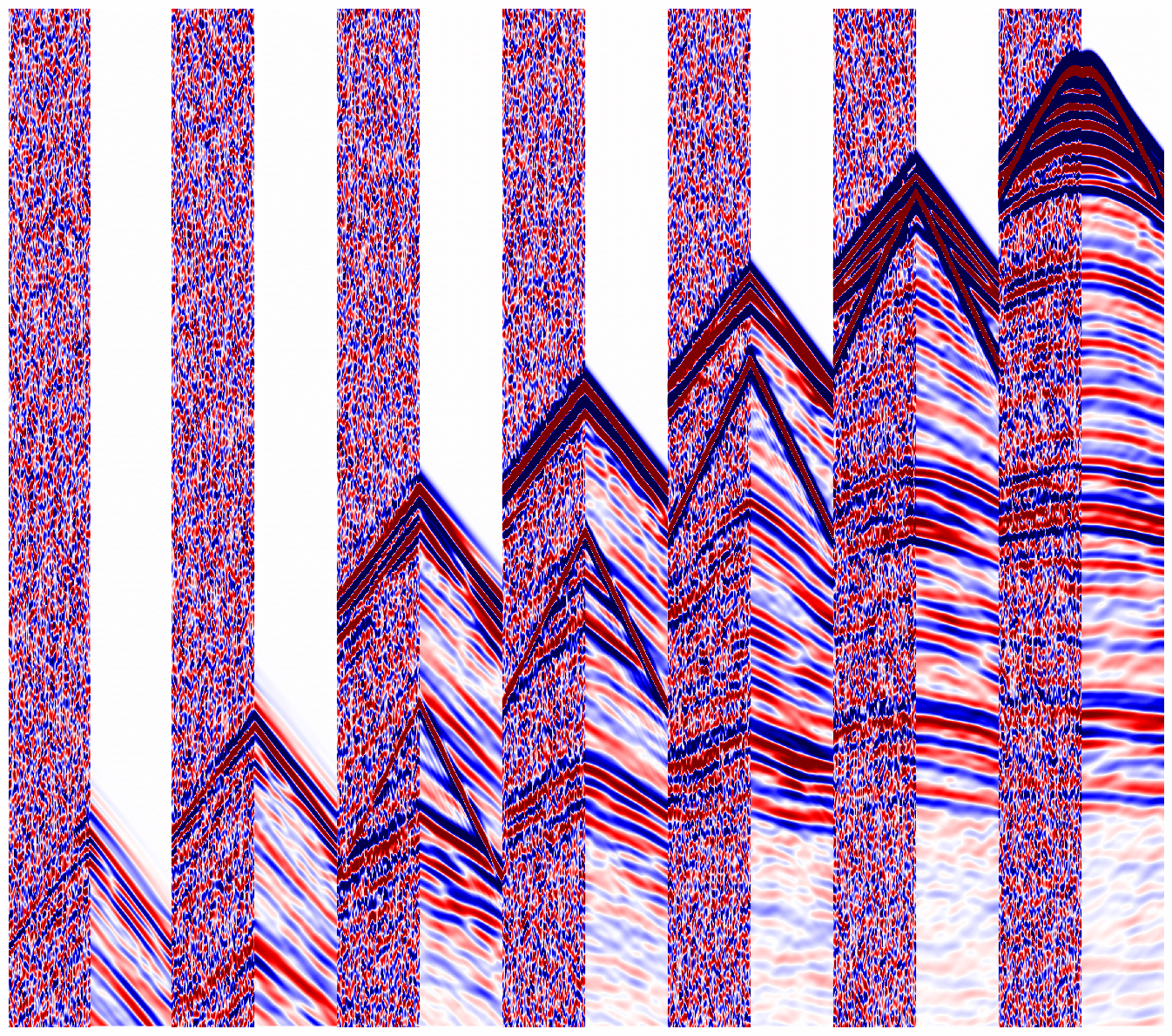}

}

\subcaption{\label{fig-misfit-compass-cig}Compass w/ CIG data fit
\(58\%\)}

\end{minipage}%
\newline
\begin{minipage}{0.50\linewidth}

\centering{

\includegraphics[width=0.8\textwidth,height=\textheight]{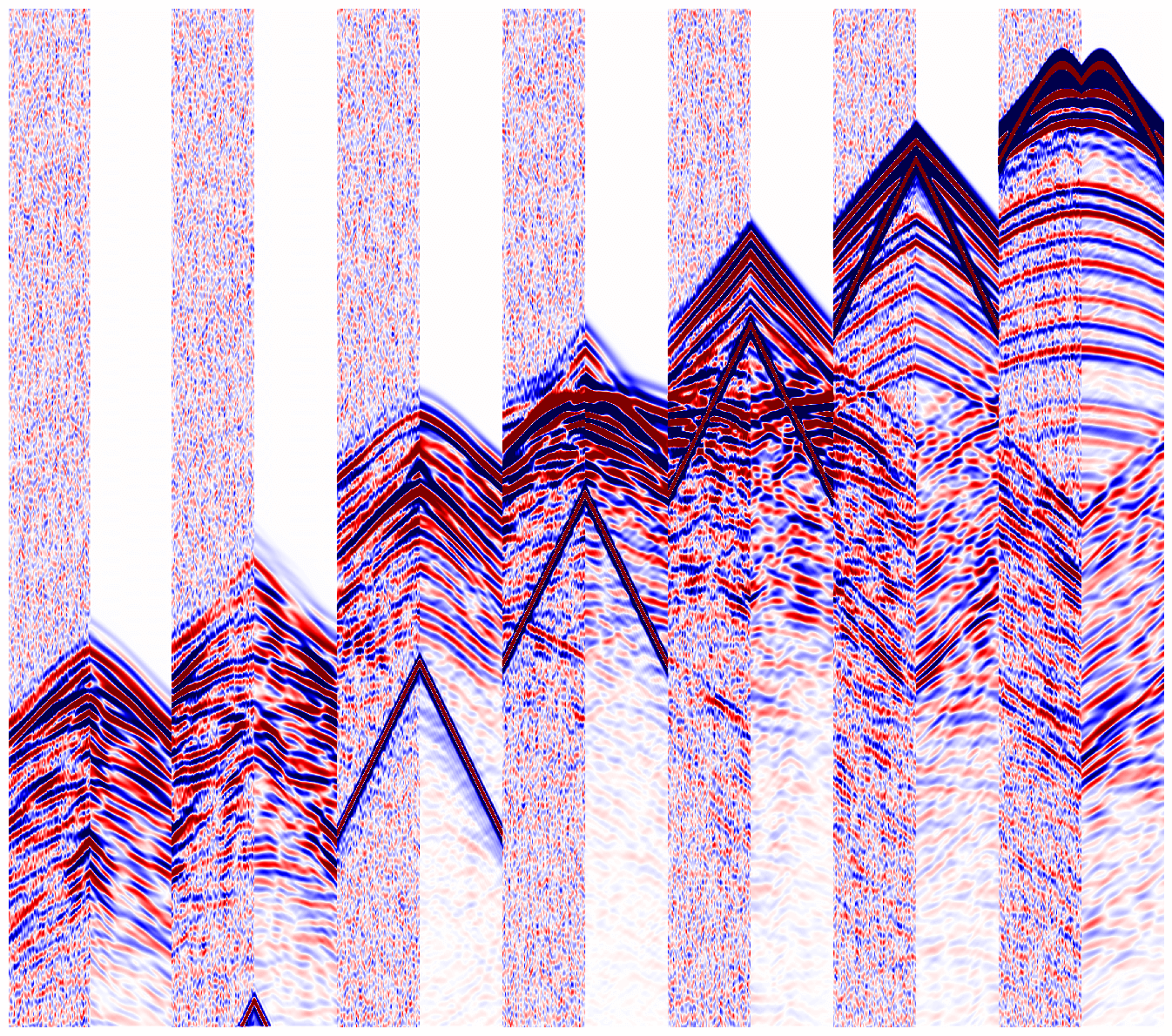}

}

\subcaption{\label{fig-misfit-compass-rtm}\texttt{Synthoseis} w/ RTM
data fit \(41\%\)}

\end{minipage}%
\begin{minipage}{0.50\linewidth}

\centering{

\includegraphics[width=0.8\textwidth,height=\textheight]{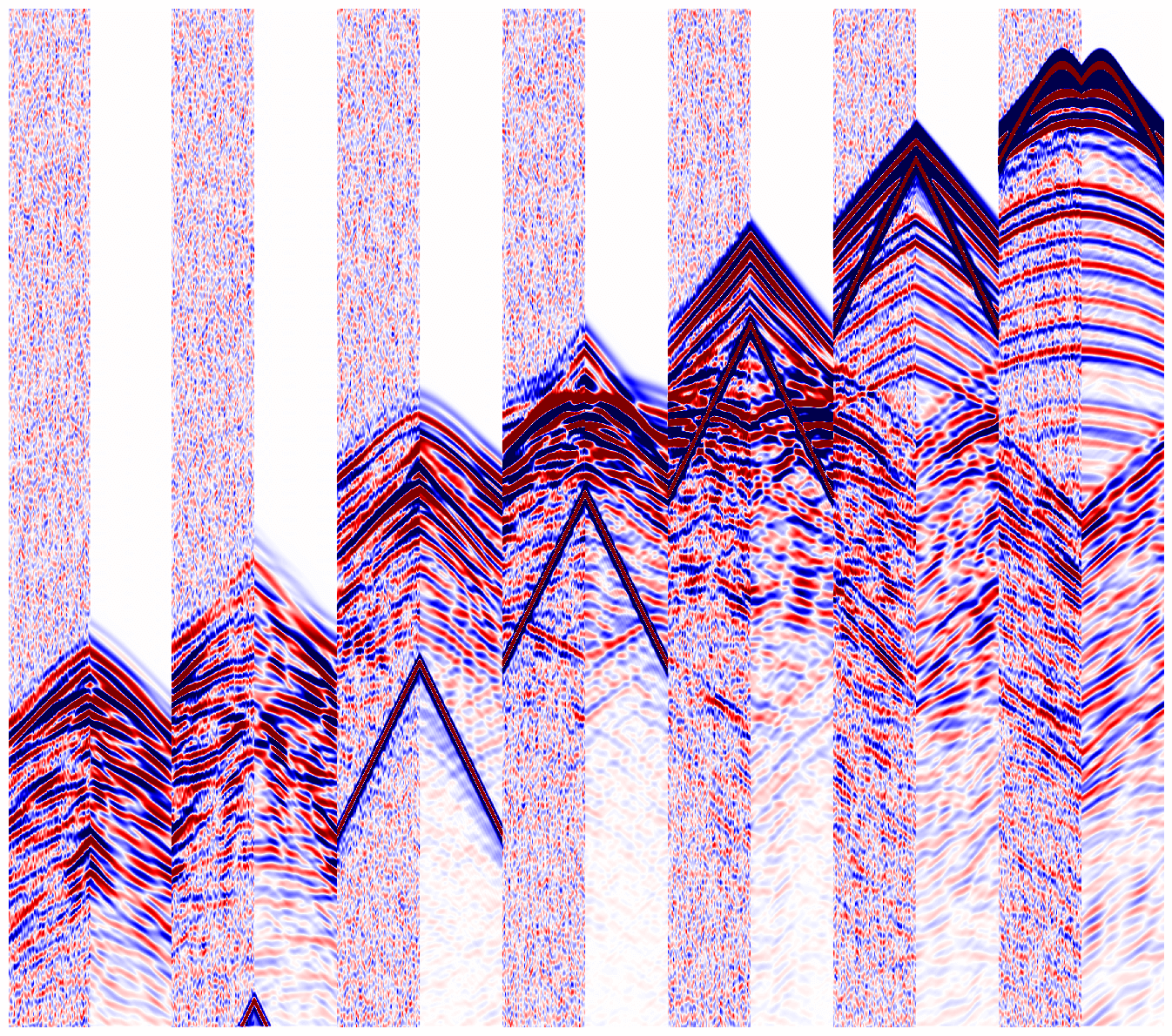}

}

\subcaption{\label{fig-misfit-compass-cig}\texttt{Synthoseis} w/ CIG
data fit \(47\%\)}

\end{minipage}%

\caption{\label{fig-datamisfit}Interleaved comparison between paired
bins of observed and synthetic shot data.}

\end{figure}

Even though the above quantitative assessment of UQ is valuable, it
relies on having access to ground truth velocity models rendering these
metric impractical for field data where the ground truth velocity models
are unknown. For this reason, it is as a final check important to make
sure that the proposed velocity-model inference produces synthetic
gathers that fit observed shot gathers. In other words, while the
generative samples \(\mathbf{x}_{\textrm{post}}\) shown previously seem
to visually match the correct prior, we will verify that these are
indeed Bayesian and respect the data likelihood
\(p(\mathbf{y} \mid \mathbf{x})\). We take the posterior samples and
pass it through the nonlinear forward wave operator
\(\mathcal{F}({\mathbf{x_{\textrm{post}}}})\) and analyze the fit to the
observed shot data \(\mathbf{y}^{\textrm{obs}}\). In
Figure~\ref{fig-datamisfit}, we show a pairwise binned comparison
between the predicted and observed shot gather interleaved with
increasing receiver coordinate. To quantify the fit, we take the
division between the noise norm and the data residual that we define as
\(\|  \boldsymbol{\varepsilon}  \|_2 / \|   \mathcal{F}(\mathbf{x_{\textrm{post}}}) - \mathbf{y}^{\textrm{obs}}   \|_2\)
and report this number as a percentage. With \(100\%\) corresponding to
the residual having the same normed magnitude as the noise, in other
words, a perfect fit. For both models, the network conditioned on the
CIGs better fits the shot gathers.

Results of the four performance metrics: (1) percentage of regions with
large errors, (2) degree of calibration, (3) posterior coverage
percentage, and (4) posterior fit of shot data are included in
Table~\ref{tbl-performance} for networks trained with RTMs or CIGs. From
these results, we conclude that conditioning on CIGs improves
performance with respect to RMSE and SSIM for the recovery results
themselves, and with respect to these four metrics that reflect the
quality of the UQ. This improvement is due to the CIGs containing more
information than the RTMs, even in situations where the
migration-velocity models are poor. As a result, CIGs provide a more
accurate representation of the original posterior that is conditioned on
raw shot records.

We include the performance on the same Compass dataset used in
\citep{yin2024wise}, showing that our Diffusion network with average
SSIM of \(0.85\) outperforms the normalizing flow, which had an average
SSIM of \(0.63\) \citep{yin2024wise}. While this increase in performance
is encouraging, we cannot definitively conclude that Diffusion is in
general better than normalizing flows since our Diffusion network uses
different architectures than the normalizing flow in \citep{yin2024wise}
and also has more training costs. A careful comparison between these two
frameworks is beyond the scope of this paper but would be a valuable
avenue for future work.

\begin{longtable}[]{@{}
  >{\raggedright\arraybackslash}p{(\columnwidth - 12\tabcolsep) * \real{0.2115}}
  >{\raggedright\arraybackslash}p{(\columnwidth - 12\tabcolsep) * \real{0.1538}}
  >{\raggedright\arraybackslash}p{(\columnwidth - 12\tabcolsep) * \real{0.1346}}
  >{\raggedright\arraybackslash}p{(\columnwidth - 12\tabcolsep) * \real{0.1346}}
  >{\raggedright\arraybackslash}p{(\columnwidth - 12\tabcolsep) * \real{0.1346}}
  >{\raggedright\arraybackslash}p{(\columnwidth - 12\tabcolsep) * \real{0.1154}}
  >{\raggedright\arraybackslash}p{(\columnwidth - 12\tabcolsep) * \real{0.1154}}@{}}
\caption{Image and uncertainty quality metrics on Compass and
\texttt{Synthoseis} datasets.}\label{tbl-performance}\tabularnewline
\toprule\noalign{}
\begin{minipage}[b]{\linewidth}\raggedright
Dataset
\end{minipage} & \begin{minipage}[b]{\linewidth}\raggedright
RMSE \(\downarrow\)
\end{minipage} & \begin{minipage}[b]{\linewidth}\raggedright
SSIM \(\uparrow\)
\end{minipage} & \begin{minipage}[b]{\linewidth}\raggedright
Coverage \([\%]\) \(\uparrow\)
\end{minipage} & \begin{minipage}[b]{\linewidth}\raggedright
UCE \(\downarrow\)
\end{minipage} & \begin{minipage}[b]{\linewidth}\raggedright
\(z\)-score \([\%]\) \(\downarrow\)
\end{minipage} & \begin{minipage}[b]{\linewidth}\raggedright
Data fit \([\%]\) \(\uparrow\)
\end{minipage} \\
\midrule\noalign{}
\endfirsthead
\toprule\noalign{}
\begin{minipage}[b]{\linewidth}\raggedright
Dataset
\end{minipage} & \begin{minipage}[b]{\linewidth}\raggedright
RMSE \(\downarrow\)
\end{minipage} & \begin{minipage}[b]{\linewidth}\raggedright
SSIM \(\uparrow\)
\end{minipage} & \begin{minipage}[b]{\linewidth}\raggedright
Coverage \([\%]\) \(\uparrow\)
\end{minipage} & \begin{minipage}[b]{\linewidth}\raggedright
UCE \(\downarrow\)
\end{minipage} & \begin{minipage}[b]{\linewidth}\raggedright
\(z\)-score \([\%]\) \(\downarrow\)
\end{minipage} & \begin{minipage}[b]{\linewidth}\raggedright
Data fit \([\%]\) \(\uparrow\)
\end{minipage} \\
\midrule\noalign{}
\endhead
\bottomrule\noalign{}
\endlastfoot
Compass w/o CIGs & \(0.12\) & \(0.81\) & \(73.8\) & \(0.013\) & \(10.7\)
& \(53.7\) \\
Compass w/ CIGs & \(\mathbf{0.11}\) & \(\mathbf{0.85}\) &
\(\mathbf{74.8}\) & \(\mathbf{0.011 }\) & \(\mathbf{9.9}\) &
\(\mathbf{54.2}\) \\
\texttt{Synthoseis} w/o CIGs & \(0.085\) & \(0.91\) & \(72.5\) &
\(0.012\) & \(7.8\) & \(77.1\) \\
\texttt{Synthoseis} w/ CIGs & \(\mathbf{0.079}\) & \(\mathbf{0.92 }\) &
\(\mathbf{74.6 }\) & \(\mathbf{0.009}\) & \(\mathbf{5.4 }\) &
\(\mathbf{79.6 }\) \\
\end{longtable}

\section{Complex case studies}\label{complex-case-studies}

So far, the stylized examples have exhibited relatively minor
complexity, which in part explains the success of the inference
discussed so far where the reconstructed velocity models significantly
bring down the residuals. Due to the the always present amortization gap
\citep{marino2018iterative, orozco2024aspire}, which corresponds to a
drop in accuracy due to an attempt to generalize the performance of the
network over many velocity models instead of focusing the inference on a
single shot dataset as in non-amortized methods
\citep{zhao2022bayesian, siahkoohi2023reliable, izzatullah2024physics},
we cannot expect the presented amortized approach to perform well on
larger models with increased geological complexity. Using an example
involving the SEAM salt model \citep{fehler2011model}, we demonstrate
how our inference approach can be extended to handle complex salt plays.
Finally, we also consider a field dataset to demonstrate the importance
of having access to relevant training data.

\subsection{Salt flooding with ASPIRE}\label{salt-flooding-with-aspire}

\begin{figure}

\begin{minipage}{0.50\linewidth}

\centering{

\includegraphics[width=1\textwidth,height=\textheight]{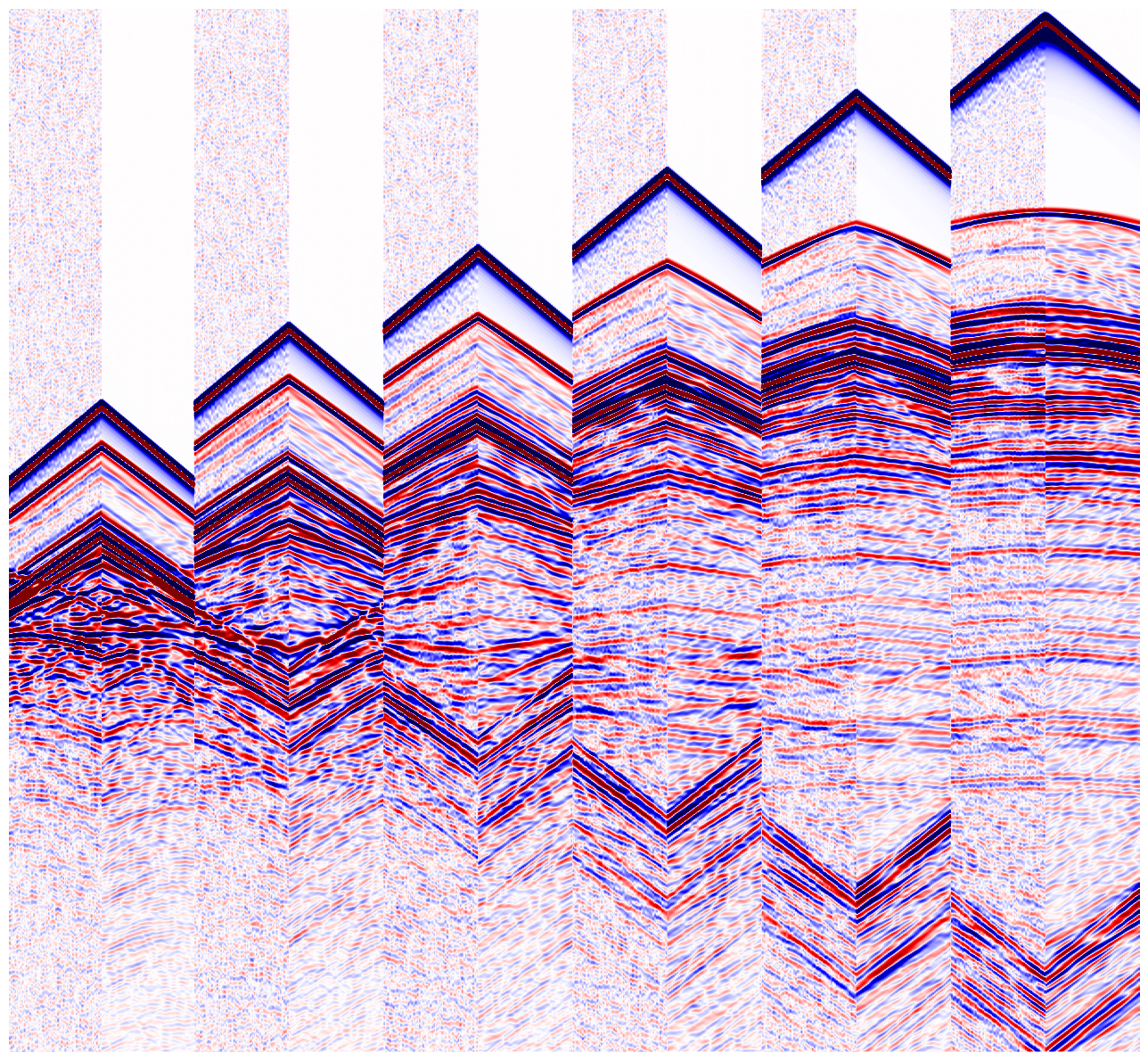}

}

\subcaption{\label{fig-misfit-1}ASPIRE 1 data fit \(14\%\)}

\end{minipage}%
\begin{minipage}{0.50\linewidth}

\centering{

\includegraphics[width=1\textwidth,height=\textheight]{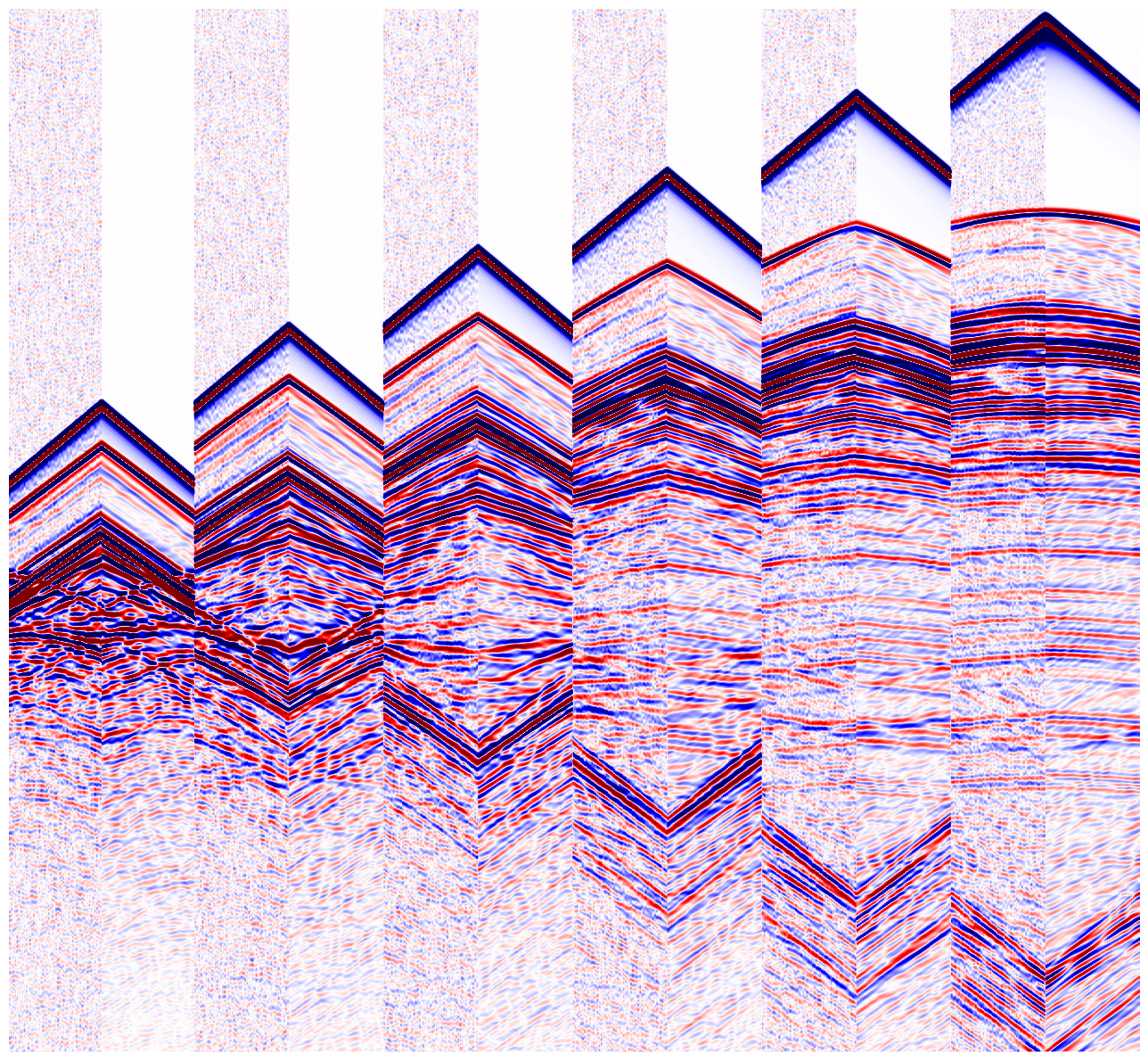}

}

\subcaption{\label{fig-misfit-2}ASPIRE 2 data fit \(17\%\)}

\end{minipage}%
\newline
\begin{minipage}{0.50\linewidth}

\end{minipage}%

\caption{\label{fig-aspire-datamisfit}Comparison of data misfit of a
shot record generated from the posterior samples for two ASPIRE
iterations by interweaving traces from the observed shot and simulated
shot gathers. The second ASPIRE iteration has improved the data fit by
using one more gradient (extended migration) at test time.}

\end{figure}

To ameliorate the amortization gap and to accommodate complexities
arising from salt in the Gulf of Mexico, we adapt and extend
methodologies introduced by \citet{orozco2024aspire} and retrain the
networks at the next iteration on migrations improved by the average of
the posterior samples produced by the current iteration. This type of
approach, known as ASPIRE \citep{orozco2024aspire}, which can be
interpreted as a probabilistic loop-unrolled gradient descent algorithm
\citep{putzky2017recurrent, gregor2010learning}, remains amortized and
has been demonstrated to close the amortization gap in transcranial
ultrasound brain imaging. Motivated by these findings, and by the fact
that ASPIRE lends itself well to iterative workflows, we propose a
methodology aimed at improving the large data residuals observed in
Figure~\ref{fig-misfit-1}. These residuals are mainly due to the
complexity of the velocity model and the lack of training data, a
situation where amortized methods are known to fail
\citep{siahkoohi2023reliable, orozco2024aspire}.

To remedy this situation and decrease the data misfit, we follow ASPIRE
and produce an improved migration-velocity model based on the average of
the posterior samples. With this improved velocity model, new CIGs
\(\overline{\mathbf{y}}_1\) are computed, as in Equation~\ref{eq-score},
but now with \(\mathbf{x}_{\textrm{mean}}\)---i.e., the posterior mean,
serving as the migration-velocity model instead of \(\mathbf{x}_0\).
After calculating CIGs for each training sample, we train a new network
on the updated dataset
\(\overline{\mathcal{D}}_1 = \{\mathbf{x}^{i},\overline{\mathbf{y}}_1^{i}\}_{i=0}^{N}\).
While this choice of posterior mean (e.g.~as opposed to posterior
median) is still open to debate, \citet{orozco2024aspire} proved that it
leads to significant improvements. Although there are additional offline
and online costs associated with this method, the inference stage
remains relatively computationally efficient with the cost of a single
migration per iteration and a low total number of iterations, in this
case two iterations.

\subsubsection{Experimental setup}\label{experimental-setup}

To demonstrate how inference with ASPIRE can be adapted to complex
settings, we consider 2D slices through the 3D SEAM model
\citep{fehler2011model}, which represents complex salt geometry typical
for the Gulf of Mexico. We create a training split of the SEAM model by
selecting a continuous subset of crosslines for training. To avoid
similarity between 2D slices, we skip \(2\) km between the nearest
training slice and the nearest test slice.

When simulating shot gathers, we generate narrow-offset 2D seismic lines
with a grid size of \(1744 \times 512\) and a spatial discretization of
\(20\) m. A marine dataset is created by simulating shot data with
sources and receivers towed at the same depth near the surface.
Surface-related multiples are avoided by applying an absorbing boundary
condition at the surface. Computational costs are reduced by using a
coarse source interval of \(1000\) m, while receivers are sampled at
\(20\) m and a maximum offset of \(6\) km. The total recording time for
each shot record is \(9\) s, and Gaussian noise is added to yield a
level of \(25\) dB in a frequency band that matches the source
signature. With these settings, \(32\) shots were simulated for a model
with constant density and varying velocity.

Migration-velocity models are created by first removing the salt from
the ground truth velocity models, followed by smoothing the result with
a Gaussian kernel of grid size \(25\) (see
Figure~\ref{fig-aspire-models}). To capture most of the energy, the
horizontal subsurface-offset migration utilizes \(50\) equally spaced
offsets, ranging from \(-2000\) m to \(2000\) m. After imaging the full
seismic line, we extract subsets to create the training pairs by taking
sliding windows of each 2D line, resulting in grids of size
\(512 \times 512\) and a total of \(700\) training data pairs. Each
ASPIRE iteration takes \(12\) GPU hours to train. While the conditional
Diffusion network is trained on these smaller patches, at inference time
the trained network is evaluated on the full grid size of
(\(512 \times 1744\)). This is feasible because our Diffusion network
predominantly relies on convolutional layers, allowing for evaluation on
grid sizes larger than those used during training. At inference, we
incur the cost of one extended migration (\(3\) min) per ASPIRE
iteration and then \(8\) sec per posterior sample.

\subsubsection{Algorithmic innovations}\label{algorithmic-innovations}

To successfully adapt ASPIRE to the complex setting of SEAM, the
following innovations are implemented:

\begin{itemize}
\item
  \textbf{Utilization of previous iterates.} In its vanilla
  implementation, gradient descent and, therefore ASPIRE, base its
  updates on gradients taken at the current model iterate (read inferred
  velocity model), ignoring information from previous iterates.
  Motivated by quasi-Newton optimization methods,
  \citet{putzky2017recurrent} proposed to add previous model iterates to
  improve convergence and approach reminiscent of quasi-Newton methods
  where previous gradients are used to approximate the Hessian.
  Similarly, we allow the networks at each iteration to take all
  previously calculated CIGs and inferred migration-velocity models as
  input. In our experiments, including this modification significantly
  improved the performance of ASPIRE.
\item
  \textbf{Incorporation of domain knowledge with salt flooding.} To
  detect the bottom of the salt, we employ the well-established method
  of salt flooding (see \citet{esser2016constrained} and the references
  therein) during which top salt is extended downwards. Due to its
  iterative structure, ASPIRE can easily accommodate salt flooding, so
  the bottom of the salt is clearly delineated by migration on the
  second iteration.
\end{itemize}

\subsubsection{Preliminary results}\label{preliminary-results}

\begin{figure}

\begin{minipage}{0.50\linewidth}

\centering{

\includegraphics[width=1\textwidth,height=\textheight]{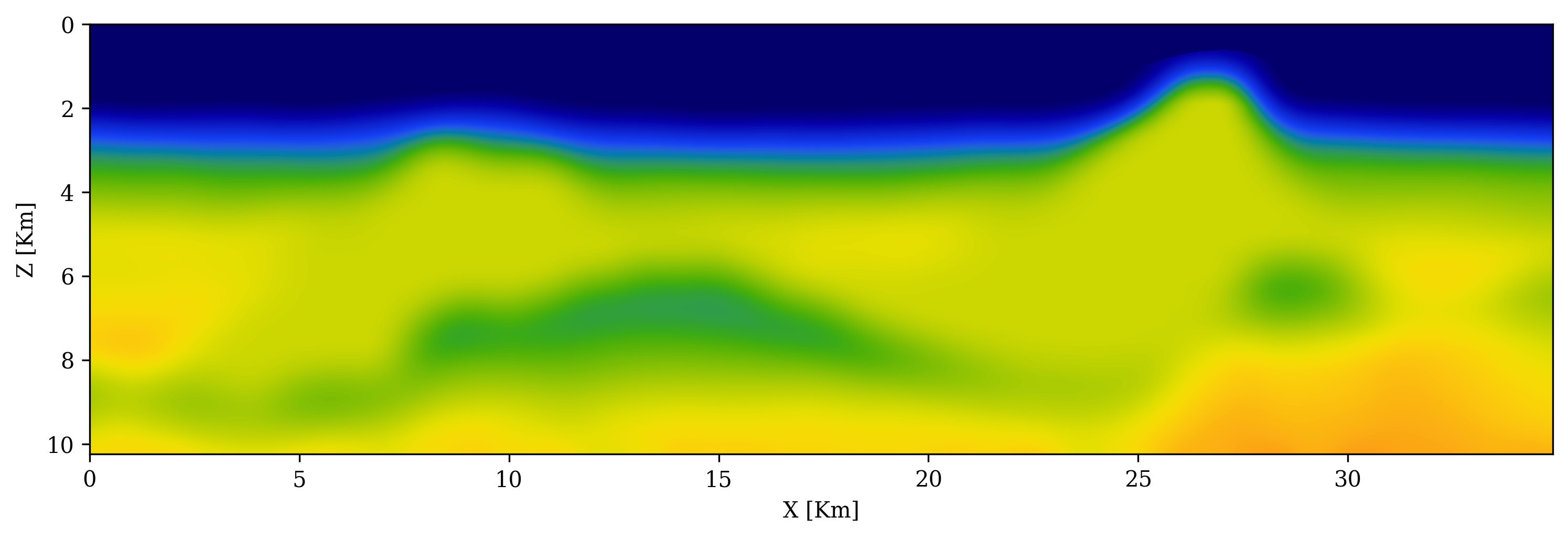}

}

\subcaption{\label{fig-seam-a}Initial velocity model}

\end{minipage}%
\begin{minipage}{0.50\linewidth}

\centering{

\includegraphics[width=1\textwidth,height=\textheight]{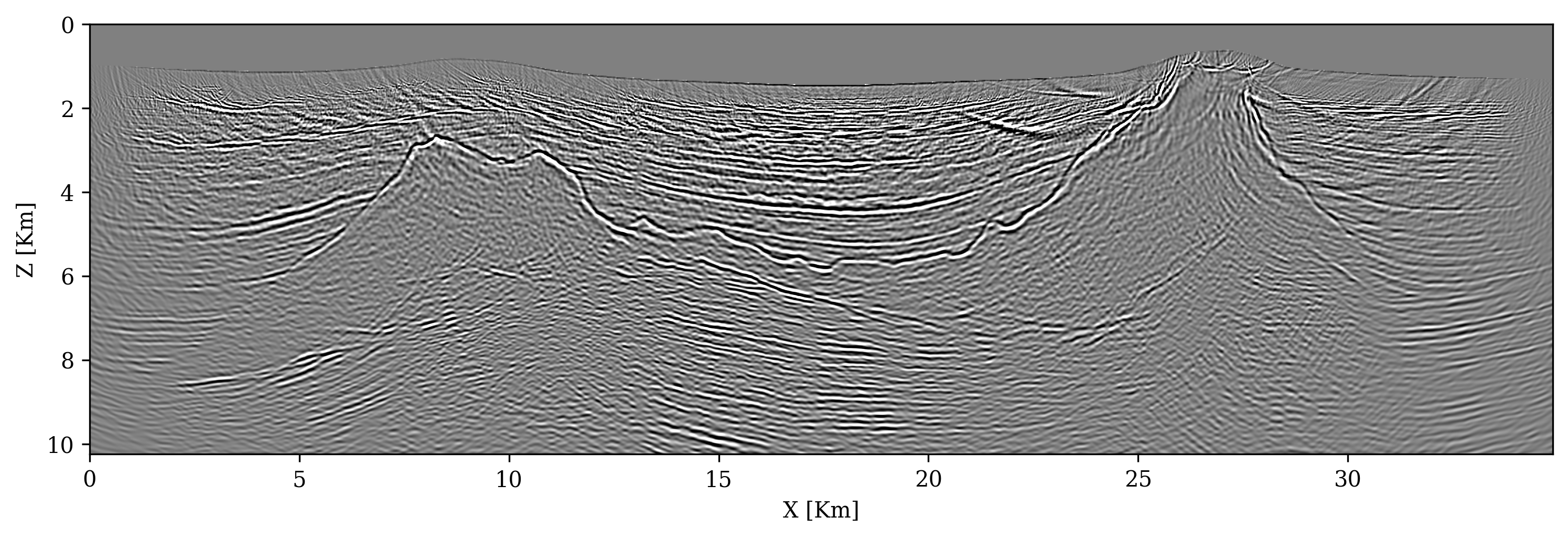}

}

\subcaption{\label{fig-seam-b}Initial migration}

\end{minipage}%
\newline
\begin{minipage}{0.50\linewidth}

\centering{

\includegraphics[width=1\textwidth,height=\textheight]{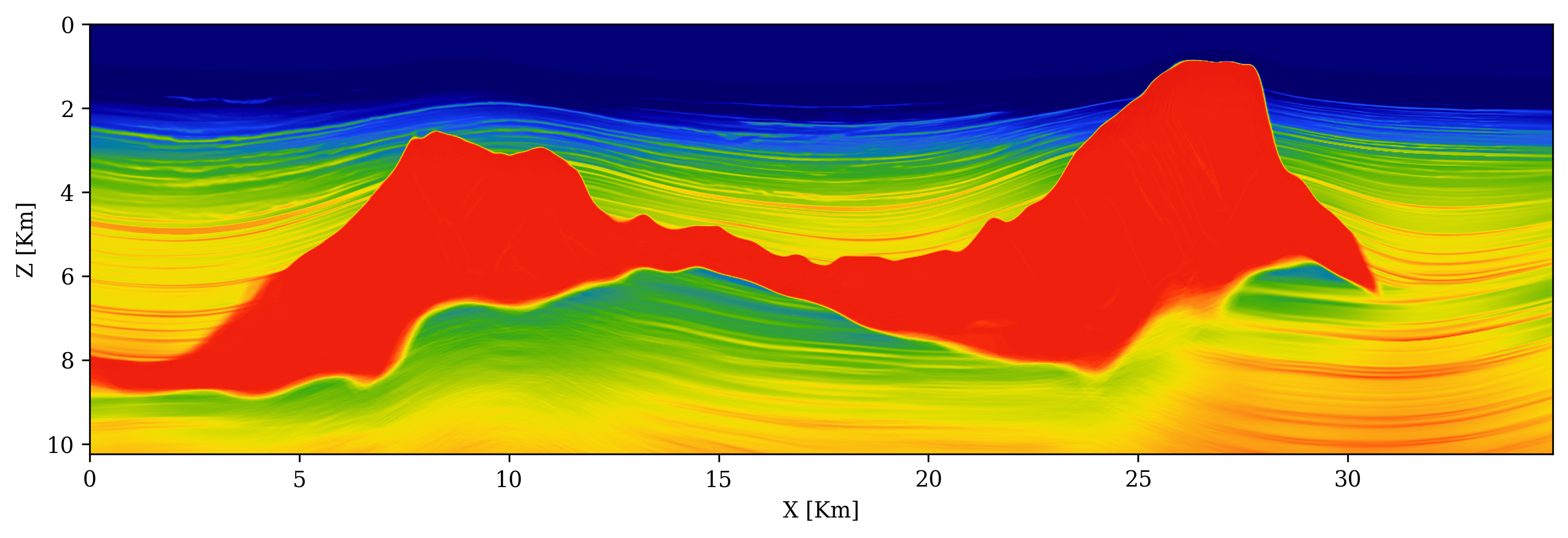}

}

\subcaption{\label{fig-seam-c}ASPIRE 1 model SSIM \(=0.70\)}

\end{minipage}%
\begin{minipage}{0.50\linewidth}

\centering{

\includegraphics[width=1\textwidth,height=\textheight]{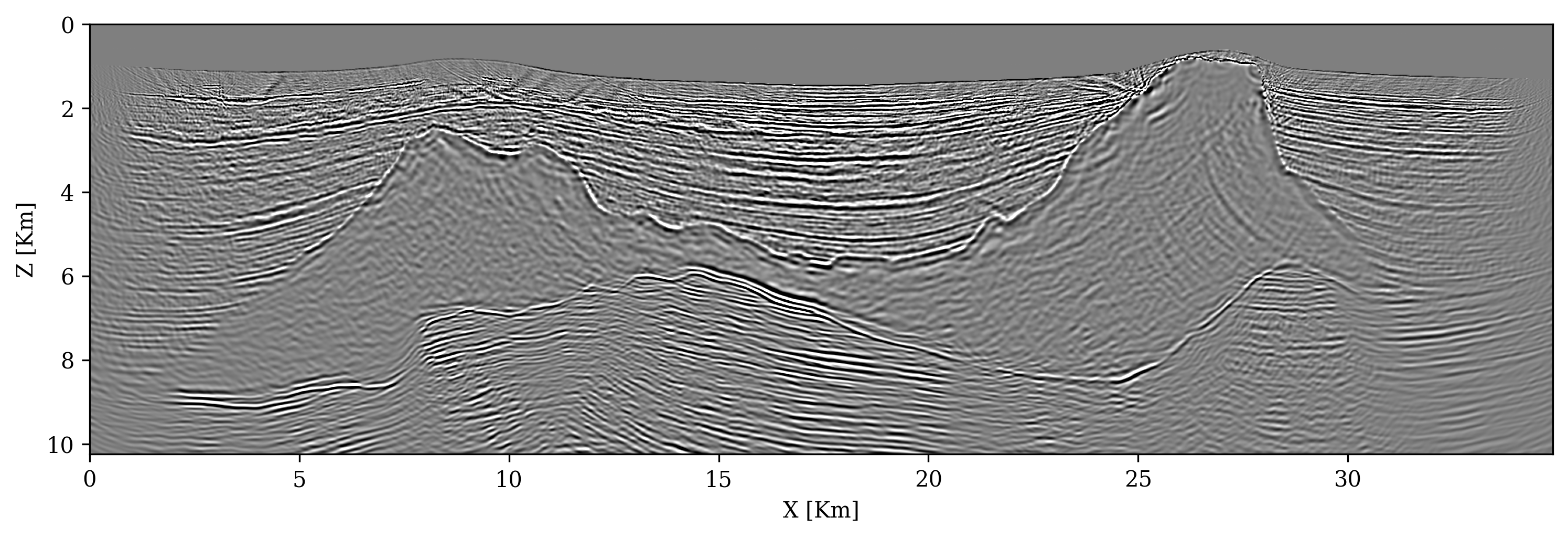}

}

\subcaption{\label{fig-seam-d}ASPIRE 1 migration}

\end{minipage}%
\newline
\begin{minipage}{0.50\linewidth}

\centering{

\includegraphics[width=1\textwidth,height=\textheight]{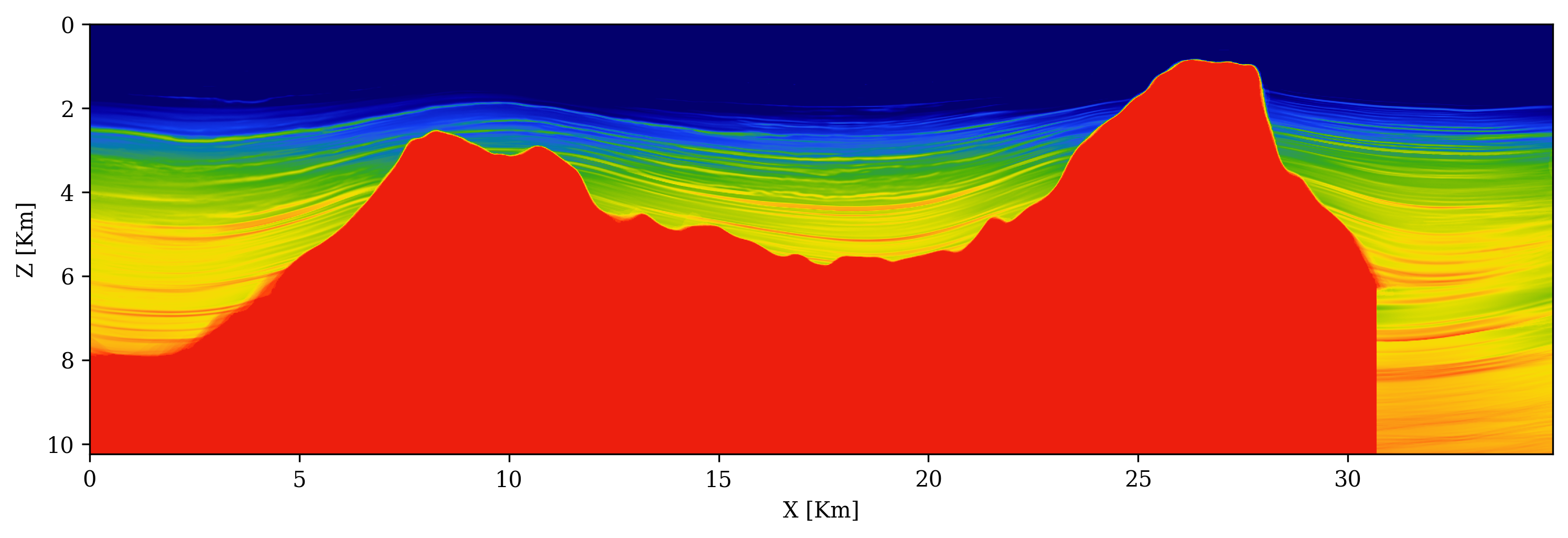}

}

\subcaption{\label{fig-seam-e}ASPIRE 1 velocity model w/ flooding}

\end{minipage}%
\begin{minipage}{0.50\linewidth}

\centering{

\includegraphics[width=1\textwidth,height=\textheight]{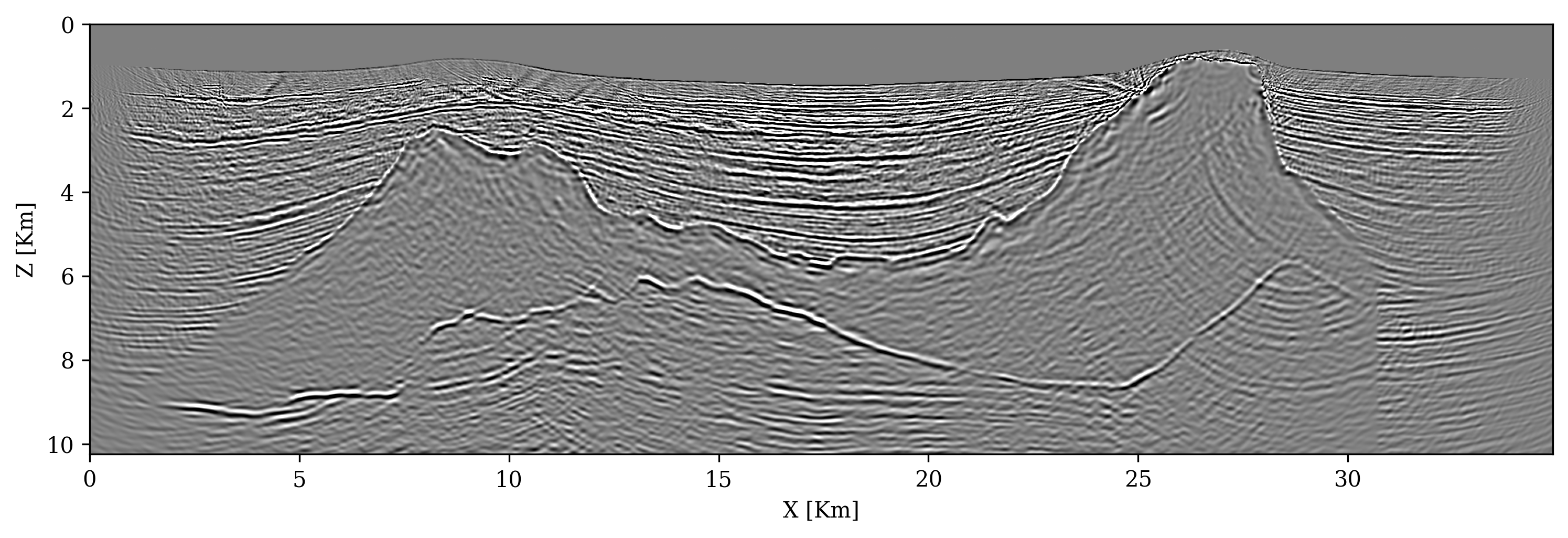}

}

\subcaption{\label{fig-seam-f}ASPIRE 1 migration w/ flooding}

\end{minipage}%
\newline
\begin{minipage}{0.50\linewidth}

\centering{

\includegraphics[width=1\textwidth,height=\textheight]{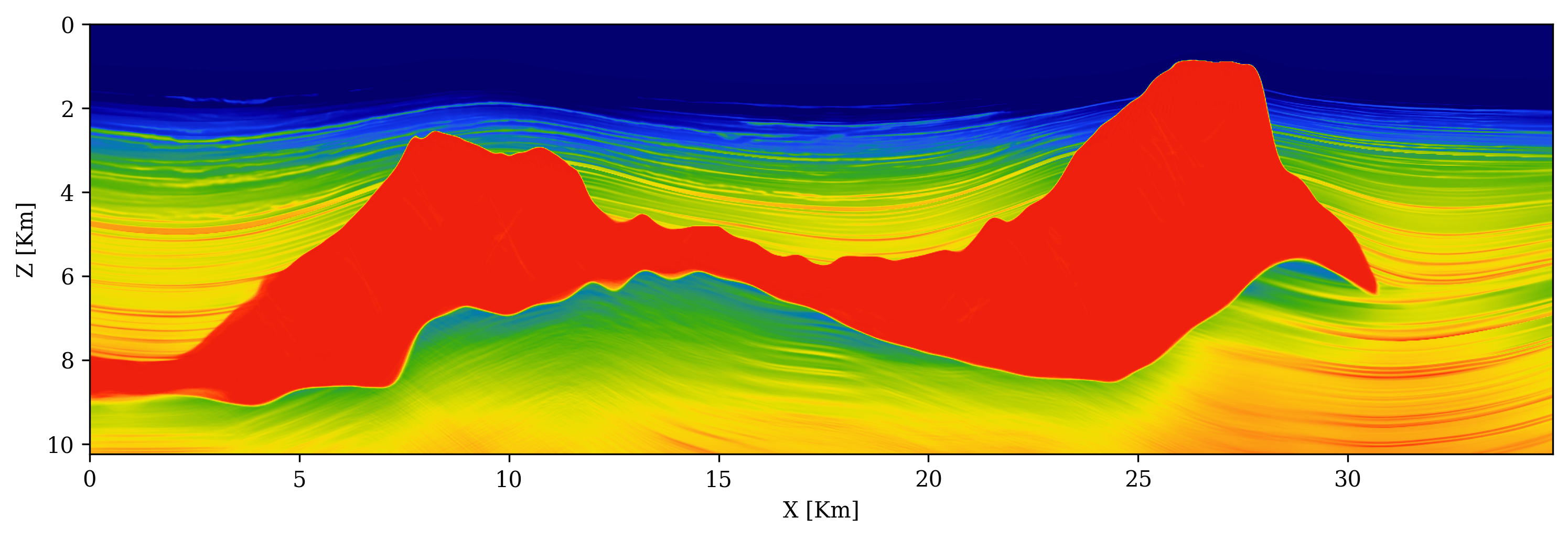}

}

\subcaption{\label{fig-seam-g}ASPIRE 2 model SSIM \(=0.72\)}

\end{minipage}%
\begin{minipage}{0.50\linewidth}

\centering{

\includegraphics[width=1\textwidth,height=\textheight]{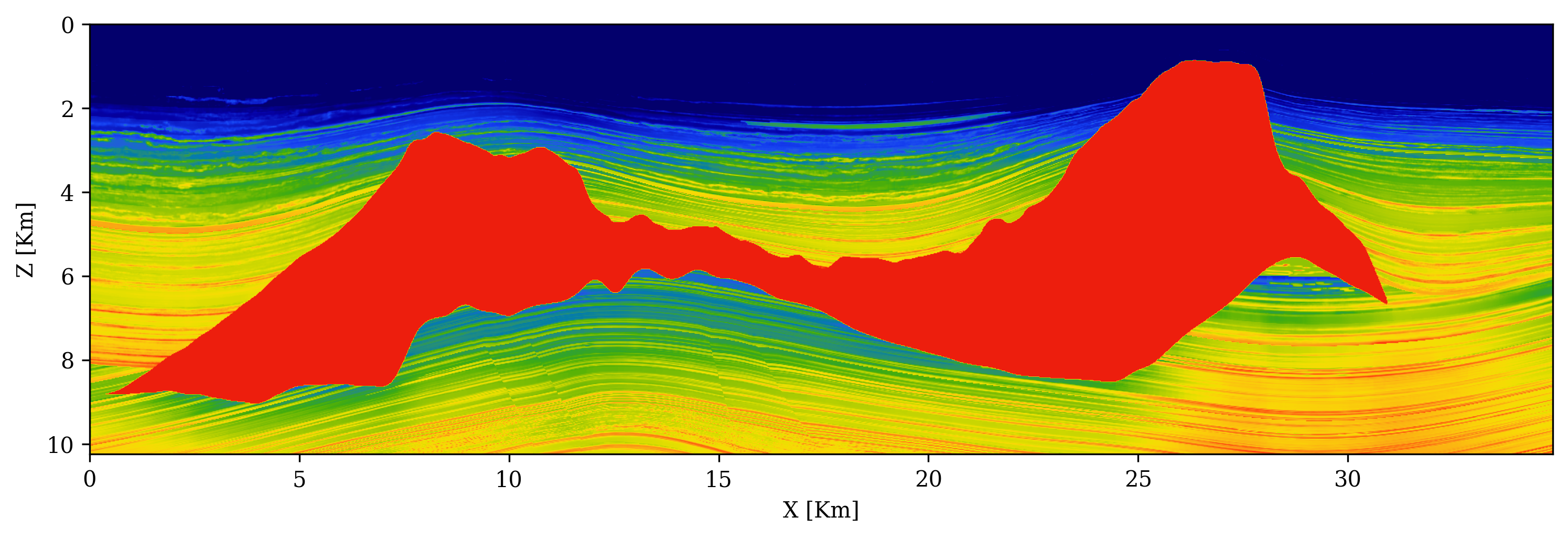}

}

\subcaption{\label{fig-seam-i}Ground truth velocity model}

\end{minipage}%

\caption{\label{fig-aspire-models}Comparison of posterior means yielded
by two iterations of ASPIRE and their corresponding reverse-time
migrations.}

\end{figure}

In Figure~\ref{fig-aspire-models}, we present the results from two
iterations of ASPIRE, starting from the smoothed migration-velocity
model depicted in Figure~\ref{fig-seam-a}. This migration-velocity model
is used to produce the initial CIGs, whose zero-offset section is
displayed in Figure~\ref{fig-seam-b}. Based on these CIGs, the first
trained ASPIRE produces posterior samples, whose average is plotted in
Figure~\ref{fig-seam-c}. While the top of salt is well recovered,
comparison with the ground truth velocity model depicted in
Figure~\ref{fig-seam-i} shows that important details are missing in the
bottom salt. This, in turn, leads to poorly resolved bottom salt in the
zero-offset migrated section Figure~\ref{fig-seam-d}. By flooding the
salt downwards, as depicted in Figure~\ref{fig-seam-e}, this issue is
largely mitigated, as observed from the migration included in
Figure~\ref{fig-seam-f}, where the bottom salt is delineated sharply and
includes the main topological features. This means that the second
trained ASPIRE on salt-flooded migrations can correctly infer the salt
bottom as shown in Figure~\ref{fig-seam-g}. Before comparing the
migration in this model to a baseline derived from the true model, let
us first consider the contraction of the inferred uncertainties
resulting from the ASPIRE iterations.

\begin{figure}

\begin{minipage}{0.50\linewidth}

\centering{

\includegraphics[width=1\textwidth,height=\textheight]{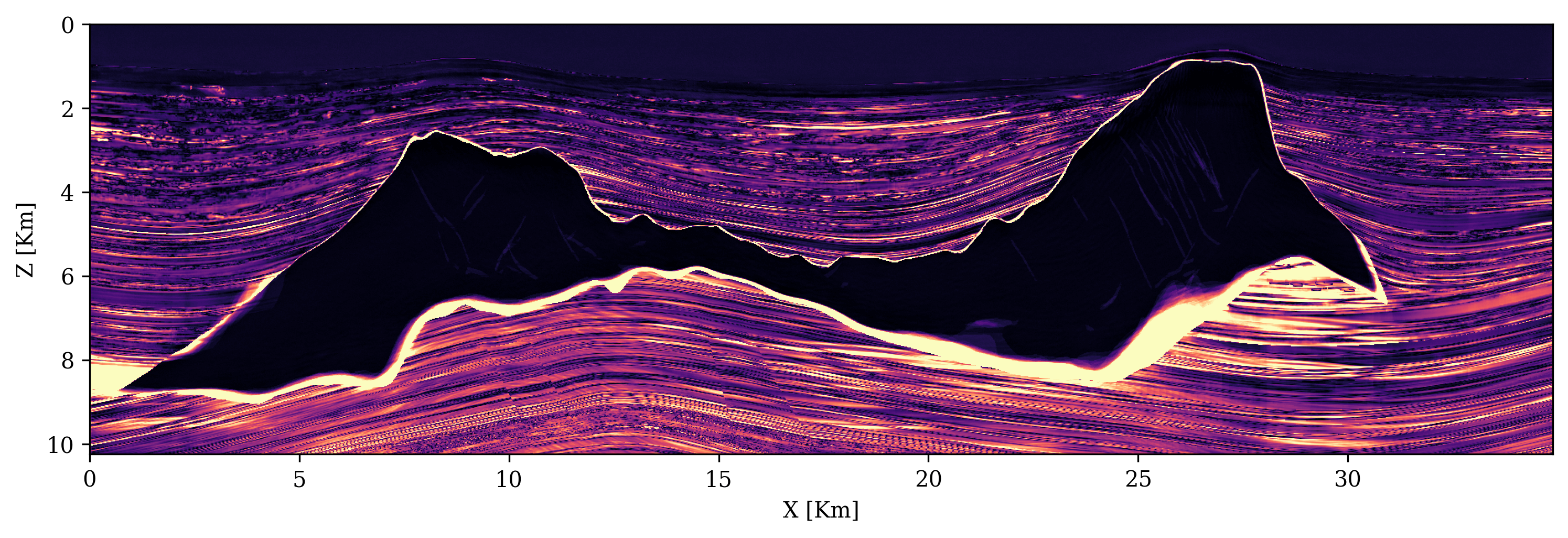}

}

\subcaption{\label{fig-aspire-error-1}ASPIRE 1 Error RMSE \(=0.26\)}

\end{minipage}%
\begin{minipage}{0.50\linewidth}

\centering{

\includegraphics[width=1\textwidth,height=\textheight]{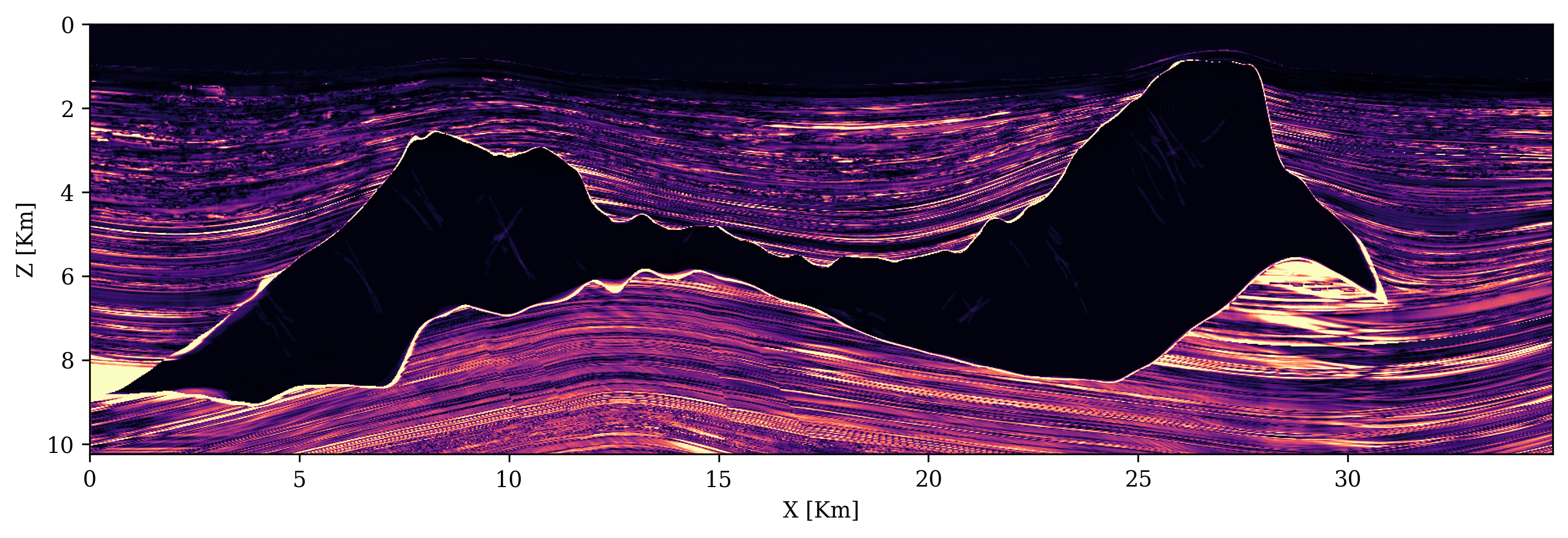}

}

\subcaption{\label{fig-aspire-error-2}ASPIRE 2 Error RMSE\(=0.20\)}

\end{minipage}%
\newline
\begin{minipage}{0.50\linewidth}

\centering{

\includegraphics[width=1\textwidth,height=\textheight]{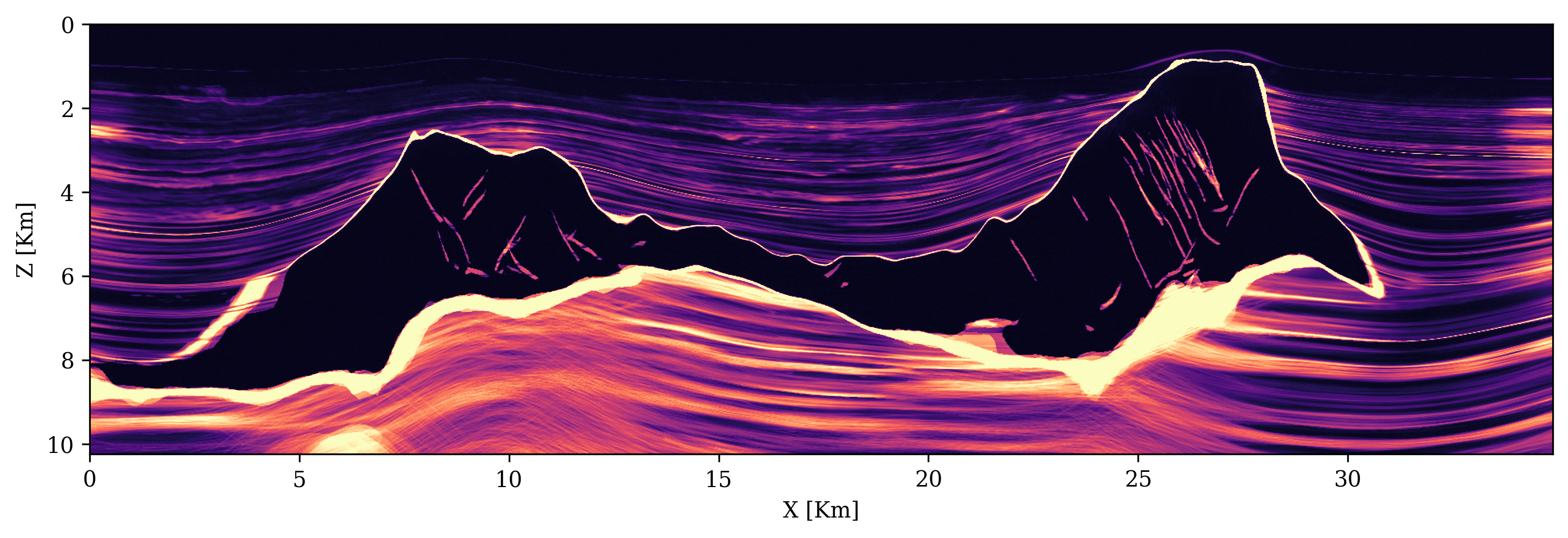}

}

\subcaption{\label{fig-aspire-std-1}ASPIRE 1 standard deviation}

\end{minipage}%
\begin{minipage}{0.50\linewidth}

\centering{

\includegraphics[width=1\textwidth,height=\textheight]{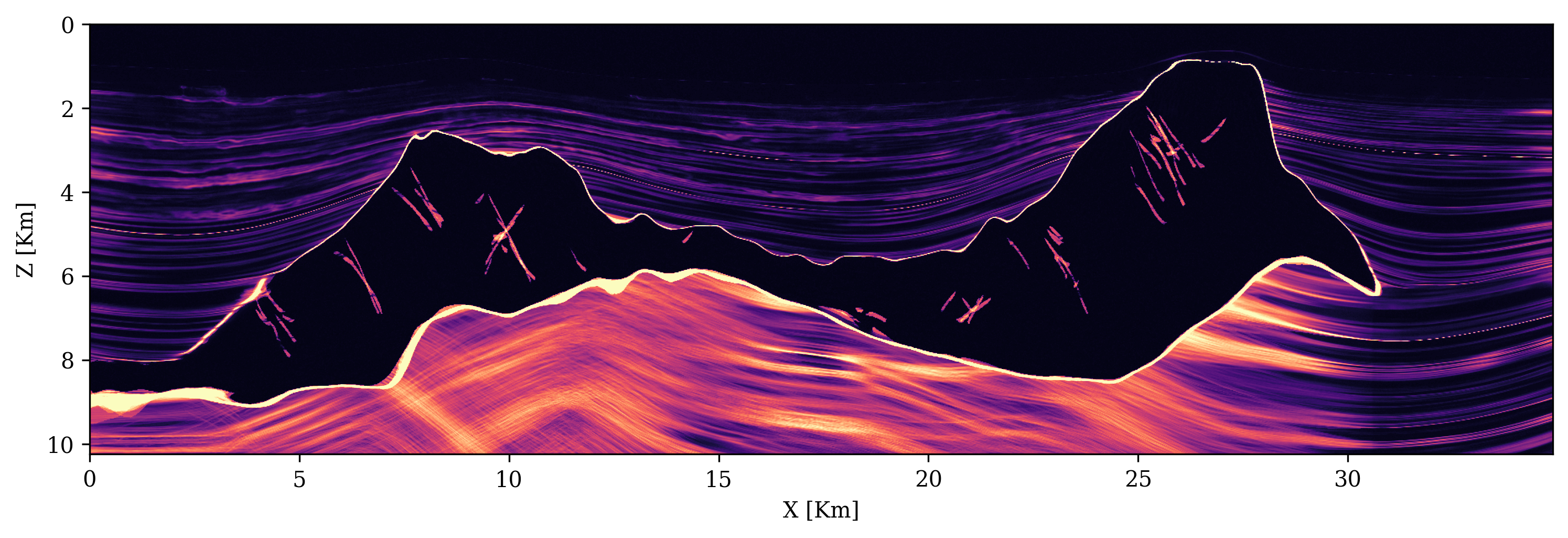}

}

\subcaption{\label{fig-aspire-std-2}ASPIRE 2 standard deviation}

\end{minipage}%
\newline
\begin{minipage}{0.50\linewidth}

\centering{

\includegraphics[width=1\textwidth,height=\textheight]{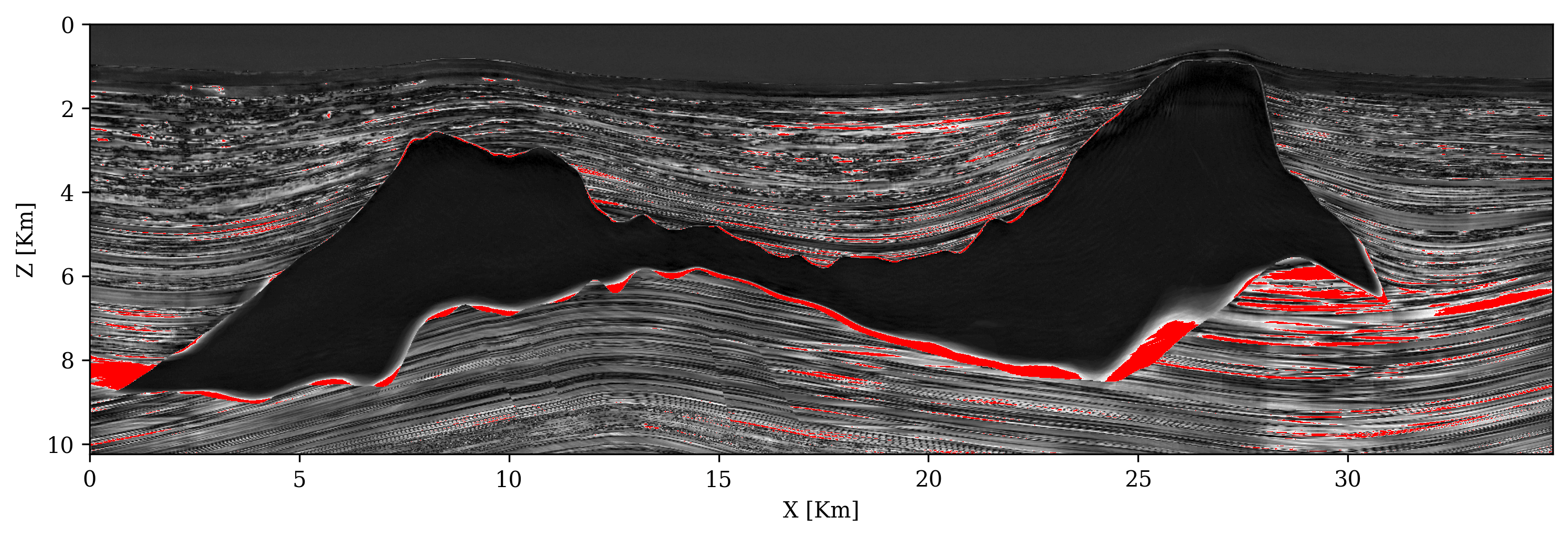}

}

\subcaption{\label{fig-aspire-support-1}ASPIRE 1 \(z\)-score
\(=3.44\%\)}

\end{minipage}%
\begin{minipage}{0.50\linewidth}

\centering{

\includegraphics[width=1\textwidth,height=\textheight]{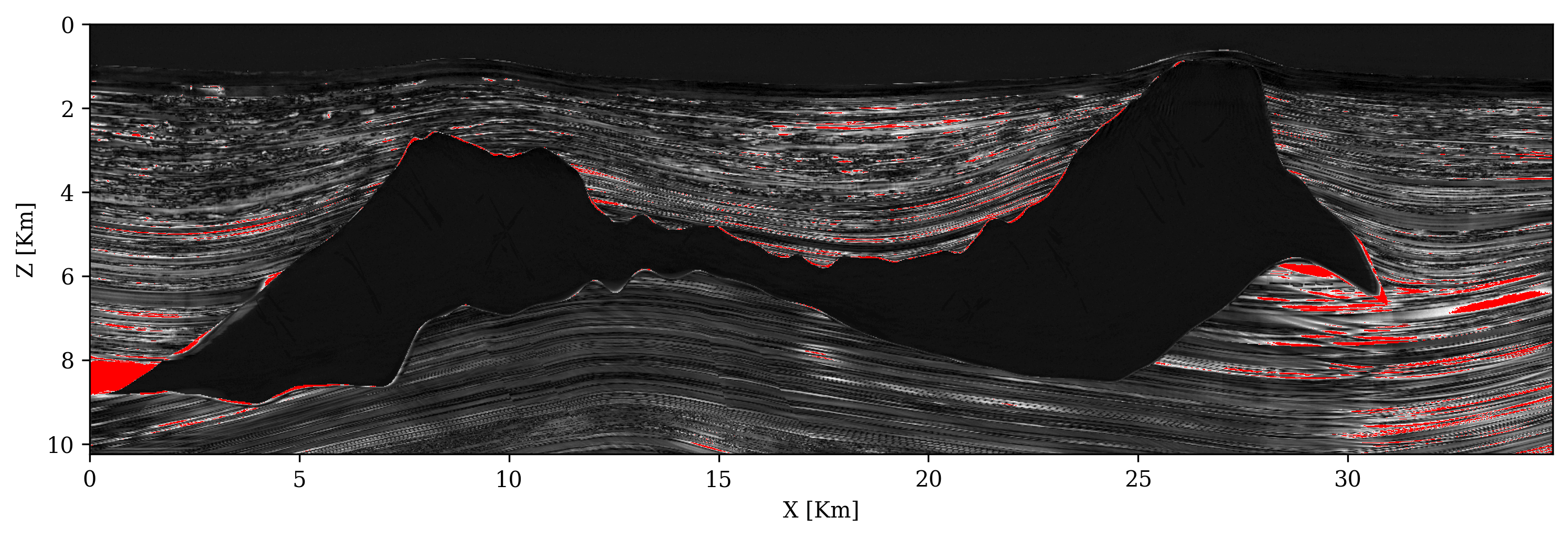}

}

\subcaption{\label{fig-aspire-support-2}ASPIRE 2 \(z\)-score
\(=2.03\%\)}

\end{minipage}%
\newline
\begin{minipage}{0.50\linewidth}

\centering{

\includegraphics[width=1\textwidth,height=\textheight]{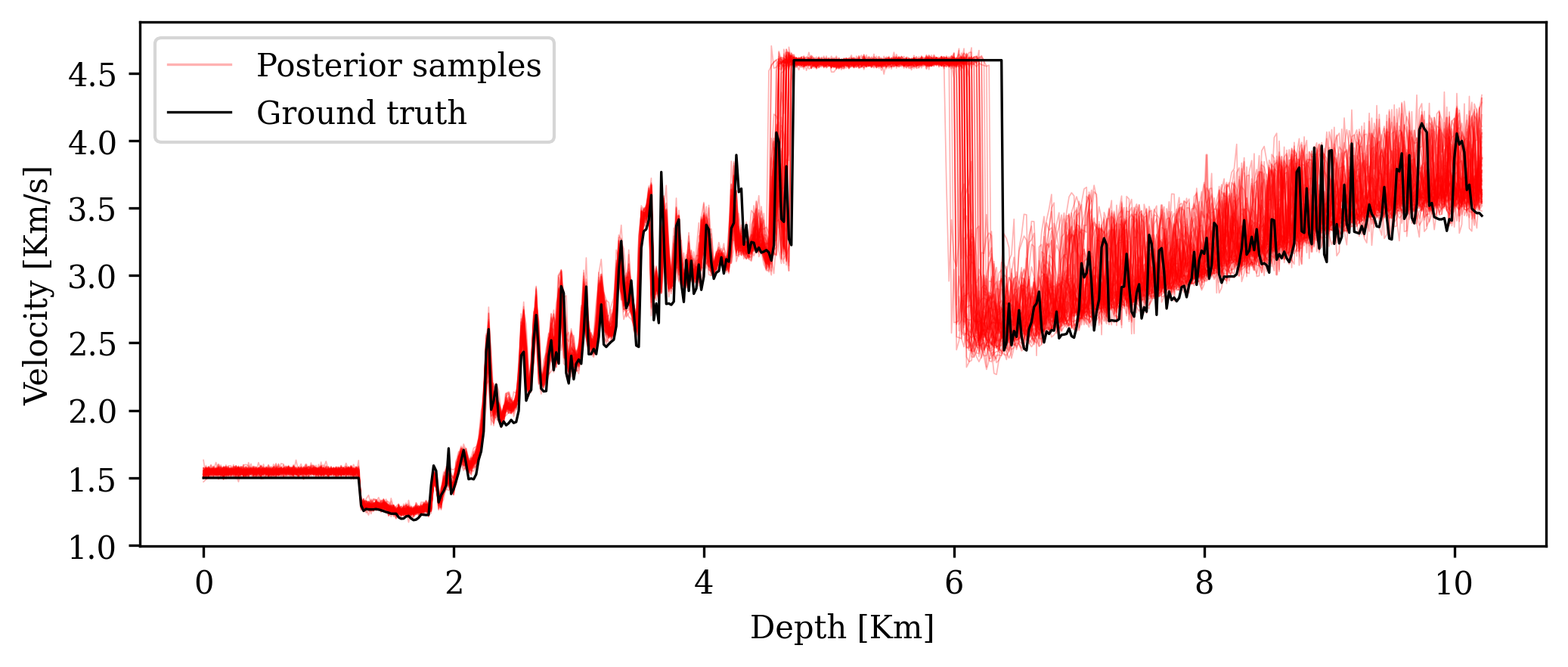}

}

\subcaption{\label{fig-aspire-traces-1}ASPIRE 1 traces coverage
\(=61.79\%\)}

\end{minipage}%
\begin{minipage}{0.50\linewidth}

\centering{

\includegraphics[width=1\textwidth,height=\textheight]{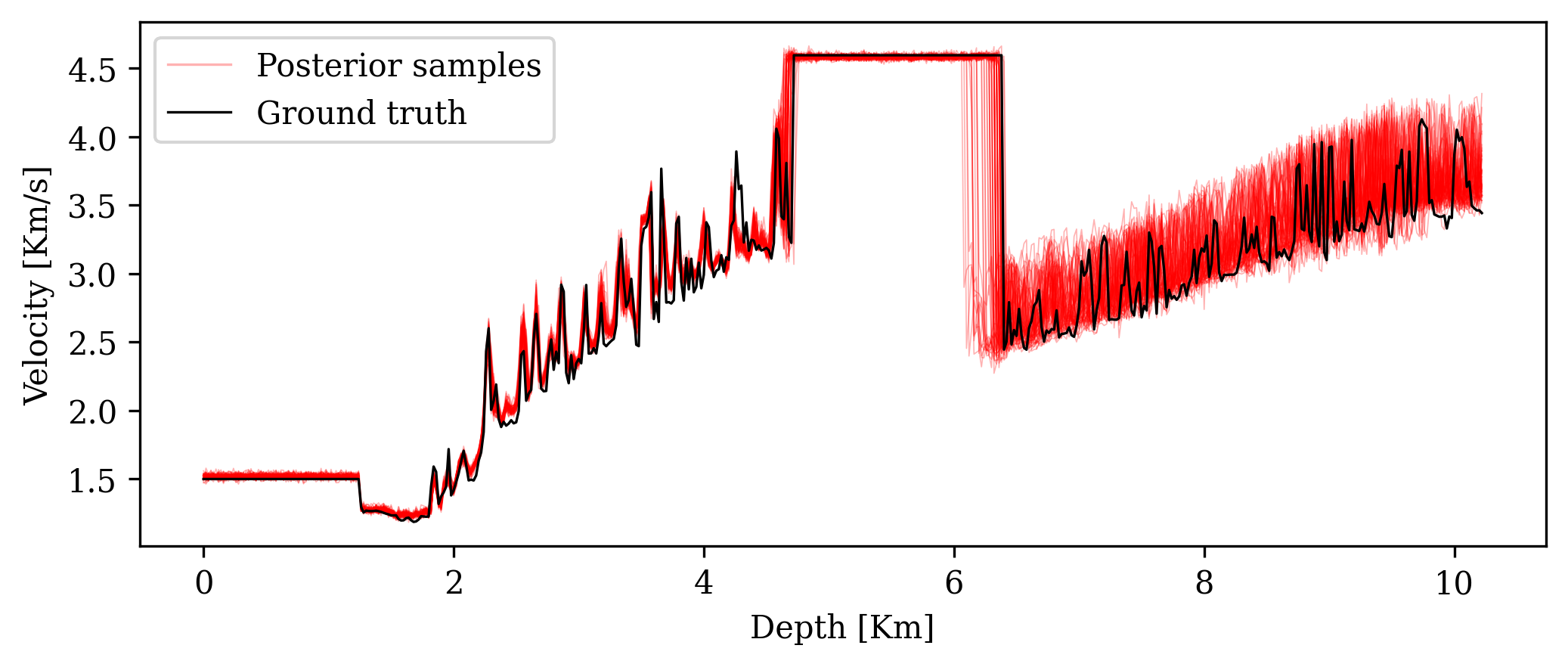}

}

\subcaption{\label{fig-aspire-traces-2}ASPIRE 2 traces coverage
\(=70.24\%\)}

\end{minipage}%
\newline
\begin{minipage}{0.50\linewidth}

\end{minipage}%

\caption{\label{fig-aspire-uq}Comparison between recovery and UQ quality
yielded by ASPIRE 1 and 2.}

\end{figure}

In addition to providing increasingly better estimates for the velocity
model, ASPIRE also includes UQ at each iteration. In
Figure~\ref{fig-aspire-uq}, different aspects of the inferred
uncertainties for ASPIRE 1 and 2 are illustrated by means of plots for
errors included in figures \ref{fig-aspire-error-1} and
\ref{fig-aspire-error-2}; posterior standard deviation in figures
\ref{fig-aspire-std-1} and \ref{fig-aspire-std-2}; \(z\)-score in
figures \ref{fig-aspire-support-1} and \ref{fig-aspire-support-2}; and
finally plots for the coverage in figures \ref{fig-aspire-traces-1} and
\ref{fig-aspire-traces-2}. From these plots, the following observations
can be made: First, errors with respect to the ground truth velocity
model at the bottom salt are greatly reduced by ASPIRE 2 due to the salt
flooding. However, errors remain at both ends of the salt due to a lack
of illumination. As expected, errors remain in the sediments below the
salt. Second, as expected, predicted uncertainties at the bottom salt
are reduced for ASPIRE 2 (see figures \ref{fig-aspire-std-1} and
\ref{fig-aspire-std-2}). The uncertainty also correlates reasonably well
with the errors. We also observe the appearance of dirty salt, which can
be explained by the fact that the salt in the SEAM dataset includes
dirty salt. Third, with very few exceptions there is a drastic
improvement in the overall \(z\)-score plots (figures
\ref{fig-aspire-support-1} and \ref{fig-aspire-support-2}) for ASPIRE 2.
Most notably, \(z\)-scores improve significantly at the bottom-salt
interface and within the sedimentary layering under the salt. Finally,
the coverage of the posterior samples is also improved. While the ground
truth bottom salt was missed in ASPIRE 1, the posterior samples yielded
by ASPIRE 2 clearly contain the step-out of the salt.

Moreover, visual improvements in the second ASPIRE iteration are
supported by quantitative uncertainty quality measures listed in the
captions of Figure~\ref{fig-aspire-uq} and in
Table~\ref{tbl-performance-aspire}, summarizing performance over eight
unseen test 2D lines. Overall, the progression of results correlates
well with the iterative focus of ASPIRE: the first iteration targets the
top of the salt and the second iteration targets the bottom of the salt.
The corresponding shot gathers for the different ASPIRE iterations are
included in Figure~\ref{fig-aspire-datamisfit} and confirm that the
model improvements lead to improved data fit.

\begin{longtable}[]{@{}
  >{\raggedright\arraybackslash}p{(\columnwidth - 12\tabcolsep) * \real{0.1833}}
  >{\raggedright\arraybackslash}p{(\columnwidth - 12\tabcolsep) * \real{0.1667}}
  >{\raggedright\arraybackslash}p{(\columnwidth - 12\tabcolsep) * \real{0.1500}}
  >{\raggedright\arraybackslash}p{(\columnwidth - 12\tabcolsep) * \real{0.1333}}
  >{\raggedright\arraybackslash}p{(\columnwidth - 12\tabcolsep) * \real{0.1333}}
  >{\raggedright\arraybackslash}p{(\columnwidth - 12\tabcolsep) * \real{0.1167}}
  >{\raggedright\arraybackslash}p{(\columnwidth - 12\tabcolsep) * \real{0.1167}}@{}}
\caption{Image and uncertainty quality metrics on SEAM dataset ASPIRE
iterations.}\label{tbl-performance-aspire}\tabularnewline
\toprule\noalign{}
\begin{minipage}[b]{\linewidth}\raggedright
Dataset
\end{minipage} & \begin{minipage}[b]{\linewidth}\raggedright
RMSE \(\downarrow\)
\end{minipage} & \begin{minipage}[b]{\linewidth}\raggedright
SSIM \(\uparrow\)
\end{minipage} & \begin{minipage}[b]{\linewidth}\raggedright
Coverage \([\%]\) \(\uparrow\)
\end{minipage} & \begin{minipage}[b]{\linewidth}\raggedright
UCE \(\downarrow\)
\end{minipage} & \begin{minipage}[b]{\linewidth}\raggedright
\(z\)-score \([\%]\) \(\downarrow\)
\end{minipage} & \begin{minipage}[b]{\linewidth}\raggedright
Data fit \([\%]\) \(\uparrow\)
\end{minipage} \\
\midrule\noalign{}
\endfirsthead
\toprule\noalign{}
\begin{minipage}[b]{\linewidth}\raggedright
Dataset
\end{minipage} & \begin{minipage}[b]{\linewidth}\raggedright
RMSE \(\downarrow\)
\end{minipage} & \begin{minipage}[b]{\linewidth}\raggedright
SSIM \(\uparrow\)
\end{minipage} & \begin{minipage}[b]{\linewidth}\raggedright
Coverage \([\%]\) \(\uparrow\)
\end{minipage} & \begin{minipage}[b]{\linewidth}\raggedright
UCE \(\downarrow\)
\end{minipage} & \begin{minipage}[b]{\linewidth}\raggedright
\(z\)-score \([\%]\) \(\downarrow\)
\end{minipage} & \begin{minipage}[b]{\linewidth}\raggedright
Data fit \([\%]\) \(\uparrow\)
\end{minipage} \\
\midrule\noalign{}
\endhead
\bottomrule\noalign{}
\endlastfoot
ASPIRE 1 & \(0.25\) & \(0.71\) & \(66.6\) & \(0.043\) &
\(\mathbf{3.44}\) & \(13.3\) \\
ASPIRE 2 & \(\mathbf{0.21}\) & \(\mathbf{0.73}\) & \(\mathbf{67.3}\) &
\(\mathbf{0.042}\) & \(3.96\) & \(\mathbf{15.5}\) \\
\end{longtable}

\subsubsection{Quality assessment}\label{quality-assessment}

\begin{figure}

\centering{

\centering{

\includegraphics[width=1\textwidth,height=\textheight]{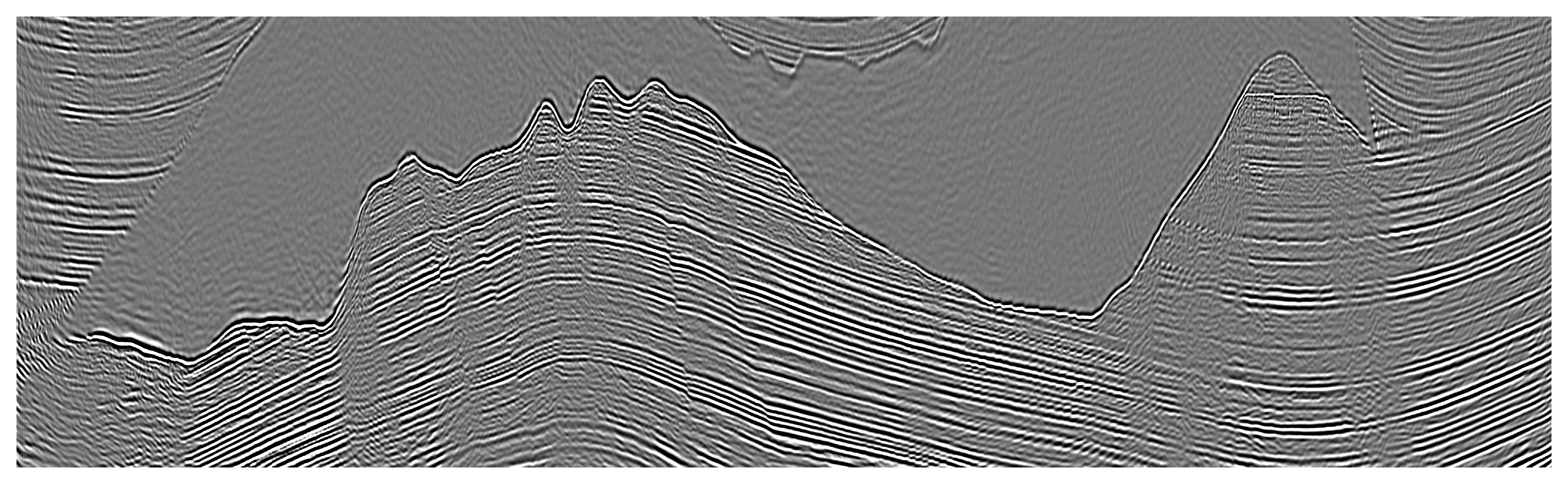}

}

\subcaption{\label{fig-downstream-rtm-gt}Migration in ground truth
velocity model}

\centering{

\includegraphics[width=1\textwidth,height=\textheight]{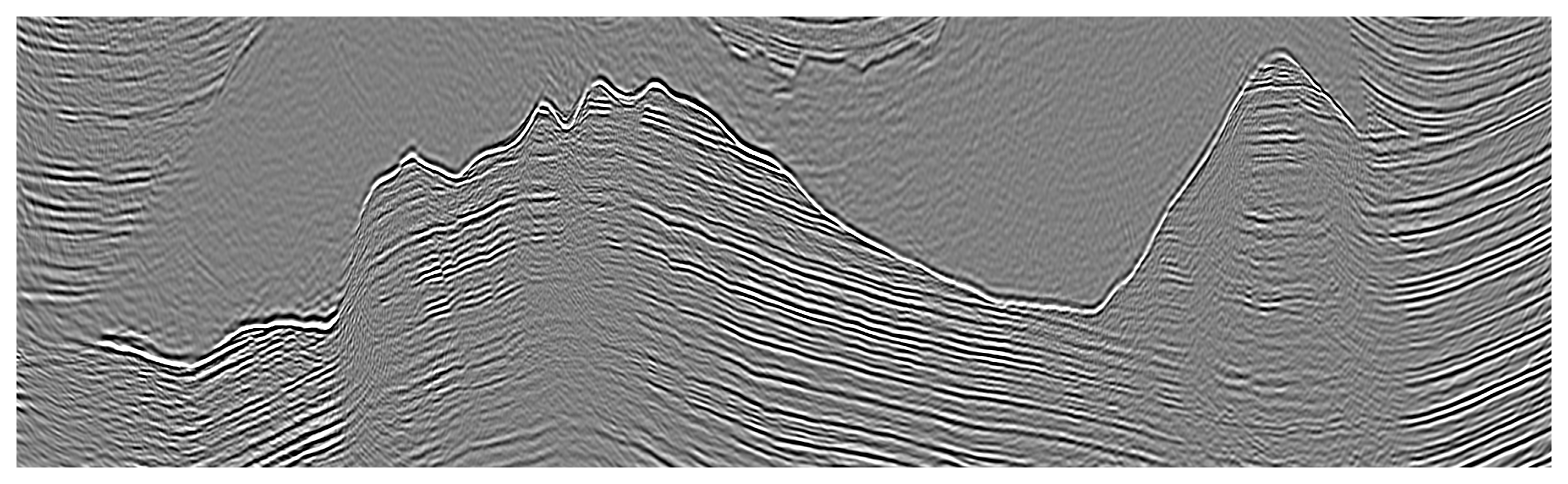}

}

\subcaption{\label{fig-downstream-rtm-v3}Mean of migrations in velocity
models produced by ASPIRE 2}

\centering{

\includegraphics[width=1\textwidth,height=\textheight]{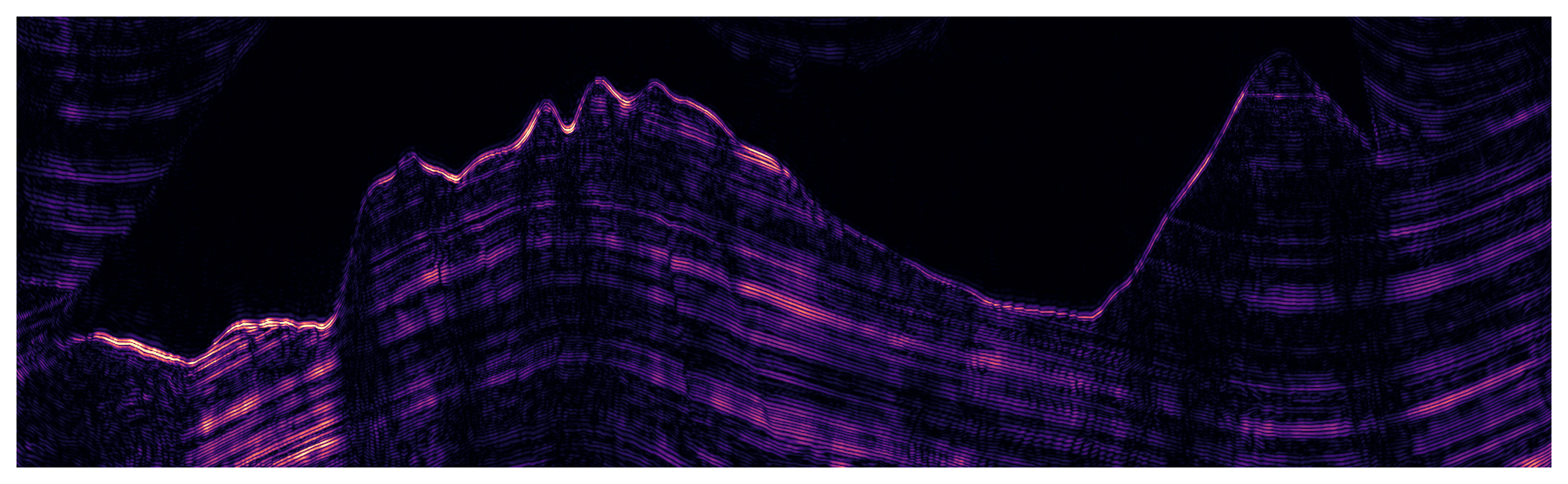}

}

\subcaption{\label{fig-downstream-error}Error between mean of migrations
in velocity models produced by ASPIRE 2 and ground truth migration}

\centering{

\includegraphics[width=1\textwidth,height=\textheight]{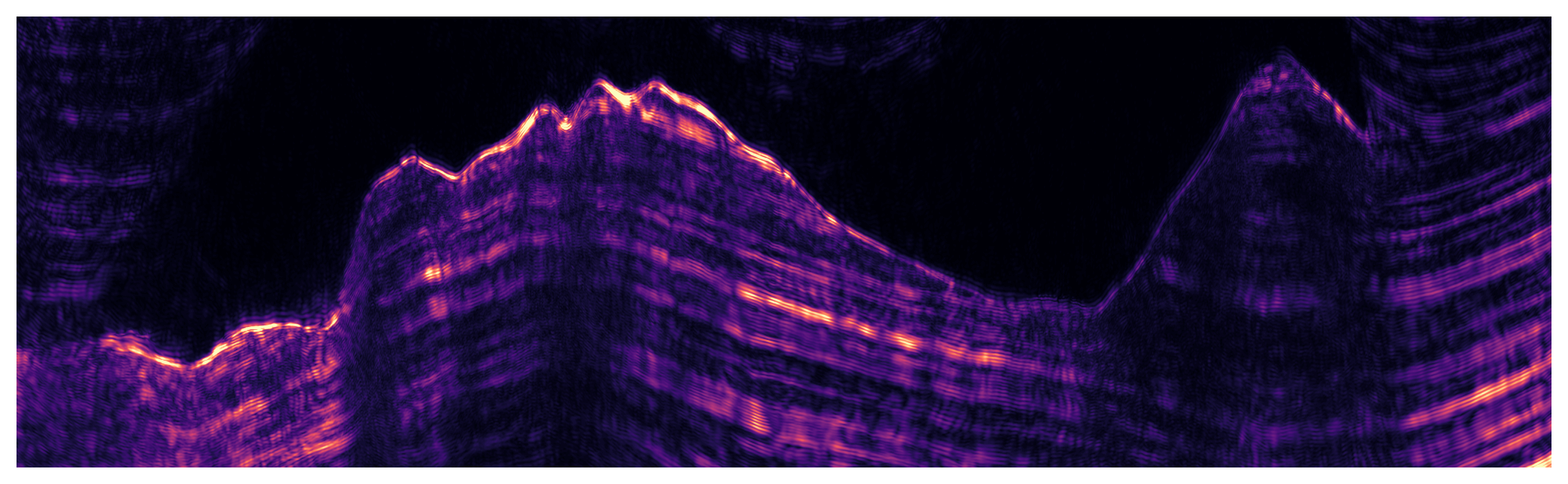}

}

\subcaption{\label{fig-downstream-std}Standard deviation of migrations
in velocity models produced by ASPIRE 2}

}

\caption{\label{fig-aspire-downstream-rtm}The migration in our final
velocity model is close to the migration in the ground truth model.}

\end{figure}

While the recovered velocity models and uncertainty quality metrics show
significant improvements between ASPIRE 1 and 2, the ultimate goal is to
assess the impact of these improvements on the final product, namely,
the migrated image in areas below the salt. For this purpose, we
included Figure~\ref{fig-aspire-downstream-rtm}, which contains results
of RTM with the ground truth velocity model; the mean of reverse-time
migrations carried out in \(16\) posterior samples for the velocities
produced by ASPIRE 2; error with respect to the migrated image with the
ground truth velocity model; and the standard-deviation of reverse-time
migrations in posterior velocity models. From these plots, the following
observations can be made: First, the migrations in the ground truth
velocity model and the inferred mean of the multiple RTMs are close,
even though issues remain. These include: errors in the delineation some
areas of bottom salt and minor shifts in imaged sedimentary layers below
the salt. While there are errors, we emphasize that errors are expected
to occur in this ill-posed problem, but that our solution comes with
uncertainty quantification that powerfully points to areas of error (see
figures \ref{fig-downstream-error} and \ref{fig-downstream-std}). We
observe that the predicted uncertainty is overly cautious, predicting
larger variability than evidenced in the plot of errors with respect to
the migration with the ground truth velocity model. We consider the fact
that uncertainty being overestimated as advantageous since it offers an
additional safeguard to overinterpretation of imaged reflectors as
opposed to being overconfident. Both migrations suffer from an overprint
due to areas of relatively poor illumination due to the interplay
between small offsets and complexity of the salt geometry, which gives
rise to (de)focusing effects that explain variations in the amplitudes
of the imaged reflectivity under the salt.

\subsection{Field data proof of
concept}\label{field-data-proof-of-concept}

Our aim is to develop ML-enabled workflows for probabilistic velocity
model building capable of scaling to industry-sized seismic datasets. To
demonstrate progress and challenges toward this goal, we apply the same
approach with pre-trained networks from the previous experiments to a
large shallow-water seismic 2D line. The aim of this exercise is
twofold: First, we aim to demonstrate that the presented methodology can
be applied to field data. Second, we aim to showcase the main
limitations of the presented approach---i.e., its reliance on training
datasets that are pertinent to the geological setting under
consideration.

\subsubsection{Shallow-water field data}\label{shallow-water-field-data}

\begin{figure}

\begin{minipage}{0.50\linewidth}

\centering{

\includegraphics[width=1\textwidth,height=\textheight]{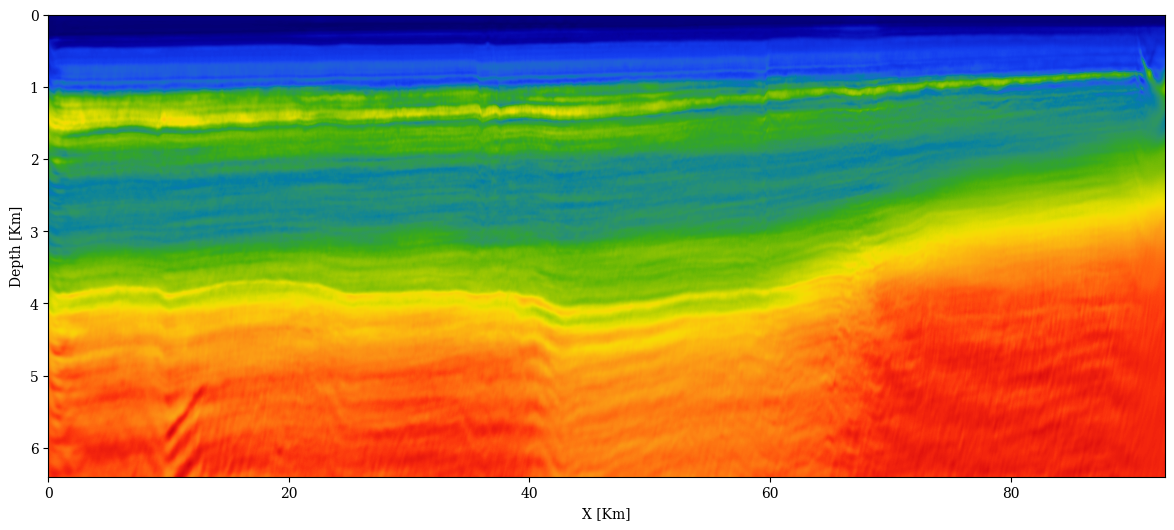}

}

\subcaption{\label{fig-galactic-b}Migration-velocity model}

\end{minipage}%
\begin{minipage}{0.50\linewidth}

\centering{

\includegraphics[width=1\textwidth,height=\textheight]{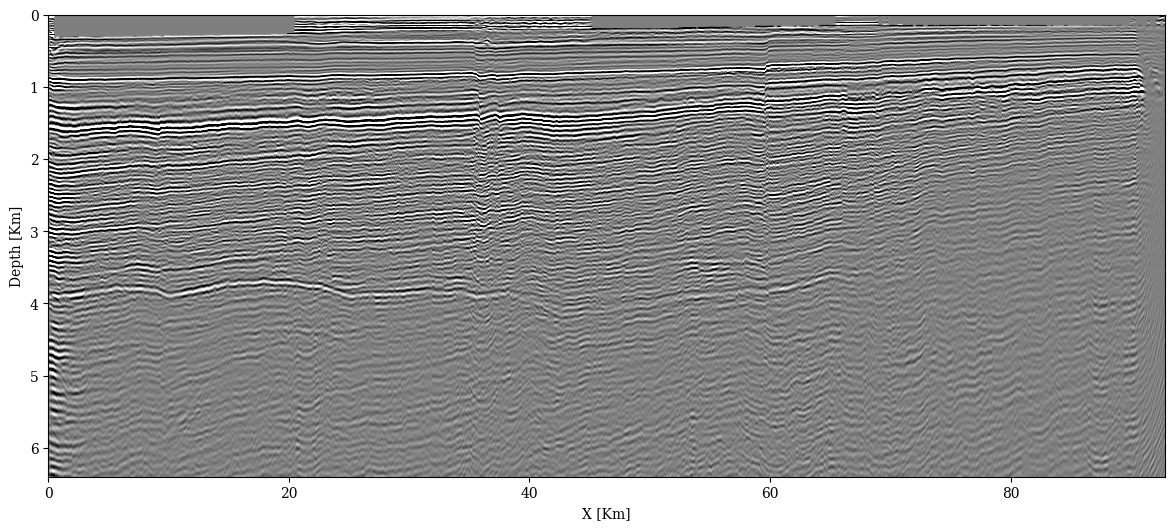}

}

\subcaption{\label{fig-galactic-a}Reverse-time migration}

\end{minipage}%
\newline
\begin{minipage}{0.50\linewidth}

\centering{

\includegraphics[width=1\textwidth,height=\textheight]{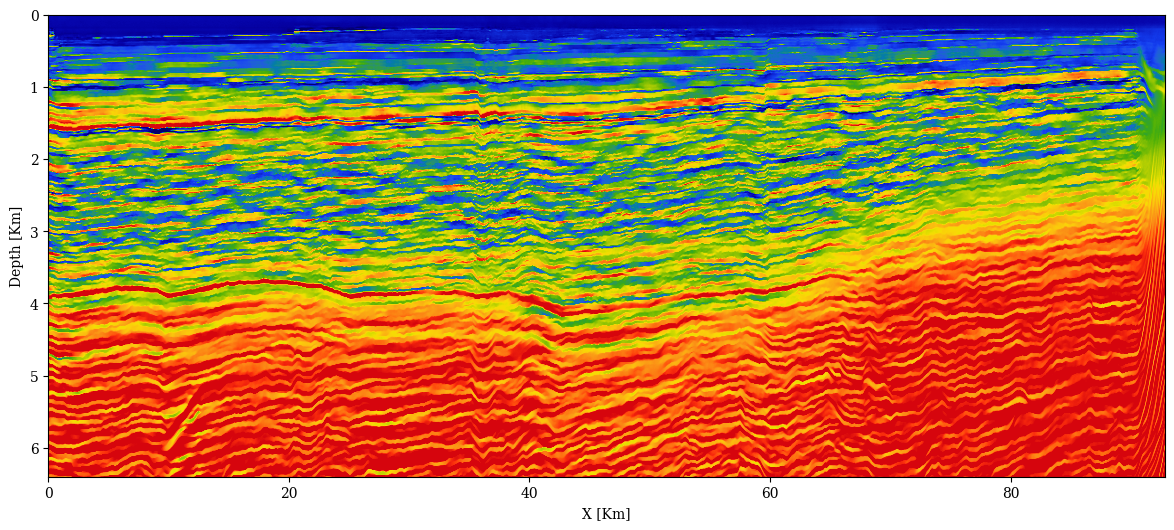}

}

\subcaption{\label{fig-galactic-c}Posterior mean trained on Synthoseis}

\end{minipage}%
\begin{minipage}{0.50\linewidth}

\centering{

\includegraphics[width=1\textwidth,height=\textheight]{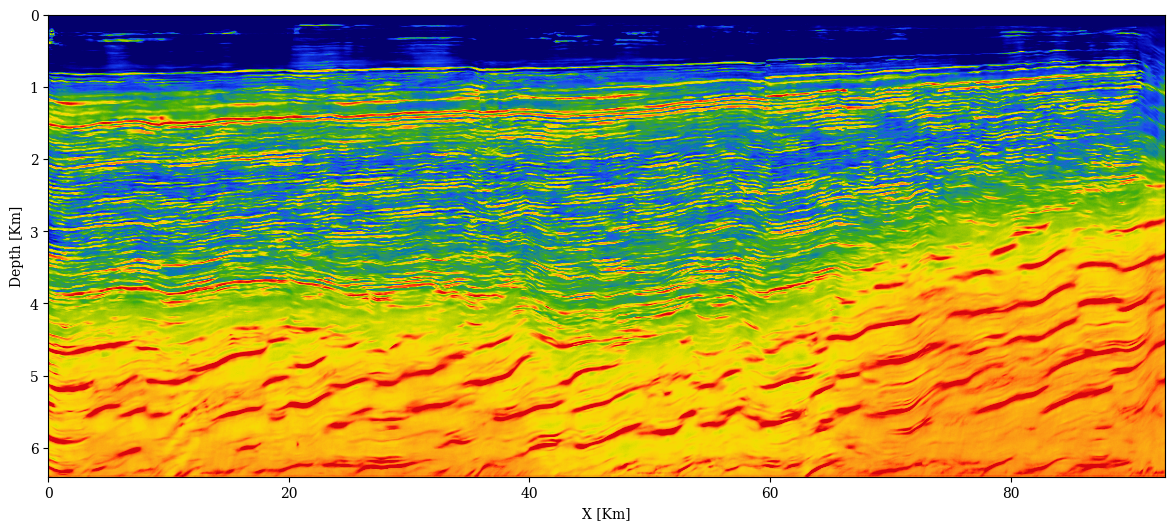}

}

\subcaption{\label{fig-galactic-d}Posterior mean trained on SEAM}

\end{minipage}%
\newline
\begin{minipage}{0.50\linewidth}

\centering{

\includegraphics[width=1\textwidth,height=\textheight]{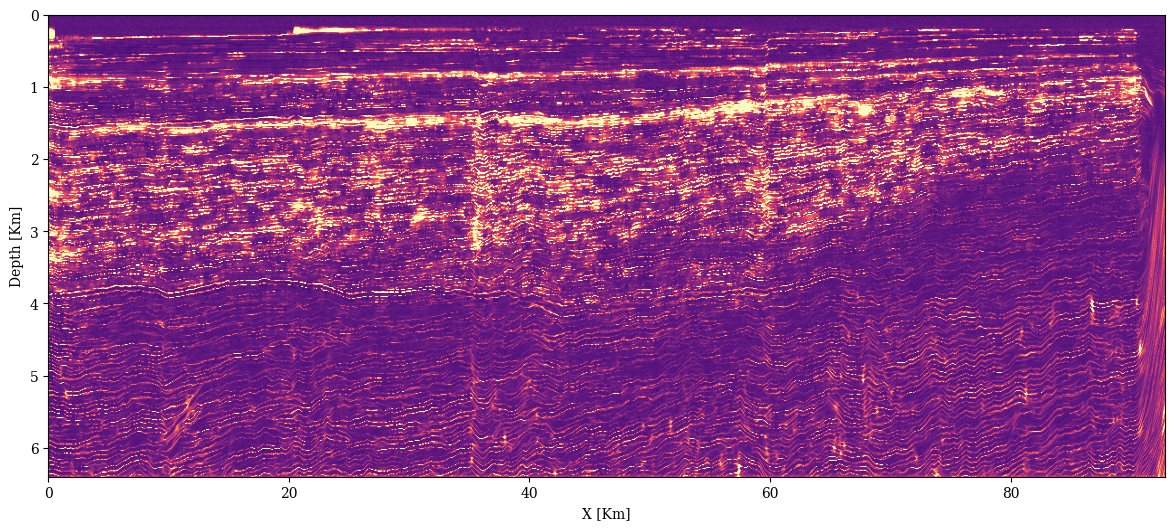}

}

\subcaption{\label{fig-galactic-e}2xUQ trained on Synthoseis}

\end{minipage}%
\begin{minipage}{0.50\linewidth}

\centering{

\includegraphics[width=1\textwidth,height=\textheight]{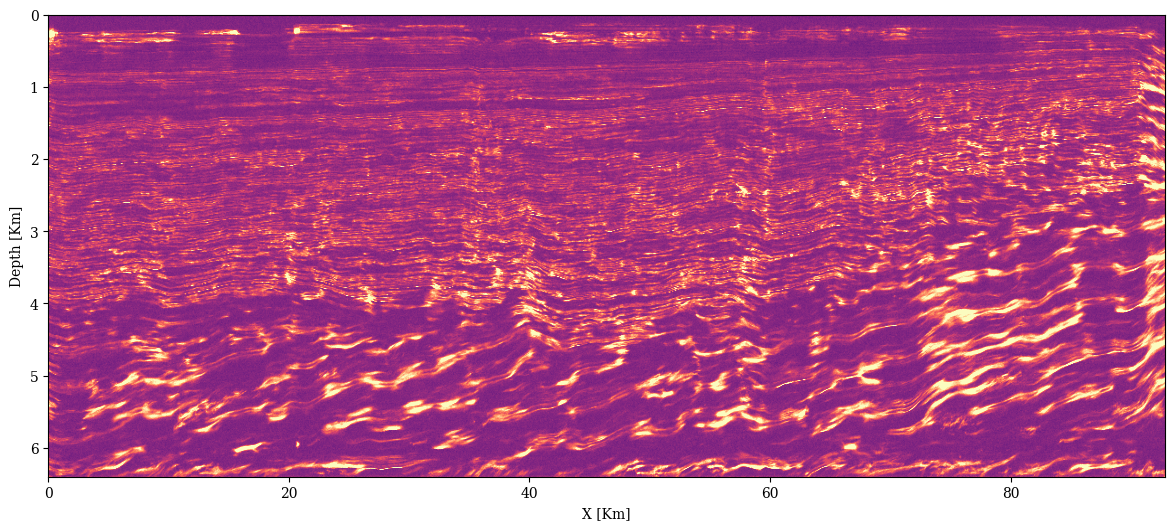}

}

\subcaption{\label{fig-galactic-f}UQ trained on SEAM}

\end{minipage}%

\caption{\label{fig-galactic}Field-data results trained on Synthoseis
versus SEAM. \emph{(a)} Migration velocity model used to produce RTM.
\emph{(b)} Observed RTM. \emph{(c)} Mean of Posterior samples from
Synthoseis. \emph{(d)} Mean of Posterior samples from SEAM. We recommend
zooming in a computer screen to see these figures.}

\end{figure}

We thank Woodside Energy Group for providing access to the dataset
referred to as ``Galactic 2D'', made available under Creative Commons
BY-SA 4.0 licence {[}https://creativecommons.org/licenses/by-sa/4.0/{]}
as part of Phase 2b of the 2023 Galactic 2D Seismic Imaging Study. We
have not made any modifications to the original work. We consider the
migration and velocity models derived from ``Galactic 2D''. These
derivatives are obtained with a traditional workflow, consisting of
migration-velocity building with FWI and RTM, as shown in figures
\ref{fig-galactic-b} and \ref{fig-galactic-a}.

The 2D line of the Galactic dataset being considered has a grid size of
\((512 \times 7024).\) To produce samples, the migration shown in
Figure~\ref{fig-galactic-a} served as input to conditional neural
networks trained on velocity models from the \texttt{Synthoseis} and
SEAM datasets. In this case, we trained a network on \texttt{Synthoseis}
samples that did not contain salt. Since the Galactic dataset does not
include non-zero offset migrations, we use the RTM included in
Figure~\ref{fig-galactic-a} as input, producing the posterior means
shown in figures \ref{fig-galactic-c} and \ref{fig-galactic-d}. For this
large line, each posterior sample takes \(30\) seconds to generate. Both
estimates are obtained from \(64\) samples of the posterior.

The posterior mean estimates produced by either network do not seem
realistic because they are strongly biased toward the datasets these
networks were trained on. The fact that the result obtained with
\texttt{Synthoseis} looks more reasonable likely stems from the fact
that \texttt{Synthoseis} dataset contains more variability in its
training set. Still, this example underlines the importance of having
access to pertinent training data, a topic we will address in the
discussion.

\section{Discussion and future work}\label{discussion-and-future-work}

Results obtained for the stylized examples and complex case studies
suggest that probabilistic velocity model building is possible as long
as training pairs in the form of velocity models and migrated image data
are available. However, having access to representative training
velocity models is challenging in practice. The examples presented here
fall short because they are biased by our geological understanding, as
expressed in synthetic earth models. To overcome these limitations and
other challenges in applying the proposed methodology to more realistic
settings, we strongly encourage our community to follow David Donoho's
\citep{donoho2024data} recipe for success in data science, which
includes \emph{(i)} making (training) datasets available; \emph{(ii)}
releasing research findings that can be reproduced in a frictionless
manner; and \emph{(iii)} establishing benchmarks to compare results
based on quantitative figures of merit. We argue that this work
contributes towards items \emph{(ii-iii)}. Item \emph{(i)} remains
challenging, but below we outline a possible strategy for addressing
this item.

\textbf{A call to arms to curate datasets to train a foundation model.}
By using the latest tools from generative AI,
\citet{erdinc2024generative} recently developed a framework to train
conditional neural networks to generate realistic Earth models
parameterized by velocity. In their approach, Diffusion networks are
trained on pairs imaged field data and well-log data, both residing in
national data repositories such as the one maintained by the National
Transition Authority in the United Kingdom. We are currently in the
process of curating hundreds of 2D seismic image-well pairs, so that a
foundational model can be trained with which realistic synthetic
velocity can be generated without ever requiring access to velocity
models themselves. As more curated datasets become available, the
quality of this foundational model will improve, as will machine
learning techniques that rely on training data.

An important avenue for future work is to explore non-amortized methods
that would build on top of the amortized results shown here.
Specifically, algorithms that use the forward operator at inference time
to specialize to the observation and improve performance
\citep{siahkoohi2023reliable, orozco2024aspire, yin2024wiser}. This
approach becomes especially pertinent when applying our method to field
data.

\section{Conclusions}\label{conclusions}

We implemented machine learning enabled workflows that, through the use
of modern conditional Diffusion neural networks, advance the state of
the art in amortized migration-velocity model building with uncertainty
quantification. In this context, amortization means that our network
generalizes across different shot datasets. We also proposed a set of
performance metrics as a benchmark to compare the effectiveness of
different uncertainty quantification methods. For complex salt
scenarios, we developed a new iterative workflow incorporating salt
flooding. Using our Bayesian probabilistic approach, multiple
migration-velocity models are generated from poor initial velocity
models (e.g., models without salt). These samples from the posterior
distribution produce different reverse-time-migrated images that contain
variability propagated from the uncertainty in the inferred velocity
model. Thanks to machine learning, the proposed approach remains
computationally feasible at inference time, with considerable but
manageable offline training costs. The results represent a demonstration
of machine learning enabled probabilistic velocity building from
short-offset acoustic reflection-only data. While our evaluation on
field datasets is still in its early stages, the initial results are
promising. However, applying the method to field data revealed a bias
toward the synthetic data distribution on which the networks were
trained. This finding highlights the need to curate realistic training
datasets to train the next generation of generative networks for solving
geophysical inverse problems.

\subsection{Acknowledgements}\label{acknowledgements}

We are grateful to the Georgia Research Alliance and the ML4Seismic
Center partners for their support in this work. We thank Devito Codes
for providing a complimentary license and support of DevitoPro which
greatly increased efficiency in producing the wave simulations. We thank
Woodside Energy Group for providing access to the dataset referred to as
``Galactic 2D'', made available under Creative Commons BY-SA 4.0 licence
{[}https://creativecommons.org/licenses/by-sa/4.0/{]} as part of Phase
2b of the 2023 Galactic 2D Seismic Imaging Study. We also thank Charles
Jones (Osokey) for the valuable discussion.

  \bibliography{paper.bib}

\end{document}